\documentclass[10pt]{article} 
\usepackage[preprint]{tmlr}
\usepackage{amsmath,amssymb,amsfonts}
\usepackage[ruled,vlined,linesnumbered]{algorithm2e}
\usepackage{float}
\usepackage{graphicx}
\usepackage{textcomp}
\usepackage{booktabs}
\usepackage{xcolor}
\usepackage{pdflscape}
\usepackage{comment}
\usepackage{multirow}
\usepackage{forest}
\usepackage{array}
\usepackage{siunitx}
\usepackage{float}
\usepackage[table]{xcolor}
\usepackage{tikz}
\usetikzlibrary{positioning}
\usetikzlibrary{arrows.meta}
\usepackage{svg}
\usepackage{titlesec}
\usepackage{url}
\usetikzlibrary{arrows.meta,positioning}

\usepackage{placeins}
\usepackage{booktabs,tabularx}
\usepackage{enumitem}
\usepackage{threeparttable,xcolor}
\usepackage{wrapfig}

\usepackage{amsmath,amsfonts,bm}

\def\eqref#1{equation~\ref{#1}}
\def\Eqref#1{Equation~\ref{#1}}

\def\1{\bm{1}}

\DeclareMathAlphabet{\mathsfit}{\encodingdefault}{\sfdefault}{m}{sl}
\SetMathAlphabet{\mathsfit}{bold}{\encodingdefault}{\sfdefault}{bx}{n}

\usepackage{hyperref}
\usepackage{url}
\usetikzlibrary{calc}

\title{Spatons: Spatiotonally Localized Building Blocks
for \\ Efficient XR Predistortion}

\author{\name Mohammadjavad Maheronnaghsh \email mj.maheronnaghsh@mail.utoronto.ca \\
      \addr Department of Electrical and Computer Engineering\\
      University of Toronto
      \AND
      \name Sushant Kondguli \email sushantkondguli@meta.com \\
      \addr Meta Reality Labs
      \AND
      \name Steve Mann \email mann@eecg.toronto.edu \\
      \addr Department of Electrical and Computer Engineering\\
      University of Toronto
      \AND
      \name Andreas Moshovos \email moshovos@eecg.toronto.edu \\
      \addr Department of Electrical and Computer Engineering\\
      University of Toronto\\
      Vector Institute}
\newcommand{\fittable}[1]{%
  \resizebox{\ifdim\width>\linewidth \linewidth\else \width\fi}{!}{#1}}
\newcommand{\tablestyle}{\footnotesize\setlength{\tabcolsep}{4pt}\renewcommand{\arraystretch}{1.05}}

\newcommand{\fitimage}[2][]{%
  \sbox0{\includegraphics[#1]{#2}}%
  \ifdim\wd0>\linewidth
    \includegraphics[width=\linewidth,#1]{#2}%
  \else
    \usebox0%
  \fi}

\def\month{MM}  
\def\year{YYYY} 
\def\openreview{\url{https://openreview.net/forum?id=XXXX}} 

\begin{document}

\maketitle

\begin{abstract}
We present Spatons, a family of compositional, spatially localized neural operators for computationally-efficient XR (eXtended reality) predistortion and distortion compensation. XR displays (AR, VR, and MR) suffer from a mix of systematic spatial (e.g., geometrical) and tonal (e.g., colour) distortions. To compensate for these distortions, frames can be pre-reverse-distorted so that the projected frame looks as intended. Since the distortions vary due to manufacturing tolerances and across projection systems, and since many XR systems are untethered and battery operated, a programmable, low energy solution is desirable. Low latency is also desirable to avoid inducing motion sickness. A Spaton is a hybrid patch-to-pixel block that combines tiny Multi-Layer Perceptrons (MLPs) with fixed-function interpolation. Using Spatons as building blocks we demonstrate effective, programmable, and computationally efficient pipelines for a variety of distortion scenarios.  On the full spatiotonal Scenario~5, Sequential-5$'$ matches CompenNet++ in SSIM (0.944 versus 0.944) and closely matches it in PSNR (27.5 versus 27.7\,dB), while remaining substantially more lightweight. We demonstrate a flexible Spaton solution using $6{\times}$ fewer FLOPs and $106{\times}$ fewer parameters vs.\ CompenNet++, and $4$--$10{\times}$ fewer memory transfers vs.\ UNet++. We characterize several efficient software implementations and analytical hardware estimates of Spaton pipelines along with appropriate optimizations and across several target XR platforms. We consider distortions drawn mostly from see-through, near-eye displays, where the spatial light path directly governs local intensity and spectral response.  Testing Spatons on a wider range of real XR devices is left to future work. Accordingly, we present this as an early, systems-level exploration: models are trained and evaluated on closed-form differentiable distortions and one real projector--camera capture set, not through-the-optics measurements. Chromatic dispersion (color fringes) remains the main artifact where Spatons trail the monolithic baselines.
\end{abstract}

\section{Introduction}
\label{sec:intro}

XR systems (AR, VR (virtual reality), MR (mixed reality), etc.) are moving from niche prototypes to
mass-market platforms, yet a persistent obstacle to visual
fidelity is \emph{distortion} introduced by the optical sensing and
projection hardware.\footnote{Large language models were used for revising code and text.}
There are three broad categories of such distortions encountered in practice: 
(1)~\emph{Spatial} (geometric, topological, warping, local
stretching, etc.);
(2)~\emph{Tonal} (photometric, radiometric, quantigraphic, or image-based tonescale adjustments, such as pixel quantity or contrast shifts, color casts / color-correction, etc.), and
(3)~\emph{Spatiotonal} (e.g., vignetting and spatially-varying color
shifts)~\cite{mannist, mann1998headmounted, mann2000comparametric, intelligentimageprocessing, tuan2006spatiotonal, ali2012comparametric, banterle2012low, antunes2022content, antunes2025cas}. The distortions vary across different hardware and often across different instances of the same hardware due to variations in manufacturing.

The standard remedy is \emph{predistortion}: apply an inverse transformation to the rendered frame so that the observed image appears correct after passing through the optics. Ideally, the predistortion system would be programmable to allow it to adapt to changes in the projection system and to variations due to manufacturing tolerances. Moreover, it would be low energy allowing deployment even in untethered, battery operated applications. Low latency is also desirable to avoid introducing
motion-to-photon latency that induces user nausea~\cite{Stauffert2020,Jerald2009}.

Past pre-distortion approaches fall in the following categories: Closed-form parametric approaches are efficient but inherently
rigid, and {struggle with} complex or coupled spatiotonal effects in real deployments. 
These motivated data-driven methods which repurpose monolithic \texttt{img2img} neural models
(e.g., UNet++, CompenNet++~\cite{zhou2018unet++,
huang2019compennet++}).
They are programmable, enabling adaptation to different projection systems and even device instance variations.  Although accurate, a UNet++-Large baseline requires
$\approx$12.1~GFLOPs and 1392~MB of {memory transfers} per
$256{\times}256$ frame, far exceeding the {representative ($\approx$2\,W)}
compute budget of wearable devices and requiring high-performance compute hardware to maintain acceptable latency. 


We formulate predistortion as a \emph{learned patch-to-pixel transformation}: for each output pixel, a small $N{\times}N$ input neighborhood provides the local context needed to compute the corrected output. Based on this formulation, we introduce Spatons, \textit{spatial-tonal operators} for AR predistortion that reduce compute and memory costs while enabling effective and programmable distortion compensation. 
A Spaton is a compact, programmable module built around a tiny machine-learning model. 
This model allows each Spaton to learn and compensate for spatial, tonal, or coupled spatiotonal distortions. 
Multiple Spatons can be composed into a short pipeline. 
This compositional structure allows spatial and tonal correction to be optimized 
separately while still supporting coupled AR distortion pipelines~\cite{mann2000comparametric,intelligentimageprocessing,tuan2006spatiotonal,chapiro2024ar}.

Spatons are trained offline for each device using paired undistorted reference frames and their corresponding distorted frames. This allows them to learn the device-specific behavior of the projection subsystem. Because they operate directly on local patches, Spatons avoid the dense intermediate feature maps of conventional image-to-image pipelines, substantially reducing compute and {memory-transfer}.  

Each Spaton is implemented as a tiny multilayer perceptron (MLP), with up to three layers and up to 64 hidden features. A Spaton may predict either (i)~a continuous sampling coordinate for spatial correction, or (ii)~a local tonal $\Delta$RGB adjustment for photometric, radiometric, or image-based correction. A hardware-friendly fixed-function bilinear interpolation unit performs the final resampling. When distortions are spatiotonally inseparable, specialized Spatons are composed sequentially to provide holistic correction without sacrificing efficiency.

We characterize the performance and costs of measured software implementations and analytically modelled hardware configurations across a variety of distortions and target platforms. 

 {The main contributions of this work are:}


\begin{enumerate}[leftmargin=*, rightmargin=0pt]

  \item \textbf{Compositional local neural architectures:}
  We introduce and characterize a family of tiny MLPs (\textbf{C-Res} and
  \textbf{{Patch2Pixel-v2}} for non-geometric, \textbf{G-Res} and
  \textbf{G-Hybrid} for spatial distortions) and demonstrate that they can effectively compensate expected distortions while being extremely lightweight in terms of number of operations, memory footprint and memory transfers. 

  \item \textbf{Joint spatiotonal correction via composition:}
  We demonstrate that sequential pipelines of 2 models (3.4-10.8K parameters,
  0.44-1.41\,GFLOPs) match monolithic quality.
  Our characterization shows that our {Sequential-5$'$} pipeline
 (\textbf{G-Res}$\to$\textbf{Patch2Pixel-v3}) is effective in compensating for a variety of distortion scenarios including tonal, spatial, and spatiotonal ones. Our preferred {Sequential-5$'$} pipeline achieves SSIM~0.944 / PSNR~27.5\,dB, matches CompenNet++ (0.944 / 27.7\,dB) while having $6{\times}$ fewer FLOPs and $106{\times}$ fewer parameters vs.\ CompenNet++, and $4$--$10{\times}$ fewer memory transfers vs.\ UNet++-Large. 

  \item 
  We evaluate across a diverse set of multiple
  mixed-distortion scenarios on \texttt{Images.cv}.
  G-Res achieves average SSIM~0.984 vs.\ 0.892 for UNet++-Large
  on five geometric types ($-$95.2\% FLOPs); C-Res / {Patch2Pixel-v2}
  match or exceed UNet++-Large photometrically at
  $13$--$20{\times}$ fewer FLOPs.

  \item \textbf{Implementation Optimizations:}
  Decoupling geometric mapping from color synthesis enables
  implementing the geometric predistortion model as a lookup table that reduces G-Res inference to $O(1)$
  operations per pixel (99.99\% compute reduction) which is not directly available to conventional img2img architectures because their outputs depend on image content and dense intermediate activations.

\item \textbf{Video workload characterization.} We consider taking advantage of the expected few differences across consecutive frames to reduce how often the MLPs of Spatons need to be invoked, and demonstrate that our lookup-table optimization renders this inter-frame reuse ineffective in practice.  

\end{enumerate}

We use ``geometric'' and ``spatial'' interchangeably. We use ``non-geometric'' to denote tonal and spatiotonal distortions that do not involve coordinate warping.

\textbf{Scope and limitations.} This is an early, systems-level study of learnable predistortion. Training and evaluation use closed-form differentiable distortion generators together with one real projector--camera capture set (Sec.~\ref{sec:realdata}). We do not claim through-the-optics validation under real device variation, sensor noise, calibration error, or temporal drift, and we treat these as future work (Sec.~\ref{sec:conclusion}). While the distortions we study are drawn mostly from see-through, near-eye displays, predistortion is a general distortion-compensation problem, and the Spaton pipeline is not tied to a specific display type: It may therefore be applicable to different XR devices. We evaluate it here using closed-form distortions motivated primarily by see-through displays and one real projector--camera capture set. Testing on physical XR devices, including AR and VR headsets, is left to future work.

\section{Background}
\label{sec:related}

\noindent\textbf{AR waveguide optics: }
Waveguide combiners dominate OST-AR (Optical See-Through Augmented Reality) displays due to their thin form
factor and pupil-expansion properties~\cite{xiong2021augmented,
ding2023waveguide}.
Geometric (reflective) waveguides offer superior color fidelity, while
diffractive designs (HoloLens, Magic Leap) are more scalable but
introduce wavelength-dependent artifacts from their grating
structures~\cite{ding2023waveguide, chapiro2024ar}.
Both architectures introduce systematic spatiotonal distortions necessitating real-time processing to maintain
fidelity~\cite{intelligentimageprocessing}.

\noindent\textbf{Digital compensation and neural predistortion: }
Early LUT-based (Lookup Table) methods lack flexibility for device-specific or
dynamic variations~\cite{li2020distortion}.
Data-driven approaches such as \textbf{CompenNet++}~\cite{huang2019compennet++}
and \textbf{UNet++}~\cite{zhou2018unet++} achieve high reconstruction
quality by treating compensation as an end-to-end img2img task, but
their monolithic architectures require $>$12\,GFLOPs and
$>$1392\,MB of {memory transfers} per $256{\times}256$
frame which is prohibitive for untethered headsets where  compute budgets are severely constrained (e.g.,  $<$2\,W).
Existing methods also typically address geometric \emph{or}
photometric correction in isolation~\cite{huang2019compennet++,
li2020distortion}.
Our patch-to-pixel framework jointly corrects coupled spatiotonal
distortions {with substantially less computation and a smaller memory footprint}, enabling real-time performance on resource-constrained hardware.

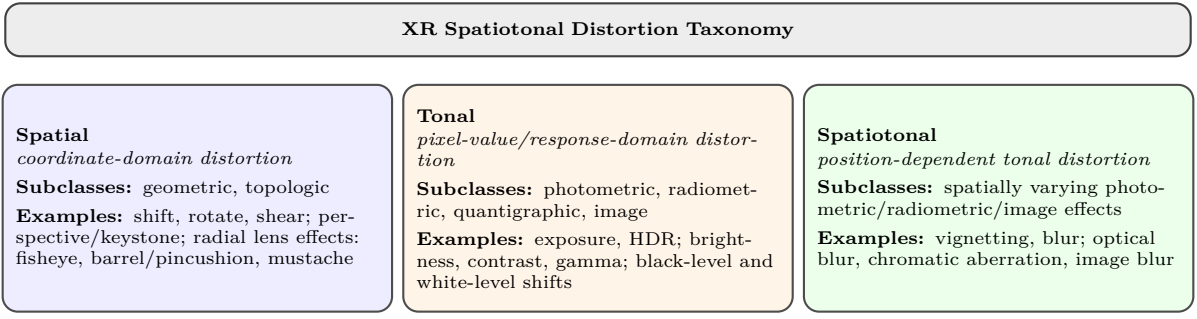
\begin{figure*}[t]
\centering
\begin{tikzpicture}[
    font=\scriptsize,
    root/.style={
        rectangle, rounded corners=2mm, draw=black!75, fill=gray!14,
        thick, align=center, text width=15.4cm, minimum height=0.7cm,
        inner sep=4pt
    },
    card/.style={
        rectangle, rounded corners=2mm, draw=black!70, thick,
        align=left, text width=4.8cm, minimum height=3.0cm,
        inner sep=5pt, anchor=north
    }
]

\node[root] (root) at (0,0) {\textbf{XR Spatiotonal Distortion Taxonomy}};

\node[card, fill=blue!7]   (spatial)     at ($(root.south)+(-5.3,-0.35)$) {%
\textbf{Spatial}\\[0.3mm]
\emph{coordinate-domain distortion}\\[1mm]
\textbf{Subclasses:} geometric, topologic\\[1mm]
\textbf{Examples:} shift, rotate, shear; perspective/keystone; radial lens effects: fisheye, barrel/pincushion, mustache
};

\node[card, fill=orange!9] (tonal)       at ($(root.south)+(0,-0.35)$) {%
\textbf{Tonal}\\[0.3mm]
\emph{pixel-value/response-domain distortion}\\[1mm]
\textbf{Subclasses:} photometric, radiometric, quantigraphic, image\\[1mm]
\textbf{Examples:} exposure, HDR; brightness, contrast, gamma; black-level and white-level shifts
};

\node[card, fill=green!8]  (spatiotonal) at ($(root.south)+(5.3,-0.35)$) {%
\textbf{Spatiotonal}\\[0.3mm]
\emph{position-dependent tonal distortion}\\[1mm]
\textbf{Subclasses:} spatially varying photometric/radiometric/image effects\\[1mm]
\textbf{Examples:} vignetting, blur; optical blur, chromatic aberration, image blur
};

\end{tikzpicture}
\caption{Taxonomy of XR spatiotonal distortions. Spatial, tonal, and spatiotonal distortions are sibling families acting on coordinates, pixel values, or position-dependent tonal response.}
\label{fig:taxonomy}
\end{figure*}

\noindent\textbf{Spatiotonal distortion modeling: } We consider a variety of distortions common to XR systems. Fig.~\ref{fig:taxonomy} presents our taxonomy which is informed by
the AR-DAVID dataset~\cite{chapiro2024ar} and
ColorVideoVDP~\cite{mantiuk2024colorvideovdp}.
Tonal (e.g. photometric, radiometric, quantigraphic, image color, etc.) effects such as Waveguide Non-Uniformity (WGNU) from interference
patterns and Elevated Black Level from ambient-light
mixing, are grounded in comparametric imaging
theory~\cite{mann2000comparametric, ali2013comparametric}.
Spatial (geometric or topological) errors (keystone, fisheye, barrel, mustache) are modeled with established optical
formulations~\cite{hornacek2022spatial, li2020distortion,
liu2021optical, fitzgibbon2001simultaneous}.
Unlike general corruption benchmarks~\cite{hendrycks2019benchmarking},
we focus strictly on the coupled artifacts intrinsic to near-eye displays. Appendix~\ref{sec:supp-distortions} (Table~\ref{tab:distortions}) lists the
distortions considered in this work and how they were modeled.

\noindent\textbf{Synthetic distortion as an evaluation protocol: }
Training and quantitatively evaluating predistortion on algorithmically
generated distortions is established practice when large paired real
datasets are unavailable. Blind geometric distortion correction~\cite{li2019blind},
FishEyeRecNet~\cite{yin2018fisheyerecnet} (which also evaluates on real
fisheye images), and QueryCDR~\cite{guo2024querycdr} build training and test
data from parametric radial/projective models overlapping those in
Table~\ref{tab:distortions}; Kageyama et al.~\cite{kageyama2024efficient}
synthesize geometric distortion and defocus for dynamic projection mapping.
The XR display-quality datasets AR-DAVID~\cite{chapiro2024ar} and
XR-DAVID~\cite{mantiuk2024colorvideovdp} algorithmically synthesize blur,
color fringes, and waveguide non-uniformity, and their perceptual impact was
measured on a physical OST-AR testbed. Our geometric models follow the
optical formulations cited above. To our knowledge, no prior near-eye XR
predistortion work jointly evaluates spatial, tonal, and coupled spatiotonal
distortions with a modular, sequentially composed pipeline designed for
untethered compute budgets: the closest projector--camera baseline,
CompenNet++, uses $106{\times}$ the parameters and ${\approx}6{\times}$ the
FLOPs of our Sequential-5$'$ under the common $256{\times}256$ accounting of
Table~\ref{tab:model_summary}.

\section{Problem Formulation}
\label{sec:problem}

The objective of neural predistortion is to find an inverse mapping
of the system distortion $\mathcal{D}_S$ such that
\begin{equation}
  I_{obs} = \mathcal{D}_S(I_{pre}) \approx E,
  \label{eq:predistort}
\end{equation}
where $I_{pre}$ is the predistorted input and $E$ is the expected image obtained by resizing the source image $I$ into the system's Region of Interest (ROI) to account for projection fields
that exceed the optical combiner's usable area.

\noindent\textbf{Forward Distortion Process: }
The optical pipeline (light engine + viewing optic) applies $n$
sequential distortion stages:
\begin{equation}
  O = \mathcal{D}_S(I) = d_n(d_{n-1}(\dots d_1(I)\dots)).
  \label{eq:forward}
\end{equation}
We assume {\emph{a priori}} calibration: the type, order, and strength of
each stage $d_i$ are known. This assumption matches the common per-device calibration setting for near-eye displays, where optical correction parameters are estimated during manufacturing or device calibration.
Knowing these parameters does not make a closed-form
inverse sufficient: the coupled spatiotonal term in~\Eqref{eq:coupling} is not analytically invertible in general,
and per-unit optical variation (constraint~(i)) means no single
closed-form inverse transfers across devices. The learned predictor
absorbs both the coupling and the per-device residual while remaining
a fixed-cost per-pixel operation.
For our study we initially use this known-distortion setting:  we use the calibrated forward model to generate clean/distorted image pairs and train the predistortion model through it. We follow these studies by emulating scenarios where the forward model is not known in closed form. We do so by training the model from image pairs instead, using the two-phase scheme of Section~\ref{sec:twophase}.

\noindent\textbf{Spatiotonal Coupling and System Constraints: }
In waveguide-based AR, the spatial light path directly dictates
photometric response.
A distorted pixel at coordinate $\mathbf{u}{=}[x,y]^T$ is modelled as
\begin{equation}
  I_{obs}(\mathbf{u}) =
    \mathcal{T}_{\text{tonal}}\!\left(I\!\left(
    \mathcal{T}_{\text{spatial}}(\mathbf{u})\right),\mathbf{u}\right),
  \label{eq:coupling}
\end{equation}
where $\mathcal{T}_{\text{spatial}}$ encodes geometric warping
(e.g., Fisheye, Keystone) and $\mathcal{T}_{\text{tonal}}$ encodes
photometric shifts (e.g., WGNU, Color Fringes).
Inverting $\mathcal{D}_S$ must satisfy three constraints:
(i)~\textbf{Manufacturing variance}: per-device calibration is
required due to unit-to-unit optical differences;
(ii)~\textbf{Hardware envelope}: the solution must fit within a
typical compute budget to prevent latency-driven motion sickness. A $<$2\,W power budget is used here as a
representative envelope; and
(iii)~\textbf{Environmental dynamics}: tonal distortions such as
elevated black level vary with ambient light, requiring real-time
adaptation.

\section{Spatons}
\label{sec:method}
Spatons are compact pre-distortion building blocks built around tiny neural networks so that they can be trained to compensate for device-specific distortion characteristics. Section~\ref{sec:train} describes the Spaton training methodology and  Section~\ref{sec:arch} details the various Spatons we considered.

\subsection{Training Procedure}
\label{sec:train}
\begin{figure*}[t]
  \centering\tablestyle
  \includegraphics[width=0.7\textwidth]{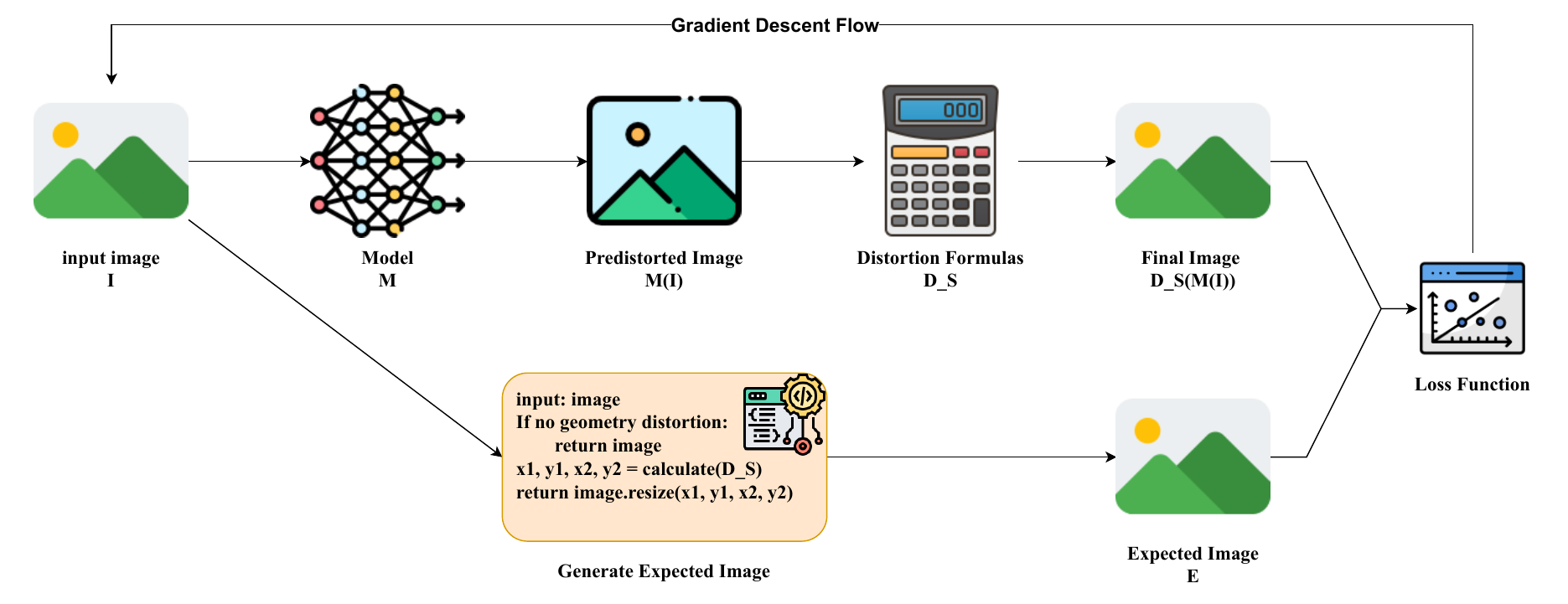}
  \caption{End-to-end training overview.}
  \label{fig:training_process}
\end{figure*}

A clean source image $I$ is passed through model $M$ to produce
a predistorted frame $M(I)$, which is then subjected to the
differentiable distortion pipeline $\mathcal{D}_S$ to yield
$I_{obs}{=}\mathcal{D}_S(M(I))$.
Gradients are back-propagated through $\mathcal{D}_S$ to minimise
the discrepancy between $I_{obs}$ and the expected image $E$
(Fig.~\ref{fig:training_process}).
When geometric distortions are present, $E$ is defined as the
largest axis-aligned square inscribed within the distorted
boundary (found in $O(L^2)$) and then resized via bilinear
interpolation, ensuring the model compensates for both pixel-level
distortion and the global ROI mapping.
Without geometric distortion, $E{=}I$.

Our loss function combines pixel-level fidelity (MSE, $L_1$~\cite{hodson2022root})
with perceptual quality (LPIPS~\cite{zhang2018unreasonable} with the AlexNet~\cite{krizhevsky2012imagenet} backbone):
\begin{equation}
\mathcal{L} = \text{MSE}(I_{obs},E) + 0.5\cdot L_1(I_{obs},E)
               + 0.3\cdot\text{LPIPS}(I_{obs},E)
  \label{eq:loss}
\end{equation}

The dataset and training hyperparameters are detailed in Section~\ref{sec:methodology}.

\subsection{Formula-Agnostic Two-Phase Training}
\label{sec:twophase}
The procedure above back-propagates through a \emph{known}, differentiable
distortion pipeline $\mathcal{D}_S$. Spatons do not require this. When only
paired clean/distorted images are available (e.g., real projector--camera
captures or a black-box renderer) we train in two phases. In \emph{Phase~1}
a forward-model network $\mathrm{NN}_1$ is fit to reproduce the distortion
from image pairs alone, $\text{clean}\!\to\!\mathrm{NN}_1\!\to\!\text{distorted}$;
$\mathrm{NN}_1$ may be arbitrarily large since it is used only at training time.
In \emph{Phase~2} we freeze $\mathrm{NN}_1$ and train the compact deployable
predistortion model $\mathrm{NN}_2$ through it,
$\text{clean}\!\to\!\mathrm{NN}_2\!\to\!\mathrm{NN}_1(\text{frozen})\!\to\!\text{clean}$.
Because gradients flow through the learned $\mathrm{NN}_1$, the true forward
process need not be differentiable or known in closed form, while Phase~2
retains a fully differentiable path for training $\mathrm{NN}_2$.

\subsection{Spaton Block Architecture}
\label{sec:arch}


Our framework recasts predistortion as a \emph{patch-to-pixel}
transformation: each output pixel is predicted independently from a
small $P{\times}P$ neighbourhood, avoiding the dense intermediate
feature maps of img2img pipelines.
We propose two composable model families respectively for  geometric and non-geometric distortions. Here, we focus on pipelines that
apply geometry compensation first, followed by tonal compensation.

\begin{figure*}
\centering\tablestyle
\includegraphics[width=0.9\linewidth]{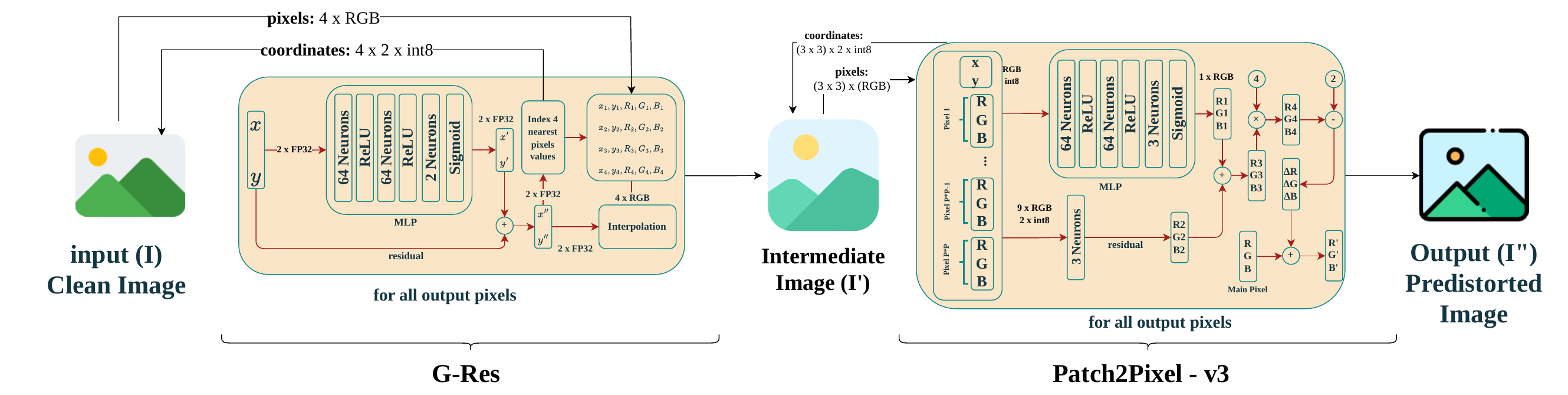}
\caption{The Sequential-5$'$ pipeline: a G-Res geometric predistortion
stage followed by a Patch2Pixel-v3 tonal predistortion stage. RGB residuals are
normalized to $[-2,+2]$. Data types: \texttt{int8} = 8-bit integer pixel index;
\texttt{FP32} = 32-bit float; RGB = one 3-channel pixel.}

\label{fig:gres}
\end{figure*}

Algorithm~\ref{alg:seq5prime} shows an example of such a pipeline comprising two stages and Fig.~\ref{fig:gres} shows an example pipeline using our preferred MLPs (described in the next section): G-Res and Patch2Pixel-v3. The first pipeline stage processes the input frame $I$ compensating for geometric distortions producing the intermediate frame $I'$. The second stage, processes $I'$ to compensate for tonal distortions producing the final output frame $I''$. 

The loop of line 1 implements the first stage. It iterates over the \textit{output} pixels producing the intermediate frame $I'$. For each output pixel the model predicts the input frame $I$ coordinates $(x',y')$ that should be used to generate it (colour values are irrelevant because warping is image-independent). These coordinates are continuous 
{and} are used to fetch the four nearest discrete pixels from $I$. Those pixels are used by the fixed-function interpolation unit (lines 4-5) to produce the output pixel. Lines 7-10 implement the second stage. For each output pixel $(x,y)$ of $I''$ (the final output frame), it fetches a $P\times P$ patch of pixels centered at $(x,y)$ (for the evaluation we use $P=3$ which we found, empirically, to work well). The coordinates $(x,y)$ and the $P^2$ pixels (RGB values) are fed as a flattened vector input to this stage's MLP. The MLP produces an RGB pixel that, along with the input pixels, is used to produce the final pixel value at $(x,y)$ of $I''$.

\begin{algorithm}[t]
\caption{Sequential~5$'$: G-Res$\rightarrow$ Patch2Pixel-v3}
\label{alg:seq5prime}
\SetAlgoLined
\DontPrintSemicolon

\KwIn{Image $I \in \mathbb{R}^{H \times W \times 3}$; patch size $P$;
      trained weights of $\mathrm{MLP}_{\mathrm{geo}}$ and $\mathrm{MLP}_{\mathrm{ton}}$}
\SetKwInOut{Intermediate}{Intermediate}
\Intermediate{Image $I' \in \mathbb{R}^{H \times W \times 3}$; geometric correction}
\SetKwInOut{Input}{Output}
\KwOut{predistorted image $I'' \in \mathbb{R}^{H \times W \times 3}$}

\BlankLine
\tcp{Stage 1: Geometric Correction (G-Res)}
\For{each pixel $(x,y)$ of $I'$}{
    $(x',y') \leftarrow \mathrm{MLP}_{\mathrm{geo}}(x,y)$
    \tcp*[r]{predict sub-pixel source coords}
    $(x_0,y_0) \leftarrow \lfloor(x',y')\rfloor$;\quad
    $(d_x,d_y) \leftarrow (x'-x_0,\; y'-y_0)$\;
    $I'(x,y) \leftarrow (1{-}d_x)(1{-}d_y)\,I(x_0,y_0)
                      + d_x(1{-}d_y)\,I(x_0{+}1,y_0)$\;
    \phantom{$I'(x,y) \leftarrow$}$
                      + (1{-}d_x)d_y\,I(x_0,y_0{+}1)
                      + d_x d_y\,I(x_0{+}1,y_0{+}1)$\;
}

\BlankLine
\tcp{Stage 2: Tonal Refinement (Patch2Pixel-v3)}
\For{each pixel $(x,y)$ of $I''$}{
    $\mathbf{f} \leftarrow \bigl[\mathrm{flatten}(x,\; y,\;  P{\times}P\text{ patch of }I'\text{ centred at }(x,y))\bigr]
     \in \mathbb{R}^{3P^2+2}$\;
    $\Delta_{\mathrm{RGB}} \leftarrow 4\cdot\sigma\bigl(\mathrm{MLP}_{\mathrm{ton}}(\mathbf{f})\bigr)-2$
    \tcp*[r]{residual $\in [-2,2]^3$}
    $I''(x,y) \leftarrow \mathrm{clamp}\bigl(I'(x,y)+\Delta_{\mathrm{RGB}},\,[0,1]\bigr)$\;
}
\Return{$I''$}\;

\BlankLine




\end{algorithm}

\subsubsection{Geometric Distortion Compensation}
We evaluate two geometric compensation models. \textbf{G-Res} is a tiny 3-layer residual MLP that maps $(x,y)\!\to\!(x',y')$ followed by the bilinear interpolation above (4{,}482 parameters at $h{=}64$; we also use $h{=}32,16$), and \textbf{G-Hybrid} is a another variant adding a shallow CNN branch. G-Res is the geometric stage of our featured Sequential-5$'$ pipeline. Full descriptions are in Appendix~\ref{sec:supp-archdetail}.

Applying the geometric compensation model produces an intermediate frame which becomes the input to the tonal compensation model.

\subsubsection{Tonal models}

We evaluate several tonal models, all 3-layer MLPs operating per pixel. All tonal models process pixels independently and ingest spatial
coordinates $(x,y)$ alongside colour values $(R,G,B)$ to handle
spatially-varying artifacts such as WGNU. \textbf{C-Res} takes a single pixel $(x,y,R,G,B)$ and predicts a normalized residual $\Delta$RGB $\in[-2,2]$; \textbf{C-Hybrid} adds a shallow CNN branch. The \textbf{Patch2Pixel} family instead ingests a $P{\times}P$ patch: \textbf{v1} predicts absolute RGB (an ablation of residual learning), \textbf{v2} predicts a normalized residual with per-pixel coordinates ($\mathbf{f}\in\mathbb{R}^{5P^2}$), and \textbf{v3} reduces the spatial input to center coordinates only ($\mathbf{f}\in\mathbb{R}^{3P^2+2}$), removing $2(P^2{-}1)$ inputs with negligible quality loss (Table~\ref{tab:mixed_scenario5}). Patch2Pixel-v3 is the tonal stage of our featured Sequential-5$'$ pipeline. Full descriptions are in Appendix~\ref{sec:supp-archdetail}.

Residual ($\Delta$) learning consistently outperforms direct colour
prediction, particularly in multi-distortion scenarios.
Patch context improves blur correction but offers diminishing
returns beyond $P{=}3$; $P{=}5$ adds cost without meaningful
quality gain.

\subsection{Implementing Spatial Correction as a LUT}
Once the geometry model is trained, it effectively implements a mapping function $(x,y) \rightarrow (x',y')$. 
This allows us to implement any of them as a lookup table (LUT) which is indexed by $(x,y)$ and where each entry contains two floating point values. For a $256\times 256$ output frame, a 64K-entry LUT is sufficient. Each entry contains two FP32 values, resulting in a 512~KB table. This eliminates geometric MLP computation by replacing the network evaluation with a simple table lookup. Lines 1-7 of Algorithm~\ref{alg:seq5prime} implement this optimization where the lookup table is pre-loaded from precomputed memory.

Implementing functions using a LUT is a standard, widely used implementation method. Our contribution here is the method that \emph{learns} how to populate the LUT from paired data. This in contrast to typical LUT deployments that use analytic LUT construction. This approach requires a closed-form inverse of the full geometric transformation, which is expected to be unavailable for the unknown or coupled geometric--photometric distortions we target (e.g., Scenario~5). The learned formulation fills the table directly from data whether or not an analytic inverse exists. Monolithic img2img baselines such as CompenNet++ cannot be reduced to a coordinate LUT because their output depends on image content.
\subsection{Sequential Composition for Spatiotonal Correction}

When geometric and tonal distortions are inseparable, we use a sequence of models.
A composition of $k$ models inherits the pixel-independence of its
components, so LUT implementation and parallelization remain applicable to
each stage.
For example, \textbf{Sequential-3}
($\text{G-Res}{\to}\text{C-Res}{\to}\text{Patch2Pixel-v2}$) first
corrects geometry, then applies a global tonal residual, and
finally refines neighbourhood-dependent artefacts.
{We empirically find that the ordering of stages has
negligible impact on quality (e.g.,\ Sequential-3 vs.\ Sequential-4
reversed have an SSIM difference ${\approx}0.003$), suggesting the pipeline
is modular and robust to reordering.}
Appendix~\ref{sec:supp-seqdefs} (Table~\ref{tab:seq_models}) lists
combinations of geometric and tonal distortion correction models we considered.

\section{Results and Analysis}
\label{sec:results}

We first consider single frame pre-distortion and evaluate single-distortion compensation first, then
multi-distortion scenarios that simulate real XR pipelines.
 Unless otherwise noted, the models use $h=64$.

\subsection{Methodology}
\label{sec:methodology}
\noindent\textbf{Dataset.}
\textbf{Images.cv} (26k images, $3{\times}256{\times}256$, two
categories) is our primary benchmark with a
70/15/15\% train/val/test split.
\textbf{CelebA-HQ 256}~\cite{karras2017progressive,liu2015deep}
(28k faces) provides a zero-shot cross-dataset generalization test; models are evaluated on it without any fine-tuning of the Images.cv-trained weights.

\noindent\textbf{Hyperparameters.}
To train our models we use AdamW~\cite{loshchilov2017decoupled} with cosine annealing and
warm restarts~\cite{loshchilov2016sgdr}; AMP (automatic mixed precision)~\cite{micikevicius2017mixed} enabled.
On Images.cv at $256{\times}256$: batch~$32$, lr~$10^{-3}$,
min-lr~$10^{-6}$, weight decay~$10^{-4}$, warmup~$5$ epochs,
$100$ total epochs.

\noindent\textbf{Evaluation Metrics.}
\label{sec:metrics}
{We report SSIM (structural similarity index)~\cite{wang2002universal} and PSNR (peak signal-to-noise ratio) for perceptual
and signal quality (higher is better), LPIPS (learned perceptual image patch similarity)~\cite{zhang2018unreasonable} and RMSE (root mean square error) for
perceptual and pixel error (lower is better), per-frame MAC (multiply-accumulate) count (one MAC counted as two FLOPs (floating-point operations)) for inference cost, and model parameter count for static memory footprint.}
{All FLOPs are computed for a $3{\times}256{\times}256$
input and include the MLPs, LUTs, and the
fixed-function bilinear interpolation.}

\subsection{Single Distortion Compensation}
\label{sec:single}

We evaluate SSIM and PSNR for the best proposed model against UNet++-Large and CompenNet++ across all single-distortion types; the full per-model, per-distortion breakdown is in Appendix~\ref{sec:supp-single} (Tables~\ref{tab:geo_single}--\ref{tab:nongeo3}). In the interest of space, we summarize the key takeaways for the single distortion scenarios, and present detailed measurements only for the multi-distortion experiments.

\subsubsection{Geometric distortions}
\textbf{G-Res} outperforms UNet++-Large on all geometric distortions (full per-distortion measurements in Appendix~\ref{sec:supp-single}, Table~\ref{tab:geo_single}).
For example, for Fisheye it achieves 40.6\,dB / SSIM\,0.993 vs.\ 30.3\,dB /
0.923 for UNet++-Large (a $\approx$35\% PSNR gain) with
99.93\% fewer parameters (4.5K vs.\ 6.8M) and 95.2\% fewer FLOPs
(581\,MFLOPs vs.\ 12.1\,GFLOPs).
G-Res processes $(x,y)$ through a 3-layer MLP (8,862 ops/pixel,
580.78\,MFLOPs total, 135.27\,MB memory transfers) and is
on par with CompenNet++ while enabling LUT caching reducing
geometric inference to $O(1)$ per pixel (a 99.99\% compute
reduction).

\subsubsection{Non-geometric (tonal \& spatiotonal) distortions}
\textbf{C-Res} and \textbf{{Patch2Pixel-v2}} ($P{=}3$) match or exceed
UNet++-Large across most tonal distortions (for example, WGNU and Color Balance; full per-distortion measurements in Appendix~\ref{sec:supp-single}, Tables~\ref{tab:nongeo1}--\ref{tab:nongeo3}), at only 616--967\,MFLOPs (vs.\ 12,129\,MFLOPs) and
142--169\,MB transfers (vs.\ 1,392\,MB).
{One application-critical exception is Color Fringes
(chromatic aberration), where our best model ($0.916$ / $29.5$\,dB)
trails both UNet++-Large ($0.941$ / $32.3$\,dB) and CompenNet++
($0.938$ / $31.9$\,dB). Because chromatic dispersion is the dominant
artifact in diffractive waveguides, we report this as a known
limitation: the per-pixel tonal MLP lacks the multi-scale spatial
support a dispersion correction requires, and a wider-context tonal
stage is left to future work.}
C-Res processes 9,398 ops/pixel via its $(x,y,R,G,B)$ input;
{Patch2Pixel-v2} ($P{=}3$) uses 14,758 ops/pixel, providing the
neighbourhood context needed for blur-like artefacts.
\textbf{Negative result:} Patch2Patch variants (patch-to-patch
output) underperform single-pixel delta models on most distortions
and are excluded from Sequential pipelines.

\subsection{Multiple Distortion Scenarios}
\label{sec:multi}

We next evaluate systems with multiple simultaneous distortions. We define five representative scenarios (referred to as ``Scenario \#'' or ``Sc.\ \#''): Sc.\ 1--2 are non-geometric, Sc.\ 3--4 are geometric, and Sc.\ 5 is the full spatiotonal mix (ten coupled distortions). The exact distortion composition of each scenario is given in Appendix~\ref{sec:supp-scenarios} (Table~\ref{tab:scenarios}).

\subsubsection{Non-geometric mixed (Scenarios 1--2)}
Table~\ref{tab:mixed_nongeo} shows results for two tonal-only
multi-distortion scenarios.
\textbf{{Patch2Pixel-v2}} ($P{=}3$) is the best single-stage model
(SSIM\,0.857 / 27.4\,dB, Scenario\,1), matching UNet++-Medium and
approaching UNet++-Large / CompenNet++ (SSIM\,0.894 / 29.4\,dB).

\subsubsection{Geometric mixed (Scenarios 3--4)}
Table~\ref{tab:mixed_geo} shows results for compound geometric
distortions.
\textbf{G-Res} maintains near-perfect correction (SSIM\,$\geq$0.975)
on both scenarios, matching CompenNet++ (0.991 / 0.983) while
using $>$$100{\times}$ fewer parameters.

\subsubsection{Spatiotonal-Mixed (Scenario~5)}
Table~\ref{tab:mixed_scenario5} presents our main result. Sequential compositions of pixel-to-pixel models consistently outperform monolithic baselines. {\textbf{Sequential-5$'$} with G-Res ($h{=}64$) and Patch2Pixel-v3 ($p{=}3$, $h{=}64$) achieves SSIM\,0.944 / PSNR\,27.5\,dB, matching CompenNet++ (0.944 / 27.7\,dB) at $\mathbf{106{\times}}$ fewer parameters and $\mathbf{6{\times}}$ fewer FLOPs. The lighter Sequential-5$'$ ($h{=}64,48$) reaches SSIM\,0.943 / PSNR\,27.4\,dB with negligible quality loss, and the smallest ($h{=}32$) still attains SSIM\,0.940 / PSNR\,27.2\,dB.} Ablations Sequential-9/10 (omitting the Patch-based tonal stage) confirm that the patch-based tonal stage is important: SSIM drops to 0.896.

{Across all Sequential compositions and hidden
dimensions, the ordering of stages has negligible effect on quality
(e.g.,\ Sequential-3 vs.\ Sequential-4 reversed: SSIM difference
${\approx}0.003$), confirming the pipeline's modularity.}

{\subsubsection{Conclusion} For best quality we
recommend \textbf{Sequential-5$'$} with G-Res ($h{=}64$) and
Patch2Pixel-v3 ($h{=}64$): $10.8$K parameters, $1{,}407$\,MFLOPs, SSIM~$0.944$ / PSNR~$27.5$\,dB matching CompenNet++ at $\mathbf{6{\times}}$ lower FLOPs and two orders of magnitude fewer parameters. The lighter Patch2Pixel-v3 at $h{=}48$ ($8.5$K parameters, the configuration used in the
Section~\ref{sec:video-extension} ; Section~\ref{sec:hardware} studies $h{=}32$ and $h{=}64$) trades $0.001$ SSIM / $0.1$\,dB PSNR for a noticeable reduction in compute, making it the balanced choice for embedded deployment. The smallest
configuration tested, Patch2Pixel-v3 at $h{=}32$ ($6.7$K
parameters), still reaches SSIM~$0.940$ / PSNR~$27.2$\,dB, useful where the parameter budget is tighter than the quality requirement.}

\begin{table}[t]
\centering\tablestyle
\begin{minipage}[t]{0.48\linewidth}
\centering
\caption{Non-geometric mixed distortions.}
\label{tab:mixed_nongeo}
\setlength{\tabcolsep}{3pt}
\begin{tabular}{lcccc}
\toprule
& \multicolumn{2}{c}{Sc.\,1} & \multicolumn{2}{c}{Sc.\,2} \\
Model & S↑ & P↑ & S↑ & P↑ \\
\midrule
Pure XR            & .416 & 8.3  & .655 & 17.3 \\
C-Res              & .691 & 22.3 & .735 & 21.4 \\
{Patch2Pixel-v2} ($p{=}3$)  & .857 & 27.4 & .863 & 23.0 \\
{Patch2Pixel-v2} ($p{=}5$) & .844 & 27.3 & .873 & 23.3 \\
UNet++-Small       & .507 & 8.6  & .617 & 19.3 \\
UNet++-Medium      & .835 & 27.6 & .840 & 22.9 \\
UNet++-Large       & .894 & 29.4 & .888 & 23.5 \\
CompenNet++        & .894 & 29.4 & .870 & 23.3 \\
\bottomrule
\end{tabular}
\\[2pt]
{\footnotesize Sc.\,1: color balance, Br\&Ct, elevated black, vignetting, blur, color fringes. Sc.\,2: elevated black, brightness±, contrast, blur, color fringes, vignetting, WGNU.\par}
\end{minipage}\hfill
\begin{minipage}[t]{0.48\linewidth}
\centering
\caption{Geometric mixed distortions.}
\label{tab:mixed_geo}
\setlength{\tabcolsep}{3pt}
\begin{tabular}{lcccc}
\toprule
& \multicolumn{2}{c}{Sc.\,3} & \multicolumn{2}{c}{Sc.\,4} \\
Model & S↑ & P↑ & S↑ & P↑ \\
\midrule
Pure XR        & .704 & 15.3 & .650 & 14.9 \\
G-Hybrid h=16  & .982 & 35.8 & .968 & 32.9 \\
G-Hybrid       & .987 & 37.4 & .974 & 34.8 \\
\textbf{G-Res} & \textbf{.987} & \textbf{37.5} & \textbf{.975} & \textbf{34.9} \\
UNet++-Small   & .857 & 24.1 & .834 & 23.7 \\
UNet++-Medium  & .877 & 27.1 & .858 & 26.9 \\
UNet++-Large   & .914 & 29.2 & .895 & 28.7 \\
CompenNet++    & .991 & 39.1 & .983 & 36.3 \\
\bottomrule
\end{tabular}
\\[2pt]
{\footnotesize Sc.\,3: keystone + fisheye. Sc.\,4: keystone + barrel + fisheye.\\
G-Res leads all UNet++ variants by ${\approx}8$\,dB PSNR and comes within ${\approx}1.5$\,dB of CompenNet++ at ${\approx}250{\times}$ fewer parameters (Table~\ref{tab:model_summary}).\par}
\end{minipage}
\end{table}

\subsection{Formula-Agnostic, Real-Data, and Non-Uniform Validation}
The training procedure above back-propagates through a known differentiable
distortion pipeline, but Spatons do not require one. Using the two-phase scheme
of Sec.~\ref{sec:twophase} (fit a forward-mimic $\mathrm{NN}_1$ from image pairs,
then train the deployable $\mathrm{NN}_2$ through the frozen mimic), we validate
predistortion in three settings, detailed in
Appendix~\ref{sec:supp-validation}. (i)~On the full spatiotonal Scenario~5, the
$8.5$K-parameter $\mathrm{NN}_2$ trained through a frozen CompenNet++ mimic
reaches $26.59$\,dB / $0.926$~SSIM which is within $0.9$\,dB of the known-formula upper
bound and performs comparably through a tiny Sequential-5$'$ mimic, showing the
deployable model does not depend on a hand-derived differentiable pipeline.
(ii)~On real projector--camera captures~\cite{huang2019compennet++}
(\texttt{lavender\_np}, $700$ pairs), the same network raises the uncorrected
baseline ($\approx\!10.3$\,dB) to $22.23$\,dB / $0.816$~SSIM, a large, stable
correction from a network two to three orders of magnitude smaller than the mimic.
(iii)~Under a spatially non-uniform four-regime operator
(keystone/fisheye $\times$ blur/contrast), $\mathrm{NN}_2$ recovers a $13$\,dB
baseline to $25.22$\,dB / $0.915$~SSIM in formula mode and up to
$28.86$\,dB / $0.941$~SSIM through a frozen learned mimic. Across all three,
achievable predistortion is bounded by the \emph{invertibility} of the forward
model rather than its raw fidelity, a smoother learned operator can be easier for
a tiny corrector to pre-invert than the exact formulas.

\subsection{Higher Resolution Training and Zero-Shot Cross-Dataset Generalization}
\label{sec:generalization}

We evaluate two distinct settings. For the FFHQ image experiments, models are trained directly on FFHQ at $1024\times1024$ and evaluated on Scenarios~2 and~4 to assess higher-resolution performance. Separately, Images.cv-trained models are evaluated zero-shot on CelebA-HQ-256 without fine-tuning for the full spatiotonal Scenario~5. On CelebA-HQ-256, Sequential-5$'$ and Sequential-7$'$ match CompenNet++ ($0.948$ / $29.1$\,dB and $0.949$ / $29.2$\,dB versus $0.948$ / $29.3$\,dB). Full results are reported in Appendix~\ref{sec:supp-generalization} (Tables~\ref{tab:ffhq_all_scenarios} and~\ref{tab:celeba_generalization}).
\begin{table}[htbp]
\centering\tablestyle
\caption{Mixed geometric + non-geometric distortions, Scenario~5 (↑\,higher better, ↓\,lower better). Full $h$-sweep, all Sequential compositions, and the complete baseline ladder are in Table~\ref{tab:supp-scenario5full}; Sequential~9/10 ablate the patch-tonal stage.}
\label{tab:mixed_scenario5}
\fittable{%
\begin{tabular}{llcccc}
\toprule
Model & $h$ & SSIM↑ & PSNR↑ & LPIPS↓ & RMSE↓ \\
\midrule
Pure XR & -- & .641 & 14.6 & .635 & .186 \\
\midrule
\textbf{{Sequential-5$'$} ($p=3$)} & \textbf{64,64} & \textbf{.944} & \textbf{27.5} & \textbf{.082} & \textbf{.043} \\
{Sequential-5$'$} ($p=3$) & 64,48 & .943 & 27.4 & .085 & .043 \\
{Sequential-5$'$} ($p=3$) & 32,32 & .939 & 27.1 & .091 & .044 \\
\textbf{{Sequential-7$'$} ($p=3$)} & \textbf{64,64,64} & \textbf{.944} & \textbf{27.6} & \textbf{.083} & \textbf{.042} \\
{Sequential-9/10} (no patch-tonal) & 64,64 & .896 & 25.8 & .172 & .051 \\
\midrule
UNet++-Large   & --  & .868 & 25.8 & .058 & .052 \\
\textbf{CompenNet++} & -- & \textbf{.944} & \textbf{27.7} & \textbf{.033} & \textbf{.041} \\
\bottomrule
\end{tabular}}
\newline
{\footnotesize \\ {Pipeline compositions defined in Table~\ref{tab:seq_models}. Full pipeline: projector colour artefacts (elevated black, Br±v2, contrast, blur, vignetting) → keystone → glass artefacts (color fringes, WGNU-v2) → barrel + fisheye.}}
\end{table}

We note that at matched SSIM, Sequential-5$'$ has higher
(worse) LPIPS than CompenNet++ ($0.082$ vs.\ $0.033$,
Table~\ref{tab:mixed_scenario5}): the per-pixel tonal stage matches structural
and signal fidelity but does not reproduce the deep-feature statistics a
convolutional backbone captures. For a predistortion operator judged primarily
on geometric and photometric accuracy under a strict compute budget we treat
SSIM/PSNR as the operative metrics, and report the LPIPS gap as a known
trade-off of the lightweight design.


\subsection{Quantization}
\label{sec:quant}

Since energy and implementation costs are primary concerns in XR, we consider quantization as it can reduce computation and memory costs. Here we study its effect on output quality. We quantize with symmetric, whole-tensor scaling with FP32 scale and bias. We did not study per-group quantization as this would require additional support in hardware and software. 
We sweep weight bits from 32 down to 4 to find the minimum precision $N_w^*$ within ${\approx}0.1$\,dB PSNR and $0.01$ SSIM of the FP32 baseline, then fix weights and sweep activation bits over the same 32--4 range for $N_a^*$ (same criterion).
Quantization is studied here purely for its effect on output quality; the resource and latency estimates of Sec.~\ref{sec:hardware} assume none of these reduced precisions.

Spatons tolerate low-bit fixed-point quantization with negligible quality loss; per-model weight/activation bits are reported in Appendix~\ref{sec:supp-quant} (Table~\ref{tab:quant}).


\subsection{Computation and Memory Efficiency Summary}
\label{sec:efficiency}

Table~\ref{tab:model_summary} provides a concise efficiency
comparison.
Our single-stage models are $13$--$21{\times}$ cheaper in FLOPs and
$6$--$10{\times}$ lower in memory transfers than UNet++-Large.
Sequential pipelines add minimal overhead; the recommended
Sequential-5$'$ ($h{=}32,32$) requires only \textbf{443\,MFLOPs} and
\textbf{169\,MB} {total memory transfers} and over $19{\times}$ less compute
than CompenNet++. In the rest of this work we focus on Sequential-5$'$.


\begin{table}[t]
\centering\tablestyle
\caption{Model efficiency summary ($3{\times}256{\times}256$ input). Full per-model breakdown in Appendix~\ref{sec:supp-costs} (Table~\ref{tab:supp_model_summary}).}
\label{tab:model_summary}
\setlength{\tabcolsep}{3pt}
{%
\begin{tabular}{lrrrr}
\toprule
Model & Params & FLOPs (M) & Transfer (MB) & SSIM Sc5 \\
\midrule
CompenNet++     & 1.15M & 8,526 & 162  & .944 \\
UNet++-Large    & 6.83M & 12,129 & 1392 & .868 \\
\midrule
\textbf{Sequential-5$'$} ($h{=}32,32$) & \textbf{3.4K} & \textbf{443} & \textbf{169} & .939 \\
Sequential-5$'$ ($h{=}64,64$) & 10.8K & 1{,}407 & 297 & .944 \\
\bottomrule
\end{tabular}
}
\end{table}

\section{Video Pre-distortion}
\label{sec:video-extension}

The preceding sections evaluated Spatons as image-level predistortion operators. AR displays, however, process continuous frame streams. Video introduces temporal coherence: many pixels remain unchanged across adjacent frames. We evaluate whether this can avoid full frame model evaluation by invoking Spatons only for output pixels that may change given the differences across adjacent frames. 
We exploit this with an inter-frame reverse-cache (\texttt{lutrevcache}) that recomputes only outputs whose source pixels changed; its CSR construction, fan-in statistics, and hardware mapping are in Appendix~\ref{sec:supp-revlut}.


We benchmark the Sequential-5$'$ video pipeline across five targets
(RTX~3090/Ryzen~5900X, Jetson Orin Nano Super CPU and iGPU, Raspberry~Pi~5)
and two datasets (YouTube-UGC~\cite{wang2019youtubeugc},
UVG~\cite{mercat2020uvg}), sweeping all nine G-Res$\times$Patch2Pixel-v3
operation modes (Table~\ref{tbl:modes}; Appendix~\ref{sec:supp-video}). For these video experiments, we use the Images.cv-trained model weights without video-specific training or fine-tuning. Three findings hold across
platforms. First, because G-Res depends only on coordinates, replacing its MLP
with a LUT preserves quality (within $0.01$\,dB PSNR / $10^{-4}$ SSIM) while
giving a $13$--$104{\times}$ speedup on CPUs (e.g.,\ $574{\to}5.5$\,ms at
UVG-1080 on Ryzen) and $2$--$6{\times}$ on GPUs. Second, once G-Res is a lookup,
Patch2Pixel-v3 dominates the workload ($95$--$99\%$ of model compute on CPUs,
$47$--$57\%$ on the RTX~3090), so video predistortion reduces to one
content-dependent per-pixel MLP evaluation per output pixel. Third, and
contrary to the usual motivation for temporal caching, the inter-frame reverse
cache (\texttt{lutrevcache}) helps only in a narrow regime (x86 CPU at $256^2$
with high coherence) and otherwise regresses, adding $124$--$515\%$ G-Res
latency at $1080^2$ because the $4.66$M-edge CSR walk outweighs the reuse
savings; our LUT optimization therefore renders inter-frame reuse ineffective in
practice. Quality is numerically portable: within each (dataset, resolution)
cell all five targets agree to $\approx$$0.01$\,dB / $10^{-4}$
(Table~\ref{tab:video_xplatform} reports the fastest configuration per target).

\section{Analytical Resource and Latency Model}
\label{sec:hardware}

This section estimates the throughput and latency of a hardware implementation of Spatons over several target device platforms. 
Unlike the \emph{measured} results of Sec.~\ref{sec:video-extension} (real C++/OpenMP/CUDA executions), this section presents a first-order \emph{analytical} model that estimates throughput and latency from per-frame arithmetic work and published device specifications; a synthesized hardware implementation is left to future work. 
This section and Appendix~\ref{sec:supp-hardware} assume FP32 storage and arithmetic throughout for weights, activations, the coordinate lookup table, and all frame buffers and do not apply the reduced precisions of Sec.~\ref{sec:quant}: quantization is an orthogonal optimization, and an FP32 baseline keeps these estimates independent of it.

We model a hardware implementation of Sequential-5$'$. Stage~1 (G-Res) is memory-bound: after LUT substitution, the coordinate table occupies 512~KB and the remaining work is only 1.97~MFLOPs of fixed-function bilinear interpolation. Stage~2 (Patch2Pixel) is compute-bound: $N$ parallel MLP instances process disjoint pixel partitions, allowing throughput to scale with $N$ until device resources are exhausted. Full dataflow (Fig.~\ref{fig:data_flow}), per-stage resource accounting (Table~\ref{tab:resource_per_stage}), the FPGA processing-element mapping, and the latency derivation are given in Appendix~\ref{sec:supp-hardware}.


Peak throughput is estimated from total per-frame arithmetic work, $\mathrm{FPS}_{\mathrm{peak}} = P_{\mathrm{peak}} / (\mathrm{FLOPs}_{\mathrm{Stage1}} + \mathrm{FLOPs}_{\mathrm{Stage2}})$, giving 287.6~MFLOPs ($h{=}32$) and 828.8~MFLOPs ($h{=}64$) per frame after LUT substitution.

\begin{table}[t]
\centering\tablestyle
{\caption{Peak FPS (throughput model above), assuming 16-PE MLP instances with 1~MAC/cycle/PE. Dashes indicate configurations exceeding device resources.}}

\label{tab:fps_scaling}
\fittable{%
\begin{tabular}{c l r r r r r r}
\toprule
$h$ & Device (Freq) & $N=1$ & $N=2$ & $N=4$ & $N=8$ & $N=16$ & $N=32$ \\
\midrule
\multirow{2}{*}{32}
& Lattice CrossLink-NX (200 MHz)~\cite{Lattice_CLNX_DS} & 22.25 & 44.51 & -- & -- & -- & -- \\
& AMD Zynq ZU3EG (600 MHz)~\cite{AMD_DS891}                 & 66.76 & 133.52 & 267.04 & 534.08 & 1068.15 & -- \\
\midrule
\multirow{2}{*}{64}
& Lattice CrossLink-NX (200 MHz)~\cite{Lattice_CLNX_DS} & 7.72 & 15.44 & -- & -- & -- & -- \\
& AMD Zynq ZU3EG (600 MHz)~\cite{AMD_DS891}                 & 23.16 & 46.33 & 92.66 & 185.32 & 370.64 & -- \\
\bottomrule
\end{tabular}%
}

\end{table}


Latency is the end-to-end delay from the first input pixel entering Stage~1 to the last output pixel leaving Stage~2. Unlike throughput, it is directly tied to XR comfort: latencies above 20~ms can become perceptible and contribute to cybersickness, while sub-20~ms is widely used as a comfort target~\cite{Stauffert2020,Jerald2009}.

Using the latency model of Appendix~\ref{sec:supp-hardware}, the Zynq ZU3EG at $N{=}16$ and 600~MHz achieves $T_{\mathrm{latency}} \approx 1.01\text{--}1.15$~ms, while low-power CrossLink-NX at $N{=}2$ and 200~MHz achieves $\approx 21.38\text{--}21.79$~ms.
The Zynq ZU3EG thus meets XR latency targets at 1{,}068~FPS ($h{=}32,N{=}16$); the CrossLink-NX (44~FPS, $N{=}2$) is an attractive low-power point despite latency slightly above the 20~ms threshold. The pixel-independent Spaton pipeline therefore spans high-throughput FPGA and very low-power embedded accelerators.

\section{Conclusion}
\label{sec:conclusion}
This work characterizes Spatons as lightweight building blocks for XR predistortion under controlled, paired-data conditions, using closed-form distortion generators—including constructed spatiotonal couplings—to study correction quality, model cost, and deployment feasibility in a model-matched setting. Spatons efficiently capture spatial, tonal, and coupled distortions while remaining compatible with practical XR constraints. We do not claim through-the-optics validation or perceptual adequacy under real device variation, sensor noise, calibration error, or temporal drift; rather, this establishes a systems-level foundation and workload characterization for efficient learnable predistortion, motivating future validation on measured optical pipelines and perceptual metrics.

We also note that the formula-agnostic two-phase scheme is bounded by the fidelity of its Phase~1 forward mimic: when the mimic cannot fully reproduce the true distortion system, the Phase~2 corrector inherits that error (a ${\approx}0.9$\,dB gap to known-formula training on Scenario~5). A fuller discussion of limitations and future work is given in Appendix~\ref{sec:supp-limitations}.



\subsubsection*{Broader Impact Statement}
This work targets efficient, low-latency distortion correction for XR displays at
substantially lower compute and memory cost than monolithic baselines, which is
expected to lower energy use on untethered devices and improve viewing comfort;
the method is application-specific with little potential for misuse.
The same pre-compensation principle has been applied to correcting a
viewer's own refractive errors~\citep{huang2014eyeglasses}, where an efficient
per-pixel corrector like ours may also be applicable.

\bibliography{main}
\bibliographystyle{tmlr}

\appendix
\section*{Appendix}

\section{Baseline Training Details}
UNet++ (all variants) and CompenNet++ were trained and evaluated under
\emph{identical} conditions to our models: the same loss (Eq.~\ref{eq:loss},
MSE $+\,0.5 L_1 +\,0.3$\,LPIPS), the same $70/15/15$ Images.cv split at
$256{\times}256$, the same expected-image/ROI target $E$ (Sec.~\ref{sec:problem}),
the same distortion pipeline and scenarios, the same AdamW optimizer with
cosine annealing and warm restarts and AMP, the same budget ($100$ epochs,
batch $32$, lr $10^{-3}$, min-lr $10^{-6}$, weight decay $10^{-4}$, $5$ warmup
epochs), and identical validation-based checkpoint selection.

UNet++-Large's weaker geometric performance (Table~\ref{tab:mixed_geo})
is architectural, not a training artifact: a dense img2img network must
synthesize spatially displaced content through its feature maps, whereas
large geometric displacements are more naturally handled by coordinate
resampling---the motivation for our separate geometric stage. Consistent
with this, UNet++-Large is competitive on tonal scenarios
(Appendix~\ref{sec:supp-single}, Table~\ref{tab:mixed_nongeo}) while CompenNet++,
which contains a dedicated WarpingNet, performs well on the geometric
scenarios ($0.991/0.983$).

\section{Distortion Formulas}
\label{sec:supp-distortions}

\begin{table}[H]
\centering\tablestyle\tablestyle
\begingroup
\caption{AR distortions.}
\label{tab:distortions}
\footnotesize
\setlength{\tabcolsep}{3pt}
\renewcommand{\arraystretch}{1.08}

\begin{tabularx}{\columnwidth}{@{}>{\raggedright\arraybackslash}p{0.31\columnwidth}
                                >{\raggedright\arraybackslash}X@{}}
\toprule
\textbf{Distortion} & \textbf{Formula} \\
\midrule
\multicolumn{2}{@{}l}{\textit{\textbf{Spatial}}} \\
\midrule
Keystone    & Perspective transform. \\
Fisheye     & $\vec{x}'{=}\vec{x}/(1{-}ar^2)$ \\
Barrel/Pincushion & $\vec{x}'{=}\vec{x}(1{\pm}ar^2{\pm}br^4)$;\; $(+)$=Barrel, $(-)$=Pincushion \\
Mustache    & $\vec{x}'{=}\vec{x}(1{-}ar^2{+}br^4)$ \\

\midrule
\multicolumn{2}{@{}l}{\textit{\textbf{Tonal}}} \\
\midrule
Brightness-v1   & $g = f + b$ \\
Contrast-v1     & $g = (f-0.5)\cdot a+0.5$ \\
Br. \& Contrast & $g = (f\cdot a-0.5)\cdot a+0.5$ \\
Elevated Black  & $g = (f+p)/(1+p)$ \\
Color Balance   & $g_c = f_c \cdot a_c$ \\

\midrule
\multicolumn{2}{@{}l}{\textit{\textbf{Spatiotonal}}} \\
\midrule
Gaussian Blur   & $g(\vec{x}) = (f*G_\sigma)(\vec{x})$ \\
Vignetting      & $g(\vec{x}) = f(\vec{x})\cdot\mathrm{clamp}(1{-}ar^2,\,0.7,\,1)$ \\
WGNU-v1         & $g(\vec{x}) = f(\vec{x})\cdot[1{+}a\sin(2\pi\nu x{+}\phi)]$ \\
WGNU-v2         & $g_c(\vec{x}) = f_c(\vec{x})\cdot[1{+}a\cdot B(\vec{x})\cdot(1{+}b(c{-}1))]$ \\
WGNU-v3         & Same as v2;\; $a{=}a(\mu_f,\sigma_f)$ \\
Brightness-v2   & $g(\vec{x}) = f(\vec{x}) + a\cdot\mathrm{pat}(\vec{x})$ \\
Brightness-v3   & Same as v2;\; $f_x,\phi_x{=}h(\mu_f)$;\; $f_y,\phi_y{=}h(\sigma_f)$ \\
Contrast-v2     & $g(\vec{x}) = (f(\vec{x}){-}0.5)\cdot\mathrm{clamp}(a\cdot\mathrm{pat}(\vec{x}),0.2,2)+0.5$ \\
Contrast-v3     & Same as v2;\; $f_x,\phi_x{=}h(\mu_f)$;\; $f_y,\phi_y{=}h(\sigma_f)$ \\
Color Fringes   & $g_R{=}f_R\cdot a_R$;\; $g_G{=}f_G$;\; $g_B{=}f_B\cdot a_B$ \\

\midrule
\multicolumn{2}{@{}p{\dimexpr\columnwidth\relax}@{}}{
\textit{Notation.}
$\vec{x}{=}(x,y)$ are normalized spatial coordinates;
$r^2{=}x^2{+}y^2$;
$f(\vec{x})$ and $g(\vec{x})$ are input and output pixel values;
$c{\in}\{0,1,2\}$ indexes RGB channels;
$a,b,p{>}0$ are scalar distortion parameters;
$\mu_f,\sigma_f$ are the per-image mean and standard deviation of $f$;
$\nu,\phi$ are spatial frequency and phase;
$\sigma$ is the Gaussian standard deviation and $G_\sigma$ its kernel;
$B(\vec{x})$ is a multi-frequency spatial blend pattern;
$\mathrm{pat}(\vec{x}){=}\frac{1}{2}[\sin(2\pi f_x x{+}\phi_x){+}\sin(2\pi f_y y{+}\phi_y)]$
is a fixed-frequency spatial pattern; and
$h(\cdot)$ maps image statistics to frequency and phase.
} \\
\bottomrule
\end{tabularx}

\endgroup
\end{table}

\section{Sequential Model Definitions}
\label{sec:supp-seqdefs}

\begin{table}[H]
\centering\tablestyle\tablestyle
\caption{Sequential model definitions.}
\label{tab:seq_models}
\fittable
{%
\begin{tabular}{ll}
\toprule
\textbf{Model} & \textbf{Pipeline Definition} \\
\midrule
{Sequential-1}      & G-Res $\to$ Patch2Pixel-v2 ($p{=}5$) \\
{Sequential-2}     & Patch2Pixel-v2 ($p{=}5$) $\to$ G-Res \\
{Sequential-3}      & G-Res $\to$ C-Res $\to$ Patch2Pixel-v2 ($p{=}5$) \\
{Sequential-4}      & C-Res $\to$ Patch2Pixel-v2 ($p{=}5$) $\to$ G-Res \\
{Sequential-5}      & G-Res $\to$ Patch2Pixel-v2 ($p{=}3$) \\
{Sequential-5$'$}   & G-Res $\to$ Patch2Pixel-v3 ($p{=}3$) \\
{Sequential-5$''$}  & G-Res $\to$ Patch2Patch ($p_\text{in}{=}5,\,p_\text{out}{=}3$) \\
{Sequential-6}      & Patch2Pixel-v2 ($p{=}3$) $\to$ G-Res \\
{Sequential-7}      & G-Res $\to$ C-Res $\to$ Patch2Pixel-v2 ($p{=}3$) \\
{Sequential-7$'$}   & G-Res $\to$ C-Res $\to$ Patch2Pixel-v3 ($p{=}3$) \\
{Sequential-7$''$}  & G-Res $\to$ C-Res $\to$ Patch2Patch ($p_\text{in}{=}5,\,p_\text{out}{=}3$) \\
{Sequential-8}      & C-Res $\to$ Patch2Pixel-v2 ($p{=}3$) $\to$ G-Res \\
{Sequential-9}      & G-Res $\to$ C-Res \\
{Sequential-10}     & C-Res $\to$ G-Res \\
\bottomrule
\end{tabular}}
\end{table}

\section{Evaluation Scenario Definitions}
\label{sec:supp-scenarios}

\begin{table}[H]
\centering\tablestyle
\caption{Summary of evaluation scenarios.}
\label{tab:scenarios}
\fittable{%
\begin{tabular}{cl p{5cm}}
\toprule
\textbf{Sc.} & \textbf{Type} & \textbf{Distortions} \\
\midrule
1 & Non-Geometric    & Color balance, Brightness (Br)\&Contrast (Ct), elev.\ black, vignetting, blur, color fringes \\
\addlinespace
2 & Non-Geometric    & Elev.\ black, Br$\pm$-v2, contrast-v2, blur, color fringes, vignetting, WGNU-v2 \\
\addlinespace
3 & Geometric & Keystone, fisheye \\
\addlinespace
4 & Geometric & Keystone, barrel, fisheye \\
\addlinespace
5 & Spatiotonal & Elev.\ black, Br$\pm$-v2, contrast-v2, blur, vignetting, keystone, color fringes, WGNU-v2, barrel, fisheye \\
\bottomrule
\end{tabular}}
\end{table}
%
%
\section{Additional Architecture Details}
\label{sec:supp-archdetail}

\subsection*{Geometric models}
\begin{itemize}
  \item \textbf{G-Res}: a tiny 3-layer residual MLP; maps $(x,y)\!\to\!(x',y')$ followed by pixel interpolation. We consider hidden dimensions $h$ of 64, 32, and 16. For $h=64$ the model has $4482$ parameters in total.
  \item \textbf{G-Hybrid}: augments G-Res with a shallow CNN branch over a $3{\times}3$ coordinate neighbourhood to capture local distortion gradients with marginally higher cost than G-Res.
\end{itemize}

\subsection*{Tonal models}
\begin{itemize}
  \item \textbf{C-Res}: 3-layer MLP with residual skip; inputs $(x,y,R,G,B)$; predicts a normalized $\Delta\text{RGB}\in[-2,2]$ via a sigmoid-scaled output.
  \item \textbf{C-Hybrid}: dual-branch design fusing a single-pixel MLP with a shallow 3-layer CNN branch; provides the strongest per-pixel tonal capacity at higher cost.
  \item \textbf{Patch2Pixel~v1:} a 3-layer residual MLP that ingests a $P{\times}P$ patch in which each of the $P^2$ pixels carries its RGB values and its own $(x,y)$ spatial coordinates, giving $\mathbf{f}\in\mathbb{R}^{5P^2}$. Predicts \emph{absolute} RGB output values directly rather than a residual correction. It is included as an ablation of residual learning. Direct colour prediction consistently underperforms residual ($\Delta$) learning, particularly in multi-distortion scenarios where the correction signal is small relative to the input.
  \item \textbf{{Patch2Pixel-v2:}} shares the same 3-layer residual MLP architecture and $\mathbf{f}\in\mathbb{R}^{5P^2}$ input as v1, but predicts a \textit{normalized} residual $\Delta$RGB $\in[-2,2]^3$ via a sigmoid-scaled output rather than absolute colours. The per-pixel coordinate embedding lets the model resolve spatially-varying tonal artefacts (e.g.\ WGNU) while the neighbourhood context handles blur-like distortions.
  \item \textbf{Patch2Pixel~v3:} a more efficient variant of v2 in which spatial information is reduced to the \emph{center} coordinates only: $\mathbf{f}=[\mathrm{flatten}(\mathbf{W}),\,x,\,y]\in\mathbb{R}^{3P^2+2}$, where $\mathbf{W}$ is the $P{\times}P$ RGB patch. This removes $2(P^2-1)$ inputs relative to v2 with negligible quality loss (see Table~\ref{tab:mixed_scenario5}), making it the preferred tonal stage in the Sequential~5$'$ pipeline.
\end{itemize}

\section{Formula-Agnostic, Real-Data, and Non-Uniform Validation}
\label{sec:supp-validation}

\subsection{Formula-Agnostic Predistortion (Two-Phase)}
\label{sec:twophase-results}
We validate the two-phase scheme of Sec.~\ref{sec:twophase} on the full
spatiotonal Scenario~5. Phase~1 learns the forward operator from
clean/distorted pairs (Table~\ref{tab:phase1}); a high-capacity CompenNet++
mimic reaches $36.94$\,dB / $0.978$ SSIM. Freezing this learned operator and
training only the $8.5$K-parameter $\mathrm{NN}_2$ through it
(Table~\ref{tab:phase2}) yields $26.59$\,dB / $0.926$ SSIM---within $0.9$\,dB
of the known-formula upper bound ($27.4$\,dB / $0.943$, same model). Using the
tiny Sequential-5$'$ as the frozen mimic instead performs comparably
($26.66$\,dB / $0.927$), showing the deployable predistortion model does not
depend on a hand-derived differentiable distortion pipeline.

\begin{table}[t]
\centering\tablestyle
\caption{Two-phase Phase~1: forward-mimic fidelity on Scenario~5 (clean $\to$ NN$_1$ $\to$ distorted).}
\label{tab:phase1}
{\begin{tabular}{lrcccc}
\toprule
NN$_1$ model & Params & PSNR↑ & SSIM↑ & LPIPS↓ & RMSE↓ \\
\midrule
CompenNet++                 & 1{,}152{,}147 & 36.94 & .978 & .008 & .014 \\
Sequential-5$'$ ($h{=}64$)  & 10{,}847      & 28.81 & .948 & .035 & .036 \\
\bottomrule
\end{tabular}}
\end{table}

\begin{table}[t]
\centering\tablestyle
\caption{Two-phase Phase~2 on Scenario~5: predistortion through the frozen NN$_1$ (clean $\to$ NN$_2$ $\to$ NN$_1^{\text{frozen}}$ $\to$ clean). NN$_2$ is Sequential-5$'$, $8.5$K params.}
\label{tab:phase2}
{\begin{tabular}{lccccr}
\toprule
Frozen NN$_1$ & PSNR↑ & SSIM↑ & LPIPS↓ & RMSE↓ & $\Delta$PSNR vs.\ Pure-AR \\
\midrule
CompenNet++      & 26.59 & .926 & .110 & .047 & $+11.95$ \\
Sequential-5$'$  & 26.66 & .927 & .112 & .047 & $+12.02$ \\
\bottomrule
\end{tabular}}
\end{table}

\subsection{Real-Data Validation (Projector--Camera Captures)}
\label{sec:realdata}
We run the full two-phase pipeline end-to-end on physically captured data
from the CompenNet++ projector--camera benchmark~\cite{huang2019compennet++}
(setup \texttt{light1/pos1/lavender\_np}: $700$ real clean/distorted pairs at
$256{\times}256$, $70/15/15$ split). Phase~1 fits a CompenNet++ forward mimic;
the surface-aware variant reaches $23.92$\,dB / $0.827$ SSIM, versus
$21.12$\,dB / $0.742$ for the black-surface variant (Table~\ref{tab:realdata}).
Freezing each mimic and training the $8.5$K-parameter Sequential-5$'$
(G-Res $h{=}64\to$Patch2Pixel-v3 $h{=}48$) $\mathrm{NN}_2$ raises the
uncorrected pure-AR baseline ($\approx\!10.3$\,dB / $0.49$ SSIM) to
$22.23$\,dB / $0.816$ and $20.62$\,dB / $0.781$ respectively---a large,
stable correction on real captures from a network two to three orders of
magnitude smaller than the mimic. Notably, the \emph{weaker} Phase-1 mimic
(black-surface) yields the \emph{stronger} Phase-2 correction: the more
faithful surface-aware mimic reproduces harder-to-invert residual structure,
so forward-mimic fidelity and achievable compensation are distinct objectives.
Because \texttt{lavender}'s distortion is a mild, largely invertible
photometric transform, a tiny predistortion network pre-compensates it almost
fully; on information-lossy operators (e.g., chroma sub-sampling)
$\mathrm{NN}_2$ could not beat the uncorrected baseline, delimiting where
predistortion is effective.

\begin{table}[t]
\centering\tablestyle
\caption{Real projector--camera data (\texttt{lavender\_np}). Top: Phase-1 forward mimics. Bottom: Phase-2 predistortion through each frozen mimic ($\mathrm{NN}_2$ = Sequential-5$'$, $h{=}64/48$, $8{,}511$ params).}
\label{tab:realdata}
\fittable{%
\begin{tabular}{llrcccc}
\toprule
Stage & Model & Params & PSNR↑ & SSIM↑ & LPIPS↓ & RMSE↓ \\
\midrule
\multicolumn{7}{@{}l}{\textit{Phase~1: forward mimic}} \\
P1 & CompenNet++ (black surface) & 1{,}152{,}147 & 21.12 & .742 & .118 & .088 \\
P1 & CompenNet++ (real surface)  & 1{,}152{,}147 & 23.92 & .827 & .066 & .064 \\
\midrule
\multicolumn{7}{@{}l}{\textit{Phase~2: predistortion ($\mathrm{NN}_2$ = Sequential-5$'$, $8{,}511$ params)}} \\
P2 & Pure AR (no $\mathrm{NN}_2$), black  & --      & 10.28 & .489 & --   & --   \\
P2 & Pure AR (no $\mathrm{NN}_2$), real   & --      & 10.25 & .487 & --   & --   \\
P2 & $\mathrm{NN}_2 \to$ frozen black     & 8{,}511 & 22.23 & .816 & .154 & .077 \\
P2 & $\mathrm{NN}_2 \to$ frozen real      & 8{,}511 & 20.62 & .781 & .196 & .093 \\
\bottomrule
\end{tabular}}
\end{table}

\subsection{Spatially Non-Uniform, Coupled Distortion}
\label{sec:nonuniform}
To stress-test predistortion under a spatially heterogeneous operator rather
than a single global distortion, we construct a composite degradation that
varies across the frame along two axes at once: top/bottom applies
keystone vs.\ fisheye warping, and left/right applies Gaussian blur vs.\
contrast loss, yielding four corner regimes blended through soft seams.
Photometric degradation (strength $0.4$) is applied before geometry
(strength $0.2$), keeping the operator differentiable. The predistortion
network $\mathrm{NN}_2$ is a compact Sequential-5$'$ (G-Res $h{=}64\to$
Patch2Pixel-v3 $p{=}3,h{=}48$; $8{,}511$ parameters), with \emph{no}
spatial-conditioning mechanism and no knowledge of the regime boundaries.
The pure-AR baseline is measured through each forward operator separately
(analytic formulas, or the specific frozen mimic), and ranges
$13.01$--$13.09$\,dB / $0.555$--$0.566$ SSIM.

\noindent\textbf{Part 1 --- Known differentiable operator.}
Trained in pure formula mode ($\text{clean}\!\to\!\mathrm{NN}_2\!\to\!\text{AR-formulas}\!\to\!\text{clean}$),
$\mathrm{NN}_2$ recovers the uncorrected Pure-AR baseline of
$13.09$\,dB / $0.566$ SSIM to $25.22$\,dB / $0.915$ SSIM
(LPIPS $0.090$, RMSE $0.055$): $+12.1$\,dB / $+0.348$ SSIM
(Table~\ref{tab:nonuniform}).

\noindent\textbf{Part 2 --- Learned formula-agnostic operator.}
Removing the formula assumption, Phase~1 mimics the same non-uniform
degradation from image pairs alone: CompenNet++ ($1{,}152{,}147$ params)
reaches $35.96$\,dB / $0.970$ SSIM and the ${\approx}100{\times}$ smaller
Sequential-5$'$ ($10{,}847$ params) $23.36$\,dB / $0.872$ SSIM. Freezing each
mimic and training $\mathrm{NN}_2$ through it, predistortion reaches
$24.78$\,dB / $0.890$ SSIM through the frozen CompenNet++ operator
($+11.7$\,dB; within $0.44$\,dB of the Part-1 formula-mode upper bound) and
$28.86$\,dB / $0.941$ SSIM through the frozen Sequential-5$'$ operator
($+15.9$\,dB, exceeding the formula-mode bound by ${\approx}3.6$\,dB). The
latter, initially counterintuitive, result reinforces Sec.~\ref{sec:realdata}:
a smoother learned operator is easier for a tiny corrector to pre-invert than
the exact, harsh four-regime formulas---achievable predistortion is bounded by
the \emph{invertibility} of the forward model, not merely its fidelity.

\begin{table}[t]
\centering\tablestyle
\caption{Spatially non-uniform four-regime degradation
(keystone/fisheye $\times$ blur/contrast). Pure-AR is measured through each
operator (rows grouped). $\mathrm{NN}_2$ is Sequential-5$'$, $8{,}511$ params.
The frozen-Seq-5$'$ Phase-2 value is $28.865$ at $3$\,dp.}
\label{tab:nonuniform}
\fittable{%
\begin{tabular}{llrccccr}
\toprule
Mode & Model & Params & PSNR↑ & SSIM↑ & LPIPS↓ & RMSE↓ & $\Delta$PSNR \\
\midrule
\multicolumn{8}{@{}l}{\textit{Known operator (analytic formulas)}} \\
Pure AR & (uncorrected)      & --      & 13.09 & .566 & --   & --   & --      \\
Formula & $\mathrm{NN}_2$    & 8{,}511 & 25.22 & .915 & .090 & .055 & $+12.1$ \\
\midrule
\multicolumn{8}{@{}l}{\textit{Phase~1: learned forward mimic}} \\
Mimic   & CompenNet++        & 1{,}152{,}147 & 35.96 & .970 & .012 & .016 & --  \\
Mimic   & Sequential-5$'$    & 10{,}847      & 23.36 & .872 & .113 & .068 & --  \\
\midrule
\multicolumn{8}{@{}l}{\textit{Phase~2: predistortion through frozen mimic}} \\
Pure AR   & through CompenNet++       & --      & 13.09 & .561 & --   & --   & --      \\
Two-phase & $\mathrm{NN}_2\to$CompenNet++  & 8{,}511 & 24.78 & .890 & .097 & .058 & $+11.7$ \\
Pure AR   & through Sequential-5$'$   & --      & 13.01 & .555 & --   & --   & --      \\
Two-phase & $\mathrm{NN}_2\to$Sequential-5$'$ & 8{,}511 & 28.86 & .941 & .054 & .036 & $+15.9$ \\
\bottomrule
\end{tabular}}
\end{table}


\section{Qualitative Results: Scenario 5 Visual Comparison}
\begin{figure}[t]
    \centering\tablestyle
    \includegraphics[width=0.7\linewidth]{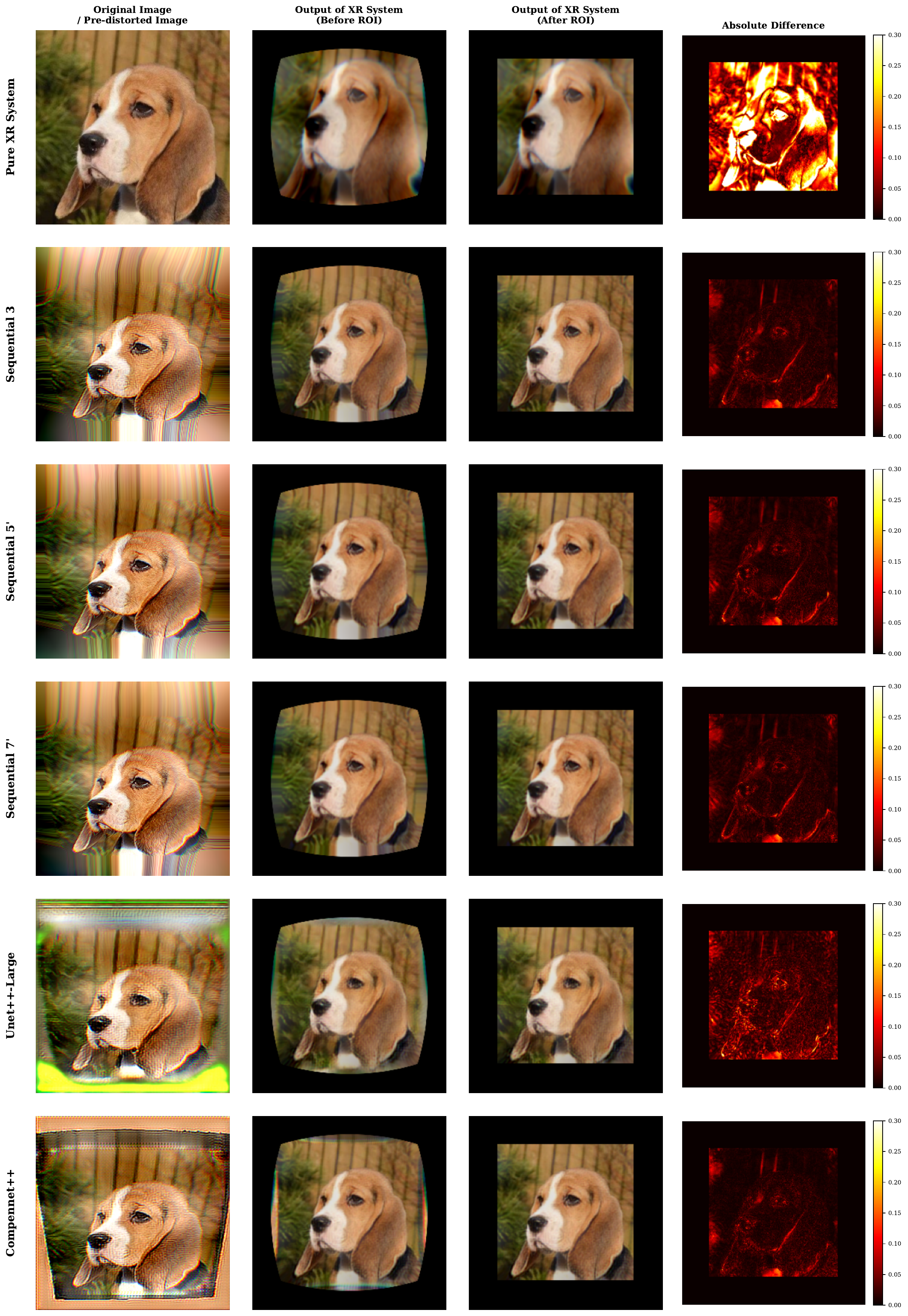}
    \caption{Qualitative comparison of distortion compensation models under the full
    spatiotonal Scenario~5 pipeline, as mentioned in the paper. Each row
    corresponds to one model. Columns show: (1) the original/pre-distorted image
    input to the model, (2) the XR system output before ROI cropping, (3) the XR
    system output after ROI cropping (the quantity measured by all metrics), and
    (4) the absolute pixel difference between the corrected output and the expected
    image, where brighter regions indicate larger residual error. All difference maps
    share the same $[0,\, 0.30]$ colormap scale.}
    \label{fig:qualitative_scenario5}
\end{figure}

Figure~\ref{fig:qualitative_scenario5} presents a qualitative comparison of all
evaluated models on a representative sample from the Scenario 5 test set (full
spatiotonal pipeline: geometric + non-geometric distortions). Each row shows the
original/pre-distorted image fed to the model, the output of the XR system before
and after ROI cropping, and the absolute difference map between the corrected output
and the expected image. The pure XR system row establishes the baseline distortion
magnitude, where strong geometric warping and tonal artifacts are clearly visible
across the entire frame. The absolute difference maps confirm the quantitative
rankings reported in Table~\ref{tab:mixed_scenario5}: Sequential~5$'$ and
Sequential~7$'$ produce visibly cleaner outputs than UNet++-Large despite using
orders of magnitude fewer parameters, while CompenNet++ remains the strongest
baseline with the lowest residual error. Notably, the Sequential models show
competitive perceptual quality in the After-ROI column despite their dramatic
efficiency advantage, with residual errors concentrated primarily at fine structural
edges rather than broad tonal regions.

\section{Applicability on higher resolution and other Datasets}
\label{sec:supp-generalization}
Tables~\ref{tab:ffhq_all_scenarios} and~\ref{tab:celeba_generalization} report two distinct evaluation settings. Table~\ref{tab:ffhq_all_scenarios} evaluates models trained directly on FFHQ at $1024\times1024$ for Scenarios~2 and~4, using the hyperparameters listed in Table~\ref{tab:hyperparameter_ffhq}. Table~\ref{tab:celeba_generalization} evaluates Images.cv-trained models zero-shot on CelebA-HQ-256 without fine-tuning for the full spatiotonal Scenario~5. These experiments assess higher-resolution training and cross-dataset generalization, respectively.

\begin{table}[htbp]
\caption{Hyperparameters of the FFHQ experiments. The rest of the hyperparameters are the same as the Images.cv dataset.}
\centering\tablestyle
\label{tab:hyperparameter_ffhq}
\begin{tabular}{|l l|}
\hline
\textbf{Hyperparameter} & \textbf{Value} \\
\hline
Batch size & 3 \\
Restart period & 6 epochs \\
Warmup epochs & 3 \\
Training epochs & 30 \\
Dataset fraction & 50\% of FFHQ dataset \\
\hline
\end{tabular}
\end{table}

\begin{table}[htbp]
\centering\tablestyle
\caption{FFHQ higher-resolution training results for Scenarios~2 and~4. Models are trained directly on FFHQ at $1024\times1024$ using the hyperparameters in Table~13. Values show improvement over baseline. Best in \textbf{bold}, second-best \underline{underlined}.}
\label{tab:ffhq_all_scenarios}
\fittable{
\begin{tabular}{|l|cc|cc|cc|}
\hline
& \multicolumn{2}{c|}{Scenario 2 (Non-Geo)} & \multicolumn{2}{c|}{Scenario 4 (Geo)} \\
Model & S↑ & P↑ & S↑ & P↑\\
\hline
Pure XR & .676 & 17.7 & .744 & 15.2 \\
G-Res (h=64) & - & - & \underline{.985 (+.241)} & \underline{40.5 (+25.4)}\\
G-Res (h=256) & - & - & .986 (+.242) & 41.0 (+25.8)\\
Patch2Pixel-v2 (p=3, h=64) & .743 (+.066) & 22.6 (+4.97) & - & - \\
Patch2Pixel-v2 (p=3, h=256) & .736 (+.060) & 22.8 (+5.11) & - & -\\
UNet++ - Large & \underline{.874 (+.198)} & \underline{25.3 (+7.62)} & - & - \\
CompenNet++ & \textbf{.890 (+.213)} & \textbf{26.0 (+8.38)} & \textbf{.991 (+.247)} & \textbf{41.1 (+26.0)} \\
\hline
\end{tabular}
}

\vspace{0.2cm}
\textbf{Abbreviations:} S = SSIM, P = PSNR (dB).
\textbf{Note:} Scenario 2 = non-geometric distortions only; Scenario 4 = geometric distortions only. Distortion details and sequential model compositions are identical to those described in the paper.
\end{table}

\begin{table}[htbp]
\centering\tablestyle
\caption{Mixed Geometric and Non-Geometric Distortions on CelebA-HQ-256 (↑ higher is better, ↓ lower is better). Models pretrained on Images.cv dataset, evaluated on CelebA-HQ-256 to assess generalization. Values show improvement over baseline. Best in \textbf{bold}, second-best \underline{underlined}. Distortions are identical to those in Scenario~5 of the paper.}
\label{tab:celeba_generalization}
\fittable{
\begin{tabular}{|l|cccc|}
\hline
& \multicolumn{4}{c|}{CelebA-HQ-256 Generalization}  \\
Model & S↑ & P↑ & L↓ & R↓ \\
\hline
Pure XR & .644 & 14.9 & .565 & .179 \\
Sequential 1 & .947\tiny{+.303} & 29.0\tiny{+14.1} & .052\tiny{-.513} & .035\tiny{-.144} \\
Sequential 2 & .946\tiny{+.302} & 28.9\tiny{+14.0} & .060\tiny{-.505} & .036\tiny{-.143} \\
Sequential 3 & .947\tiny{+.303} & 29.1\tiny{+14.2} & .048\tiny{-.517} & .035\tiny{-.144} \\
Sequential 4 & .946\tiny{+.302} & 29.0\tiny{+14.1} & .057\tiny{-.508} & .035\tiny{-.144} \\
Sequential 5 & .948\tiny{+.304} & 29.1\tiny{+14.2} & .066\tiny{-.499} & .035\tiny{-.144} \\
Sequential 6 & .946\tiny{+.302} & 28.9\tiny{+14.0} & .073\tiny{-.492} & .036\tiny{-.143} \\
Sequential 7 & .949\tiny{+.305} & 29.2\tiny{+14.3} & .064\tiny{-.501} & .035\tiny{-.144} \\
Sequential 8 & .948\tiny{+.304} & 29.1\tiny{+14.2} & .070\tiny{-.495} & .035\tiny{-.144} \\
Sequential 9 & .906\tiny{+.262} & 27.1\tiny{+12.2} & .149\tiny{-.416} & .044\tiny{-.135} \\
Sequential 10 & .905\tiny{+.261} & 27.1\tiny{+12.2} & .149\tiny{-.416} & .044\tiny{-.135} \\
Unet++ - Small & .814\tiny{+.170} & 22.9\tiny{+8.0} & .150\tiny{-.415} & .072\tiny{-.107} \\
Unet++ - Medium & .854\tiny{+.210} & 25.9\tiny{+11.0} & .077\tiny{-.488} & .050\tiny{-.129} \\
Unet++ - Large & .888\tiny{+.244} & 27.3\tiny{+12.4} & .055\tiny{-.510} & .043\tiny{-.136} \\
CompenNet++ & .948\tiny{+.304} & 29.3\tiny{+14.4} & .032\tiny{-.533} & .034\tiny{-.145} \\\hline
\end{tabular}
}

\vspace{0.2cm}
\textbf{Abbreviations:} S = SSIM, P = PSNR (dB), L = LPIPS (lower is better, shown as negative improvement), R = RMSE (lower is better, shown as negative improvement). \\
\textbf{Note:} All models were trained on Images.cv dataset and evaluated on CelebA-HQ-256 without fine-tuning to assess cross-dataset generalization. Sequential model compositions are identical to those in the paper. Sequential 9 and 10 demonstrate the importance of complete pipelines by omitting photometric correction models.
\end{table}

\section{Memory and Compute Costs}
\label{sec:supp-costs}
Table~\ref{tab:flops_rules} defines the component-level rules used to compute FLOPs, memory reads, and memory writes for every architectural primitive used in this work, ensuring full reproducibility of all efficiency numbers reported in the paper. Table~\ref{tab:supp_model_summary} applies these rules to report the total parameters, FLOPs, and memory bandwidth for all models evaluated, including both baselines and proposed architectures. Together, these tables substantiate the efficiency claims in Section~\ref{sec:efficiency}, showing that our sequential models require 13--21$\times$ fewer FLOPs and 6--10$\times$ less memory bandwidth than UNet++-Large, while matching its quality or exceeding it.

\begin{table*}[t]
\centering\tablestyle
\caption{Base Component Rules for FLOPs, Memory Reads, and Memory Writes}

\small
\textbf{Notation:}
\begin{itemize}
    \item $N$ = num\_elements, $B$ = batch\_size, $s$ = scale\_factor
    \item $C$ = channels, $C_{in}$ = input channels, $C_{out}$ = output channels
    \item $H, W$ = input spatial dimensions, $H_{out}, W_{out}$ = output spatial dimensions
    \item $K$ = kernel size (can be tuple $K_h \times K_w$), $in_f$ = input features, $out_f$ = output features
    \item $groups$ = number of groups for grouped convolution
    \item $stride$ = convolution stride (default: 1), $padding$ = zero-padding (default: 0)
\end{itemize}

\textbf{Output Dimension Formulas} (all divisions are integer division $\lfloor \cdot \rfloor$):
\begin{itemize}
    \item Conv2d: $H_{out} = \lfloor (H + 2 \times padding - K) / stride \rfloor + 1$
    \item TransConv2d: $H_{out} = (H - 1) \times stride - 2 \times padding + K + output\_padding$
    \item MaxPool/AvgPool: $H_{out} = \lfloor (H - K) / stride \rfloor + 1$ (default stride = $K$)
\end{itemize}

\vspace{0.5em}
\fittable{%
\label{tab:flops_rules}
\begin{tabular}{|l|l|l|l|l|}
\hline
\textbf{Component} & \textbf{Input/Output Dims} & \textbf{FLOPs} & \textbf{Memory Reads (bytes)} & \textbf{Memory Writes (bytes)} \\
\hline
Conv2d & $(C_{in}, H, W) \to (C_{out}, H_{out}, W_{out})$ & $2 \cdot H_{out} \cdot W_{out} \cdot C_{in} \cdot K^2 \cdot C_{out} / groups$ & $C_{in} \cdot H \cdot W \cdot 4 + C_{out} \cdot \frac{C_{in}}{groups} \cdot K^2 \cdot 4$ & $C_{out} \cdot H_{out} \cdot W_{out} \cdot 4$ \\
\hline
TransConv2d & $(C_{in}, H, W) \to (C_{out}, H_{out}, W_{out})$ & $2 \cdot H_{out} \cdot W_{out} \cdot C_{in} \cdot K^2 \cdot C_{out} / groups$ & $C_{in} \cdot H \cdot W \cdot 4 + \frac{C_{in}}{groups} \cdot C_{out} \cdot K^2 \cdot 4$ & $C_{out} \cdot H_{out} \cdot W_{out} \cdot 4$ \\
\hline
Linear & $(B, in_f) \to (B, out_f)$ & $2 \cdot B \cdot in_f \cdot out_f$ & $B \cdot in_f \cdot 4 + in_f \cdot out_f \cdot 4$ & $B \cdot out_f \cdot 4$ \\
\hline
BatchNorm & $(C, H, W) \to (C, H, W)$ & $2 \cdot C \cdot H \cdot W$ & $C \cdot H \cdot W \cdot 4 + 4 \cdot C \cdot 4$ & $C \cdot H \cdot W \cdot 4$ \\
\hline
ReLU & $N \to N$ & $N$ & $N \cdot 4$ & $N \cdot 4$ \\
\hline
LeakyReLU & $N \to N$ & $2 \cdot N$ & $N \cdot 4$ & $N \cdot 4$ \\
\hline
Sigmoid & $N \to N$ & $4 \cdot N$ & $N \cdot 4$ & $N \cdot 4$ \\
\hline
Tanh & $N \to N$ & $4 \cdot N$ & $N \cdot 4$ & $N \cdot 4$ \\
\hline
Clamp & $N \to N$ & $0$ & $N \cdot 4$ & $N \cdot 4$ \\
\hline
Scale ($ax+b$) & $N \to N$ & $2 \cdot N$ & $N \cdot 4$ & $N \cdot 4$ \\
\hline
GridSample (Interpolation) & $(3, H, W) + grid \to (3, H, W)$ & $8 \cdot H \cdot W \cdot 3$ & $3 \cdot H \cdot W \cdot 4 + H \cdot W \cdot 2 \cdot 4$ & $H \cdot W \cdot 3 \cdot 4$ \\
\hline
MaxPool & $(C, H, W) \to (C, H_{out}, W_{out})$ & $(K^2 - 1) \cdot H_{out} \cdot W_{out} \cdot C$ & $C \cdot H \cdot W \cdot 4$ & $C \cdot H_{out} \cdot W_{out} \cdot 4$ \\
\hline
AvgPool & $(C, H, W) \to (C, H_{out}, W_{out})$ & $(K^2 + 1) \cdot H_{out} \cdot W_{out} \cdot C$ & $C \cdot H \cdot W \cdot 4$ & $C \cdot H_{out} \cdot W_{out} \cdot 4$ \\
\hline
Concat & $[T_1, T_2, ...] \to T_{out}$ & $0$ & $\sum(size_i) \cdot 4$ & $size_{out} \cdot 4$ \\
\hline
Add & $N, N \to N$ & $N$ & $2 \cdot N \cdot 4$ & $N \cdot 4$ \\
\hline
Multiply & $N, N \to N$ & $N$ & $2 \cdot N \cdot 4$ & $N \cdot 4$ \\
\hline
Upsample (bilinear) & $(C, H, W) \to (C, Hs, Ws)$ & $8 \cdot (Hs) \cdot (Ws) \cdot C$ & $C \cdot H \cdot W \cdot 4$ & $C \cdot (Hs) \cdot (Ws) \cdot 4$ \\
\hline
Upsample (nearest) & $(C, H, W) \to (C, Hs, Ws)$ & $0$ & $C \cdot H \cdot W \cdot 4$ & $C \cdot (Hs) \cdot (Ws) \cdot 4$ \\
\hline
Patch Extract & $(3, H, W) \to (HW, K^2 \cdot 3)$ & $0$ & $3 \cdot H \cdot W \cdot 4$ & $H \cdot W \cdot K^2 \cdot 3 \cdot 4$ \\
\hline
Coordinate Gen & $HW$ pixels $\to (x, y)$ per pixel & $0$ & $0$ & $0$ \\
\hline
Identity & $N \to N$ & $0$ & $0$ & $0$ \\
\hline
\end{tabular}%
}

\vspace{0.5em}
\footnotesize
\textbf{Notes:}
\begin{itemize}
    \item All memory values in bytes (float32 = 4 bytes). Divide by $1,048,576$ for MB.
    \item Biases are excluded from all calculations.
    \item All calculations assume \textbf{inference mode} (BatchNorm uses running\_mean/var, not computed).
    \item \textbf{BatchNorm reads}: $4 \times C \times 4$ = reading 4 parameters per channel (mean, var, $\gamma$, $\beta$) $\times$ 4 bytes each.
    \item \textbf{GridSample} and \textbf{Patch Extract} use fixed $C=3$ (RGB) as they operate on image input/output.
    \item \textbf{Patch Extract} writes $K^2 \times 3$ values per Pixel. For Win models (Patch2Pixel (v2)/Win/Act), coordinates are generated implicitly (0 memory) and the Linear layer combines Patches + coords into $K^2 \times 5$ input features.
\end{itemize}

\end{table*}

\begin{table*}[t]
\centering\tablestyle
\caption{Model Summary: FLOPs, Memory, and Parameters}
\label{tab:supp_model_summary}
\fittable{%
\begin{tabular}{|l|r|r|r|r|r|r|}
\hline
\textbf{Model} & \textbf{Parameters} & \textbf{Size (MB)} & \textbf{Total FLOPs} & \textbf{Reads (MB)} & \textbf{Writes (MB)} & \textbf{Transfer (MB)} \\
\hline
\multicolumn{7}{|l|}{\textit{Baseline Models}} \\
\hline
CompenNet++ & 1,152,147 & 4.40 & 8,525,512,704 & 85.16 & 77.25 & 162.41 \\
UNet++ Large & 6,825,011 & 26.04 & 12,129,370,112 & 723.35 & 669.12 & 1392.48 \\
UNet++ Small & 2,552,561 & 9.74 & 5,606,285,312 & 652.91 & 625.53 & 1278.44 \\
UNet++ Medium & 3,241,487 & 12.37 & 6,358,827,008 & 666.36 & 634.25 & 1300.61 \\
\hline
\multicolumn{7}{|l|}{\textit{Single-Stage Models}} \\
\hline
C-Pixel2Pixel (h=64) & 4,757 & 0.02 & 615,907,328 & 72.52 & 69.25 & 141.77 \\
G-Res (h=64) & 4,482 & 0.02 & 580,780,032 & 68.52 & 66.75 & 135.27 \\
G-Hybrid (h=16) & 1,118 & 0.00 & 144,572,416 & 31.00 & 28.75 & 59.75 \\
G-Hybrid (h=64) & 7,150 & 0.03 & 927,858,688 & 87.03 & 84.75 & 171.78 \\
C-Hybrid (h=16) & 1,437 & 0.01 & 184,614,912 & 33.76 & 30.00 & 63.76 \\
C-Hybrid (h=64) & 7,885 & 0.03 & 1,022,427,136 & 89.78 & 86.00 & 175.78 \\
Patch2Pixel-v1 (h=64, p=3) & 7,437 & 0.03 & 966,590,464 & 90.28 & 73.75 & 164.03 \\
Patch2Pixel-v1 (h=64, p=5) & 12,797 & 0.05 & 1,669,136,384 & 130.30 & 85.75 & 216.05 \\
Patch2Pixel-v2 (h=64, p=3) & 7,437 & 0.03 & 967,180,288 & 93.28 & 76.00 & 169.28 \\
Patch2Pixel-v2 (h=64, p=5) & 12,797 & 0.05 & 1,669,726,208 & 133.30 & 88.00 & 221.30 \\
Patch2Pixel-v3 (h=64, p=3) & 6,365 & 0.02 & 826,671,104 &  85.27 & 76.00 & 161.27 \\
Patch2Pixel-v3 (h=64, p=5) & 9,581 & 0.04 & 1,248,198,656 & 109.29 & 88.00 & 197.29 \\

\hline
\multicolumn{7}{|l|}{\textit{Multi-Stage Models (Sequential models, see the sequential model details)}} \\
\hline
Sequential 1/2 (h=64) & 17,279 & 0.07 & 2,250,506,240 & 201.81 & 154.75 & 356.56 \\
Sequential 3/4 (h=64) & 22,036 & 0.08 & 2,866,413,568 & 274.33 & 224.00 & 498.33 \\
Sequential 5/6 (h=64) & 11,919 & 0.05 & 1,547,960,320 & 161.79 & 142.75 & 304.54 \\
Sequential 7/8 (h=64) & 16,676 & 0.06 & 2,163,867,648 & 234.31 & 212.00 & 446.31 \\
Sequential 9/10 (h=64) & 9,239 & 0.04 & 1,196,687,360 & 141.03 & 136.00 & 277.03 \\
Sequential 5 (h=32) & 3,847 & 0.01 & 516,161,536 & 97.76 & 78.75 & 176.51 \\
Sequential 5 (h=16) & 1,479 & 0.01 & 201,588,736 & 65.76 & 46.75 & 112.51 \\
Sequential 7 (h=32) & 5,142 & 0.02 & 691,666,944 & 138.27 & 116.00 & 254.27 \\
Sequential 7 (h=16) & 1,878 & 0.01 & 57,556,480 & 90.26 & 68.00 & 158.26 \\
\textbf{Sequential 5' }(h=64) & 10,847 & 0.04 & 1,407,451,136&   153.79 & 142.75 &      296.54 \\
\textbf{Sequential 5' }(h=32) &3,423&0.01&442,761,216
&89.76 &78.75&168.51 \\
Sequential 5'' (h=64) &  36,567&0.14&1,054,283,660 &94.99 &89.37 &184.36 \\
Sequential 7' (h=64) & 15,604 & 0.06 & 2,023,358,464 & 226.31 & 212.00 & 438.31\\
Sequential 7' (h=32) & 4,788&0.02&618,266,624 &130.27&116.00&246.27\\
Sequential 7'' (h=64) & 41,324 &       0.16 &   1,670,190,988 &  167.50&     158.62 & 326.12\\
\hline
\end{tabular}%
}
\vspace{0.5em}
\footnotesize
\begin{itemize}
    \item Input: $3 \times 256 \times 256$ RGB image. Output: $3 \times 256 \times 256$ (same size).
    \item Size (MB) = Parameters $\times$ 4 bytes / $1,048,576$.
    \item Transfer = Reads + Writes (total memory bandwidth).
    \item h = hidden dimension, p = Patch size.
    \item Sequential models chain multiple single-stage models; configurations detailed in the paper.
    \item Seq5 = G-Res $\rightarrow$ Patch2Pixel-v2 (p=3); Seq7 = G-Res $\rightarrow$ C-Pixel2Pixel $\rightarrow$ Patch2Pixel-v2 (p=3).
\end{itemize}
\end{table*}

\section{Detailed Single-Distortion Results}
\label{sec:supp-single}
The paper reports only the best-performing model per distortion type. Here we provide the full results for all models across all single distortions. Table~\ref{tab:geo_single} covers the five geometric distortions (Keystone, Fisheye, Pincushion, Barrel, Mustache). Table~\ref{tab:nongeo1} covers non-geometric distortions including all three WGNU variants, Color Balance, Elevated Black Level, Blur, Color Fringes, Vignetting, and Brightness \& Contrast. Table~\ref{tab:nongeo2} provides the full breakdown of all Brightness variants (v1/v2/v3, increase/decrease). Table~\ref{tab:nongeo3} provides the full breakdown of all Contrast variants (v1/v2/v3, increase/decrease). These tables also include negative results — notably the Patch2Patch variants — which consistently underperform single-pixel delta models and are therefore excluded from the recommended sequential pipelines.

\begin{table*}[t]
\centering\tablestyle
\caption{Geometric Distortions Performance (↑ higher is better, ↓ lower is better).}
\label{tab:geo_single}
\fittable{
\begin{tabular}{|l|cc|cc|cc|cc|cc|}
\hline
& \multicolumn{2}{c|}{Keystone} & \multicolumn{2}{c|}{Fisheye} & \multicolumn{2}{c|}{Pincushion} & \multicolumn{2}{c|}{Barrel} & \multicolumn{2}{c|}{Mustache} \\
Model & S↑ & P↑ & S↑ & P↑ & S↑ & P↑ & S↑ & P↑ & S↑ & P↑\\
\hline
Pure XR & .541 & 13.2 & .648 & 15.5 & .479 & 14.3 & .569 & 16.1 & .516 & 14.7\\
G-Hybrid - H=16 & .979\tiny{+.438} & 35.3\tiny{+22.1} & .990\tiny{+.342} & 38.4\tiny{+23.0} & .821\tiny{+.342} & 23.7\tiny{+9.4} & .990\tiny{+.421} & 38.7\tiny{+22.6} & .983\tiny{+.467} & 36.5\tiny{+21.8}\\
G-Hybrid - H=64 & .987\tiny{+.446} & 38.1\tiny{+24.9} & .993\tiny{+.345} & 40.4\tiny{+25.0} & .958\tiny{+.479} & 32.5\tiny{+18.2} & .992\tiny{+.423} & 39.7\tiny{+23.6} & .988\tiny{+.472} & 38.8\tiny{+24.2}\\
G-Res & .987\tiny{+.446} & 38.2\tiny{+25.0} & .993\tiny{+.345} & 40.6\tiny{+25.1} & .958\tiny{+.479} & 32.5\tiny{+18.2} & .992\tiny{+.423} & 39.7\tiny{+23.7} & .988\tiny{+.472} & 39.0\tiny{+24.3} \\
unet++ - Small & .785\tiny{+.244} & 22.7\tiny{+9.5} & .850\tiny{+.202} & 23.8\tiny{+8.4} & .726\tiny{+.247} & 21.2\tiny{+6.9} & .857\tiny{+.288} & 23.5\tiny{+7.4} & .794\tiny{+.278} & 22.6\tiny{+7.9}\\
unet++ - Medium & .820\tiny{+.279} & 25.5\tiny{+12.3} & .875\tiny{+.227} & 27.5\tiny{+12.1} & .775\tiny{+.296} & 25.3\tiny{+11.0} & .898\tiny{+.329} & 29.2\tiny{+13.1} & .831\tiny{+.315} & 27.0\tiny{+12.3}\\
unet++ - Large & .867\tiny{+.326} & 28.2\tiny{+14.9} & .923\tiny{+.275} & 30.3\tiny{+14.8} & .833\tiny{+.354} & 27.1\tiny{+12.8} & .938\tiny{+.369} & 31.6\tiny{+15.5} & .901\tiny{+.385} & 29.7\tiny{+15.1} \\
compennet++ &
.990 \tiny{+.449}&
39.3 \tiny{+26.1}&
.994\tiny{+.346} & 41.4\tiny{+25.9} & .957\tiny{+.478} & 32.1\tiny{+17.8} & .993\tiny{+.424} & 40.4\tiny{+24.3} & .991\tiny{+.475} & 39.8\tiny{+25.1}\\
\hline
\end{tabular}
}
\end{table*}

\begin{table*}[t]
\centering\tablestyle
\caption{Non-Geometric Distortions Part 1: WGNU Variations and Basic Distortions (↑ higher is better, ↓ lower is better).}
\label{tab:nongeo1}
\fittable{
\begin{tabular}{|l|cc|cc|cc|cc|cc|cc|cc|cc|cc|}
\hline
& \multicolumn{2}{c|}{WG-v1} & \multicolumn{2}{c|}{WG-v2} & \multicolumn{2}{c|}{WG-v3} & \multicolumn{2}{c|}{ColBal} & \multicolumn{2}{c|}{ElevBlk} & \multicolumn{2}{c|}{Blur} & \multicolumn{2}{c|}{ColFrg} & \multicolumn{2}{c|}{Vign} & \multicolumn{2}{c|}{Br\&Ct}\\
Model & S↑ & P↑ & S↑ & P↑ & S↑ & P↑ & S↑ & P↑ & S↑ & P↑ & S↑ & P↑ & S↑ & P↑ & S↑ & P↑ & S↑ & P↑\\
\hline
Pure XR & .996 & 34.4 & .984 & 25.4 & .976 & 23.4 & .922 & 17.8 & .971 & 29.5 & .823 & 26.5 & .746 & 22.6 & .955 & 18.5 & .492 & 9.4 \\
C-Hybrid - H=16 & .980\tiny{-.016} & 29.4\tiny{-5.0} & .969\tiny{-.015} & 27.8\tiny{+2.4} & .963\tiny{-.013} & 24.7\tiny{+1.3} & .991\tiny{+.069} & 41.5\tiny{+23.7} & .974\tiny{+.003} & 35.4\tiny{+5.9} & .869\tiny{+.046} & 27.0\tiny{+0.5} & .802\tiny{+.056} & 24.3\tiny{+1.7} & .945\tiny{-.010} & 20.9\tiny{+2.4} & .329\tiny{-.163} & 4.5\tiny{-4.9} \\
C-Hybrid - H=64 & .986\tiny{-.010} & 34.8\tiny{+0.4} & .975\tiny{-.009} & 29.8\tiny{+4.4} & .971\tiny{-.005} & 24.8\tiny{+1.4} & .996\tiny{+.074} & 47.7\tiny{+29.9} & .980\tiny{+.009} & 39.1\tiny{+9.6} & .875\tiny{+.052} & 27.3\tiny{+0.8} & .841\tiny{+.095} & 26.3\tiny{+3.7} & .956\tiny{+.001} & 22.6\tiny{+4.1} & .329\tiny{-.163} & 4.5\tiny{-4.9} \\
Patch2Pixel (v1) - P=3 & .986\tiny{-.010} & 34.3\tiny{-0.1} & .986\tiny{+.002} & 33.8\tiny{+8.4} & .978\tiny{+.002} & 25.2\tiny{+1.8} & .999\tiny{+.077} & 52.6\tiny{+34.8} & .979\tiny{+.008} & 38.1\tiny{+8.6} & .876\tiny{+.053} & 27.1\tiny{+0.6} & .800\tiny{+.054} & 24.9\tiny{+2.3} & .961\tiny{+.006} & 22.9\tiny{+4.4} & .329\tiny{-.163} & 4.5\tiny{-4.9} \\
Patch2Pixel (v1) - P=5 & .989\tiny{-.007} & 35.4\tiny{+1.0} & .994\tiny{+.010} & 35.5\tiny{+10.1} & .977\tiny{+.001} & 25.2\tiny{+1.8} & .998\tiny{+.076} & 52.5\tiny{+34.7} & .984\tiny{+.013} & 41.4\tiny{+11.9} & .873\tiny{+.050} & 26.9\tiny{+0.4} & .876\tiny{+.130} & 28.0\tiny{+5.4} & .968\tiny{+.013} & 23.1\tiny{+4.6} & .329\tiny{-.163} & 4.5\tiny{-4.9} \\
C-Pixel2Pixel & .999\tiny{+.003} & 44.5\tiny{+10.1} & .996\tiny{+.012} & 37.0\tiny{+11.6} & .975\tiny{-.001} & 23.6\tiny{+0.2} & 1.00\tiny{+.078} & 84.2\tiny{+66.4} & .986\tiny{+.015} & 43.0\tiny{+13.5} & .832\tiny{+.009} & 27.1\tiny{+0.6} & .743\tiny{-.003} & 22.6\tiny{+0} & .970\tiny{+.015} & 23.3\tiny{+4.8} & .999\tiny{+.507} & 57.7\tiny{+48.3} \\
Patch2Pixel (v2) - P=3 & .999\tiny{+.003} & 44.7\tiny{+10.3} & .996\tiny{+.012} & 37.1\tiny{+11.7} & .976\tiny{+0} & 23.6\tiny{+0.2} & 1.00\tiny{+.078} & 67.1\tiny{+49.3} & .986\tiny{+.015} & 42.7\tiny{+13.2} & .889\tiny{+.066} & 27.9\tiny{+1.4} & .848\tiny{+.102} & 25.8\tiny{+3.2} & .972\tiny{+.017} & 23.2\tiny{+4.7} & .997\tiny{+.505} & 49.1\tiny{+39.7} \\
Patch2Pixel (v2) - P=5 & .998\tiny{+.002} & 44.3\tiny{+9.9} & .996\tiny{+.012} & 37.0\tiny{+11.6} & .976\tiny{+0} & 25.2\tiny{+1.8} & .999\tiny{+.077} & 66.2\tiny{+48.4} & .986\tiny{+.015} & 42.7\tiny{+13.2} & .886\tiny{+.063} & 27.8\tiny{+1.3} & .916\tiny{+.170} & 29.5\tiny{+6.9} & .976\tiny{+.021} & 23.2\tiny{+4.7} & .995\tiny{+.503} & 47.2\tiny{+37.8} \\
Patch2Patch ($h$=32, $p_{in}$=5, $p_{out}$=3) & .996\tiny{+.000} & 40.0\tiny{+5.6} & .989\tiny{+.005} & 34.2\tiny{+8.8} & .975\tiny{-.001} & 23.6\tiny{+0.2} & .953\tiny{+.031} & 30.2\tiny{+12.4} & .984\tiny{+.013} & 40.6\tiny{+11.1} & .794\tiny{-.029} & 25.7\tiny{-0.8} & .706\tiny{-.040} & 22.6\tiny{+0.0} & .959\tiny{+.004} & 22.9\tiny{+4.4} & .465\tiny{-.027} & 15.6\tiny{+6.2} \\
Patch2Patch ($h$=64, $p_{in}$=5, $p_{out}$=3) & .997\tiny{+.001} & 42.0\tiny{+7.6} & .990\tiny{+.006} & 34.8\tiny{+9.4} & .975\tiny{-.001} & 23.6\tiny{+0.2} & .954\tiny{+.032} & 30.3\tiny{+12.5} & .984\tiny{+.013} & 40.5\tiny{+11.0} & .798\tiny{-.025} & 25.9\tiny{-0.6} & .711\tiny{-.035} & 22.7\tiny{+0.1} & .960\tiny{+.005} & 22.9\tiny{+4.4} & .468\tiny{-.024} & 15.6\tiny{+6.2} \\
Patch2Patch ($h$=128, $p_{in}$=5, $p_{out}$=3) & .997\tiny{+.001} & 41.7\tiny{+7.3} & .990\tiny{+.006} & 34.9\tiny{+9.5} & .975\tiny{-.001} & 23.6\tiny{+0.2} & .954\tiny{+.032} & 30.3\tiny{+12.5} & .984\tiny{+.013} & 40.6\tiny{+11.1} & .798\tiny{-.025} & 25.9\tiny{-0.6} & .710\tiny{-.036} & 22.7\tiny{+0.1} & .961\tiny{+.006} & 22.9\tiny{+4.4} & .465\tiny{-.027} & 15.7\tiny{+6.3} \\
Patch2Patch ($h$=256, $p_{in}$=5, $p_{out}$=3) & .997\tiny{+.001} & 42.6\tiny{+8.2} & .990\tiny{+.006} & 34.9\tiny{+9.5} & .975\tiny{-.001} & 23.6\tiny{+0.2} & .955\tiny{+.033} & 30.3\tiny{+12.5} & .985\tiny{+.014} & 40.6\tiny{+11.1} & .800\tiny{-.023} & 25.9\tiny{-0.6} & .712\tiny{-.034} & 22.8\tiny{+0.2} & .960\tiny{+.005} & 22.9\tiny{+4.4} & .469\tiny{-.023} & 15.7\tiny{+6.3} \\
Patch2Pixel (v3) ($h$=64, $p$=3) & .999\tiny{+.003} & 44.5\tiny{+10.1} & .996\tiny{+.012} & 37.1\tiny{+11.7} & .975\tiny{-.001} & 23.6\tiny{+0.2} & 1.00\tiny{+.078} & 72.2\tiny{+54.4} & .986\tiny{+.015} & 42.7\tiny{+13.2} & .883\tiny{+.060} & 27.7\tiny{+1.2} & .850\tiny{+.104} & 25.9\tiny{+3.3} & .972\tiny{+.017} & 23.2\tiny{+4.7} & .998\tiny{+.506} & 50.2\tiny{+40.8} \\
unet++ - Small & .820\tiny{-.176} & 22.2\tiny{-12.2} & .816\tiny{-.168} & 22.0\tiny{-3.4} & .832\tiny{-.144} & 20.5\tiny{-2.9} & .881\tiny{-.041} & 22.8\tiny{+5.0} & .813\tiny{-.158} & 22.1\tiny{-7.4} & .786\tiny{-.037} & 24.1\tiny{-2.4} & .607\tiny{-.139} & 19.9\tiny{-2.7} & .819\tiny{-.136} & 19.8\tiny{+1.3} & .801\tiny{+.309} & 21.8\tiny{+12.4} \\
unet++ - Medium & .920\tiny{-.076} & 28.5\tiny{-5.9} & .917\tiny{-.067} & 27.9\tiny{+2.5} & .892\tiny{-.084} & 23.0\tiny{-0.4} & .967\tiny{+.045} & 35.7\tiny{+17.9} & .905\tiny{-.066} & 28.1\tiny{-1.4} & .872\tiny{+.049} & 27.7\tiny{+1.2} & .800\tiny{+.054} & 25.4\tiny{+2.8} & .905\tiny{-.050} & 22.1\tiny{+3.6} & .941\tiny{+.449} & 27.3\tiny{+17.9} \\
unet++ - Large & .983\tiny{-.013} & 37.8\tiny{+3.4} & .981\tiny{-.003} & 34.5\tiny{+9.1} & .964\tiny{-.012} & 24.4\tiny{+1.0} & .984\tiny{+.062} & 39.2\tiny{+21.4} & .972\tiny{+.001} & 37.5\tiny{+8.0} & .888\tiny{+.065} & 28.2\tiny{+1.7} & .941\tiny{+.195} & 32.3\tiny{+9.7} & .956\tiny{+.001} & 23.0\tiny{+4.5} & .988\tiny{+.496} & 40.3\tiny{+30.9} \\
compennet++ & .996\tiny{+0} & 42.4\tiny{+8.0} & .989\tiny{+.005} & 35.5\tiny{+10.1} & .950\tiny{-.026} & 23.4\tiny{+0} & .999\tiny{+.077} & 52.1\tiny{+34.3} & .985\tiny{+.014} & 41.9\tiny{+12.4} & .890\tiny{+.067} & 28.2\tiny{+1.7} & .938\tiny{+.192} & 31.9\tiny{+9.3} & .952\tiny{-.003} & 22.9\tiny{+4.4} & .999\tiny{+.507} & 50.8\tiny{+41.4} \\

\hline
\end{tabular}
}

\vspace{0.2cm}
\textbf{Abbreviations:} WG = WGNU (Waveguide Non-Uniformity), ColBal = Color Balance, ElevBlk = Elevated Black Level, ColFrg = Color Fringes, Vign = Vignetting, Br\&Ct = Brightness \& Contrast, v1/v2/v3 = variation 1/2/3
\end{table*}

\begin{table*}[t]
\centering\tablestyle
\caption{Non-Geometric Distortions Part 2: Brightness Variations (↑ higher is better, ↓ lower is better).}
\label{tab:nongeo2}
\fittable{
\begin{tabular}{|l|cc|cc|cc|cc|cc|cc|cc|cc|}
\hline
& \multicolumn{2}{c|}{Br- v1} & \multicolumn{2}{c|}{Br+ v1} & \multicolumn{2}{c|}{Br- v2} & \multicolumn{2}{c|}{Br+ v2} & \multicolumn{2}{c|}{Br± v2} & \multicolumn{2}{c|}{Br- v3} & \multicolumn{2}{c|}{Br+ v3} & \multicolumn{2}{c|}{Br± v3} \\
Model & S↑ & P↑ & S↑ & P↑ & S↑ & P↑ & S↑ & P↑ & S↑ & P↑ & S↑ & P↑ & S↑ & P↑ & S↑ & P↑ \\
\hline
Pure XR & .514 & 11.5 & .752 & 11.3 & .862 & 19.9 & .927 & 19.9 & .887 & 20.4 & .851 & 19.4 & .923 & 19.4 & .896 & 20.4\\
C-Hybrid ($h=16$) & .942\tiny{+.428} & 20.3\tiny{+8.8} & .853\tiny{+.101} & 20.1\tiny{+8.8} & .946\tiny{+.084} & 25.6\tiny{+5.7} & .943\tiny{+.016} & 25.5\tiny{+5.6} & .884\tiny{-.003} & 21.0\tiny{+0.6} & .943\tiny{+.092} & 24.6\tiny{+5.2} & .943\tiny{+.020} & 25.2\tiny{+5.8} & .872\tiny{-.024} & 20.0\tiny{-0.4}\\
C-Hybrid ($h=64$) & .948\tiny{+.434} & 20.3\tiny{+8.8} & .853\tiny{+.101} & 20.1\tiny{+8.8} & .964\tiny{+.102} & 27.7\tiny{+7.8} & .948\tiny{+.021} & 27.0\tiny{+7.1} & .908\tiny{+.021} & 22.9\tiny{+2.5} & .944\tiny{+.093} & 24.9\tiny{+5.5} & .944\tiny{+.021} & 25.2\tiny{+5.8} & .878\tiny{-.018} & 20.5\tiny{+0.1}\\
Patch2Pixel-v1 ($p=3$) & .943\tiny{+.429} & 20.4\tiny{+8.9} & .852\tiny{+.100} & 20.1\tiny{+8.8} & .978\tiny{+.116} & 29.9\tiny{+10.0} & .962\tiny{+.035} & 31.2\tiny{+11.3} & .966\tiny{+.079} & 29.3\tiny{+8.9} & .948\tiny{+.097} & 25.0\tiny{+5.6} & .948\tiny{+.025} & 25.7\tiny{+6.3} & .881\tiny{-.015} & 20.5\tiny{+0.1}\\
Patch2Pixel-v1 ($p=5$) & .951\tiny{+.437} & 20.3\tiny{+8.8} & .857\tiny{+.105} & 20.1\tiny{+8.8} & .984\tiny{+.122} & 30.9\tiny{+11.0} & .965\tiny{+.038} & 32.3\tiny{+12.4} & .971\tiny{+.084} & 30.1\tiny{+9.7} & .951\tiny{+.100} & 25.2\tiny{+5.8} & .949\tiny{+.026} & 25.7\tiny{+6.3} & .882\tiny{-.014} & 20.5\tiny{+0.1} \\
C-Pixel2Pixel & .941\tiny{+.427} & 20.4\tiny{+8.9} & .849\tiny{+.097} & 20.2\tiny{+8.9} & .990\tiny{+.128} & 31.6\tiny{+11.7} & .967\tiny{+.040} & 33.5\tiny{+13.6} & .980\tiny{+.093} & 32.5\tiny{+12.1} & .952\tiny{+.101} & 25.2\tiny{+5.8} & .951\tiny{+.028} & 25.8\tiny{+6.4} & .885\tiny{-.011} & 20.6\tiny{+0.2} \\
Patch2Pixel-v2 ($p=3$) & .941\tiny{+.427} & 20.3\tiny{+8.8} & .845\tiny{+.093} & 20.1\tiny{+8.8} & .991\tiny{+.129} & 31.5\tiny{+11.6} & .967\tiny{+.040} & 33.2\tiny{+13.3} & .980\tiny{+.093} & 32.6\tiny{+12.2} & .952\tiny{+.101} & 25.2\tiny{+5.8} & .948\tiny{+.025} & 25.7\tiny{+6.3} & .883\tiny{-.013} & 20.6\tiny{+0.2}\\
Patch2Pixel-v2 ($p=5$) & .951\tiny{+.437} & 20.3\tiny{+8.8} & .855\tiny{+.103} & 20.1\tiny{+8.8} & .991\tiny{+.129} & 31.5\tiny{+11.6} & .968\tiny{+.041} & 33.2\tiny{+13.3} & .980\tiny{+.093} & 32.5\tiny{+12.1} & .953\tiny{+.102} & 25.2\tiny{+5.8} & .950\tiny{+.027} & 25.7\tiny{+6.3} & .881\tiny{-.015} & 20.6\tiny{+0.2} \\
Patch2Patch ($h$=32) & .939\tiny{+.425} & 20.3\tiny{+8.8} & .843\tiny{+.091} & 20.0\tiny{+8.7} & .989\tiny{+.127} & 31.5\tiny{+11.6} & .967\tiny{+.040} & 33.0\tiny{+13.1} & .977\tiny{+.090} & 32.3\tiny{+11.9} & .952\tiny{+.101} & 25.1\tiny{+5.7} & .950\tiny{+.027} & 25.6\tiny{+6.2} & .886\tiny{-.010} & 20.5\tiny{+0.1} \\
Patch2Patch ($h$=64) & .940\tiny{+.426} & 20.3\tiny{+8.8} & .840\tiny{+.088} & 20.0\tiny{+8.7} & .990\tiny{+.128} & 31.5\tiny{+11.6} & .967\tiny{+.040} & 33.1\tiny{+13.2} & .979\tiny{+.092} & 32.5\tiny{+12.1} & .952\tiny{+.101} & 25.2\tiny{+5.8} & .950\tiny{+.027} & 25.7\tiny{+6.3} & .885\tiny{-.011} & 20.6\tiny{+0.2} \\
Patch2Patch ($h$=128) & .939\tiny{+.425} & 20.3\tiny{+8.8} & .840\tiny{+.088} & 20.0\tiny{+8.7} & .990\tiny{+.128} & 31.5\tiny{+11.6} & .967\tiny{+.040} & 33.1\tiny{+13.2} & .980\tiny{+.093} & 32.5\tiny{+12.1} & .952\tiny{+.101} & 25.1\tiny{+5.7} & .950\tiny{+.027} & 25.7\tiny{+6.3} & .885\tiny{-.011} & 20.6\tiny{+0.2} \\
Patch2Patch ($h$=256) & .939\tiny{+.425} & 20.3\tiny{+8.8} & .840\tiny{+.088} & 20.0\tiny{+8.7} & .990\tiny{+.128} & 31.6\tiny{+11.7} & .967\tiny{+.040} & 33.1\tiny{+13.2} & .980\tiny{+.093} & 32.5\tiny{+12.1} & .952\tiny{+.101} & 25.1\tiny{+5.7} & .950\tiny{+.027} & 25.7\tiny{+6.3} & .886\tiny{-.010} & 20.6\tiny{+0.2} \\
Patch2Pixel (v3) ($h$=64) & .941\tiny{+.427} & 20.3\tiny{+8.8} & .845\tiny{+.093} & 20.1\tiny{+8.8} & .990\tiny{+.128} & 31.6\tiny{+11.7} & .967\tiny{+.040} & 33.2\tiny{+13.3} & .980\tiny{+.093} & 32.5\tiny{+12.1} & .952\tiny{+.101} & 25.2\tiny{+5.8} & .949\tiny{+.026} & 25.7\tiny{+6.3} & .881\tiny{-.015} & 20.6\tiny{+0.2} \\
unet++ - Small & .774\tiny{+.260} & 18.3\tiny{+6.8} & .680\tiny{-.072} & 17.9\tiny{+6.6} & .862\tiny{+0} & 22.0\tiny{+2.1} & .791\tiny{-.136} & 21.8\tiny{+1.9} & .809\tiny{-.078} & 21.6\tiny{+1.2} & .813\tiny{-.038} & 20.6\tiny{+1.2} & .761\tiny{-.162} & 20.5\tiny{+1.1} & .740\tiny{-.156} & 18.2\tiny{-2.2} \\
unet++ - Medium & .859\tiny{+.345} & 19.6\tiny{+8.1} & .814\tiny{+.062} & 19.8\tiny{+8.5} & .919\tiny{+.057} & 26.5\tiny{+6.6} & .885\tiny{-.042} & 27.0\tiny{+7.1} & .898\tiny{+.011} & 26.7\tiny{+6.3} & .919\tiny{+.068} & 24.7\tiny{+5.3} & .864\tiny{-.059} & 23.8\tiny{+4.4} & .779\tiny{-.117} & 19.8\tiny{-0.6} \\
unet++ - Large & .924\tiny{+.410} & 20.1\tiny{+8.6} & .809\tiny{+.057} & 19.8\tiny{+8.5} & .978\tiny{+.116} & 30.7\tiny{+10.8} & .953\tiny{+.026} & 31.9\tiny{+12.0} & .965\tiny{+.078} & 31.3\tiny{+10.9} & .942\tiny{+.091} & 25.0\tiny{+5.6} & .936\tiny{+.013} & 25.5\tiny{+6.1} & .843\tiny{-.053} & 20.4\tiny{+0} \\
compennet++ & .922\tiny{+.408} & 20.1\tiny{+8.6} & .804\tiny{+.052} & 19.8\tiny{+8.5} & .982\tiny{+.120} & 30.9\tiny{+11.0} & .962\tiny{+.035} & 32.5\tiny{+12.6} & .963\tiny{+.076} & 31.2\tiny{+10.8} & .939\tiny{+.088} & 25.1\tiny{+5.7} & .940\tiny{+.017} & 25.6\tiny{+6.2} & .822\tiny{-.074} & 20.4\tiny{+0.0} \\
\hline
\end{tabular}
}

\vspace{0.2cm}
\textbf{Abbreviations:} Br = Brightness, + = increase, - = decrease, ± = both increase and decrease, v1/v2/v3 = variation 1/2/3
\end{table*}

\begin{table*}[t]
\centering\tablestyle
\caption{Non-Geometric Distortions Part 3: Contrast Variations (↑ higher is better, ↓ lower is better).}
\label{tab:nongeo3}
\fittable{
\begin{tabular}{|l|cc|cc|cc|cc|cc|cc|}
\hline
& \multicolumn{2}{c|}{Ct+ v1} & \multicolumn{2}{c|}{Ct- v1} & \multicolumn{2}{c|}{Ct+ v2} & \multicolumn{2}{c|}{Ct- v2} & \multicolumn{2}{c|}{Ct+ v3} & \multicolumn{2}{c|}{Ct- v3} \\
Model & S↑ & P↑ & S↑ & P↑ & S↑ & P↑ & S↑ & P↑ & S↑ & P↑ & S↑ & P↑ \\
\hline
Pure XR & .806 & 21.9 & .857 & 18.9 & .953 & 28.1 & .949 & 28.4 & .951 & 28.3 & .952 & 28.2 \\
C-Hybrid ($h=16$) & .993\tiny{+.187} & 36.7\tiny{+14.8} & .882\tiny{+.025} & 21.3\tiny{+2.5} & .940\tiny{-.013} & 26.1\tiny{-2.0} & .941\tiny{-.008} & 26.5\tiny{-1.9} & .940\tiny{-.011} & 26.3\tiny{-2.0} & .941\tiny{-.011} & 26.2\tiny{-2.0} \\
C-Hybrid ($h=64$) & .995\tiny{+.189} & 44.7\tiny{+22.8} & .887\tiny{+.030} & 21.8\tiny{+2.9} & .953\tiny{+0} & 28.8\tiny{+0.7} & .950\tiny{+.001} & 28.4\tiny{+0} & .945\tiny{-.006} & 27.9\tiny{-0.4} & .944\tiny{-.008} & 27.5\tiny{-0.7} \\
Patch2Pixel-v1 ($p=3$) & .998\tiny{+.192} & 50.8\tiny{+28.9} & .889\tiny{+.032} & 22.1\tiny{+3.2} & .970\tiny{+.017} & 30.5\tiny{+2.4} & .965\tiny{+.016} & 29.3\tiny{+0.9} & .950\tiny{-.001} & 28.1\tiny{-0.2} & .947\tiny{-.005} & 28.2\tiny{+0} \\
Patch2Pixel-v1 ($p=5$) & .999\tiny{+.193} & 53.1\tiny{+31.2} & .892\tiny{+.035} & 22.0\tiny{+3.2} & .955\tiny{+.002} & 28.9\tiny{+0.8} & .973\tiny{+.024} & 32.3\tiny{+3.9} & .949\tiny{-.002} & 28.4\tiny{+0.1} & .949\tiny{-.003} & 28.2\tiny{+0} \\
C-Pixel2Pixel & 1.00\tiny{+.194} & 81.6\tiny{+59.7} & .893\tiny{+.036} & 22.2\tiny{+3.4} & .985\tiny{+.032} & 34.4\tiny{+6.3} & .989\tiny{+.040} & 35.9\tiny{+7.5} & .951\tiny{+0} & 28.5\tiny{+0.2} & .951\tiny{-.001} & 28.4\tiny{+0.2} \\
Patch2Pixel-v2 ($p=3$) & 1.00\tiny{+.194} & 64.9\tiny{+43.0} & .889\tiny{+.032} & 22.2\tiny{+3.3} & .985\tiny{+.032} & 34.8\tiny{+6.7} & .989\tiny{+.040} & 36.3\tiny{+7.9} & .951\tiny{+0} & 28.5\tiny{+0.2} & .950\tiny{-.002} & 28.4\tiny{+0.2} \\
Patch2Pixel-v2 ($p=5$) & 1.00\tiny{+.194} & 64.4\tiny{+42.5} & .897\tiny{+.040} & 22.2\tiny{+3.3} & .984\tiny{+.031} & 34.7\tiny{+6.6} & .988\tiny{+.039} & 36.0\tiny{+7.6} & .951\tiny{+0} & 28.5\tiny{+0.2} & .951\tiny{-.001} & 28.4\tiny{+0.2} \\
Patch2Patch ($h$=32) & .936\tiny{+.130} & 28.4\tiny{+6.5} & .858\tiny{+.001} & 21.2\tiny{+2.3} & .974\tiny{+.021} & 32.2\tiny{+4.1} & .978\tiny{+.029} & 33.4\tiny{+5.0} &
.951\tiny{-.000} & 28.3\tiny{+.0} &
.951\tiny{-.001} & 28.3\tiny{+.1}\\
Patch2Patch ($h$=64) & .936\tiny{+.130} & 28.4\tiny{+6.5} & .858\tiny{+.001} & 21.2\tiny{+2.3} & .975\tiny{+.022} & 32.5\tiny{+4.4} & .978\tiny{+.029} & 33.5\tiny{+5.1} &
.951\tiny{-.000} & 28.3\tiny{+.0} &
.951\tiny{-.001} & 28.2\tiny{+.0} \\
Patch2Patch ($h$=128) & .937\tiny{+.131} & 28.5\tiny{+6.6} & .857\tiny{+.000} & 21.3\tiny{+2.4} & .975\tiny{+.022} & 32.7\tiny{+4.6} & .979\tiny{+.030} & 33.6\tiny{+5.2} &
.950\tiny{-.001} & 28.3\tiny{+.0} &
.951\tiny{-.001} & 28.2\tiny{+.0}\\
Patch2Patch ($h$=256) & .937\tiny{+.131} & 28.5\tiny{+6.6} & .857\tiny{+.000} & 21.3\tiny{+2.4} & .975\tiny{+.022} & 32.8\tiny{+4.7} & .979\tiny{+.030} & 33.6\tiny{+5.2} &
.951\tiny{-.000} & 28.3\tiny{+.0} &
.950\tiny{-.002} & 28.2\tiny{+.0} \\
Patch2Pixel-v3 ($h$=64) & 1.00\tiny{+.194} & 69.4\tiny{+47.5} & .889\tiny{+.032} & 22.2\tiny{+3.3} & .985\tiny{+.032} & 34.8\tiny{+6.7} & .989\tiny{+.040} & 36.3\tiny{+7.9} &
.950\tiny{-.001} & 28.5\tiny{+.2} &
.950\tiny{-.002} & 28.4\tiny{+.2} \\
unet++ - Small & .851\tiny{+.045} & 22.4\tiny{+0.5} & .759\tiny{-.098} & 19.4\tiny{+0.6} & .870\tiny{-.083} & 22.3\tiny{-5.8} & .812\tiny{-.137} & 21.9\tiny{-6.5} & .785\tiny{-.166} & 21.2\tiny{-7.1} & .785\tiny{-.167} & 21.3\tiny{-6.9} \\
unet++ - Medium & .923\tiny{+.117} & 28.7\tiny{+6.8} & .861\tiny{+.004} & 21.8\tiny{+2.9} & .903\tiny{-.050} & 27.4\tiny{-0.7} & .905\tiny{-.044} & 27.6\tiny{-0.8} & .879\tiny{-.072} & 25.4\tiny{-2.9} & .877\tiny{-.075} & 25.4\tiny{-2.8} \\
unet++ - Large & .986\tiny{+.180} & 39.9\tiny{+18.0} & .874\tiny{+.017} & 21.9\tiny{+3.1} & .970\tiny{+.017} & 33.4\tiny{+5.3} & .974\tiny{+.025} & 34.3\tiny{+5.9} & .936\tiny{-.015} & 28.0\tiny{-0.3} & .938\tiny{-.014} & 28.0\tiny{-0.2} \\
compennet++ & .999\tiny{+.193} & 51.0\tiny{+29.1} & .867\tiny{+.010} & 22.0\tiny{+3.1} & .979\tiny{+.026} & 34.4\tiny{+6.3} & .984\tiny{+.035} & 35.8\tiny{+7.4} & .939\tiny{-.012} & 28.3\tiny{+0.0} & .938\tiny{-.014} & 28.2\tiny{+0.0} \\
\hline
\end{tabular}
}

\vspace{0.2cm}
\textbf{Abbreviations:} Ct = Contrast, + = increase, - = decrease, v1/v2/v3 = variation 1/2/3
\end{table*}

\section{Detailed Multi-Distortion Results}
\label{sec:supp-multi}
Tables~\ref{tab:nongeo_mixed} and~\ref{tab:geo_mixed} report the full model comparison for the non-geometric mixed scenarios (Scenarios~1--2) and geometric mixed scenarios (Scenarios~3--4) respectively, complementing the summary results in the paper. Table~\ref{tab:supp-scenario5full} gives the complete Scenario~5 ablation across all Sequential compositions and hidden dimensions.

\begin{table}[t]
\centering\tablestyle
\caption{Non-Geometric Mixed Distortions (↑ higher is better, ↓ lower is better). Scenario 1: described in the paper.}
\label{tab:nongeo_mixed}
\fittable{
\begin{tabular}{|l|cc|cc|}
\hline
& \multicolumn{2}{c|}{Scenario 1} & \multicolumn{2}{c|}{Scenario 2} \\
Model & S↑ & P↑ & S↑ & P↑ \\
\hline
Pure XR & .416 & 8.3 & .655 & 17.3 \\
C-Hybrid ($h=16$) & .782\tiny{+.366} & 24.8\tiny{+16.5} & .724\tiny{+.069} & 19.3\tiny{+2.0} \\
C-Hybrid ($h=64$) & .633\tiny{+.217} & 8.7\tiny{+0.4} & .798\tiny{+.143} & 21.5\tiny{+4.2} \\
Patch2Pixel-v1 ($p=3$) & .620\tiny{+.204} & 8.7\tiny{+0.4} & .797\tiny{+.142} & 21.6\tiny{+4.4} \\
Patch2Pixel (v1) ($p=5$) & .615\tiny{+.199} & 8.7\tiny{+0.4} & .843\tiny{+.188} & 22.7\tiny{+5.4} \\
C-Pixel2Pixel & .691\tiny{+.275} & 22.3\tiny{+14.0} & .735\tiny{+.080} & 21.4\tiny{+4.1} \\
Patch2Pixel-v2 ($p=3$) & .857\tiny{+.441} & 27.4\tiny{+19.1} & .863\tiny{+.208} & 23.0\tiny{+5.8} \\
Patch2Pixel-v2 ($p=5$) & .844\tiny{+.428} & 27.3\tiny{+19.1} & .873\tiny{+.218} & 23.3\tiny{+6.0} \\
Patch2Patch ($h$=32) & .665\tiny{+.249} & 23.0\tiny{+14.7} & .681\tiny{+.026} & 21.1\tiny{+3.8} \\
Patch2Patch ($h$=64) & .662\tiny{+.246} & 23.0\tiny{+14.7} & .681\tiny{+.026} & 21.2\tiny{+3.9} \\
Patch2Patch ($h$=128) & .678\tiny{+.262} & 23.5\tiny{+15.2} & .682\tiny{+.027} & 21.3\tiny{+4.0} \\
Patch2Patch ($h$=256) & .681\tiny{+.265} & 23.5\tiny{+15.2} & .679\tiny{+.024} & 21.3\tiny{+4.0} \\
Patch2Pixel-v3 ($h$=64) & .853\tiny{+.437} & 27.2\tiny{+18.9} & .858\tiny{+.203} & 23.0\tiny{+5.7} \\
UNet++ - Small & .507\tiny{+.091} & 8.6\tiny{+0.4} & .617\tiny{-.038} & 19.3\tiny{+2.0} \\
UNet++ - Medium & .835\tiny{+.419} & 27.6\tiny{+19.3} & .840\tiny{+.185} & 22.9\tiny{+5.6} \\
UNet++ - Large & .894\tiny{+.478} & 29.4\tiny{+21.1} & .888\tiny{+.233} & 23.5\tiny{+6.2} \\
CompenNet++ & .894\tiny{+.478} & 29.4\tiny{+21.1} & .870\tiny{+.215} & 23.3\tiny{+6.0} \\
\hline
\end{tabular}
}
\end{table}

\begin{table}[t]
\centering\tablestyle
\caption{Geometric Mixed Distortions (↑ higher is better, ↓ lower is better). Scenarios 3 \&  4: described in the paper.}
\label{tab:geo_mixed}
\fittable{
\begin{tabular}{|l|cc|cc|}
\hline
& \multicolumn{2}{c|}{Scenario 3} & \multicolumn{2}{c|}{Scenario 4} \\
Model & S↑ & P↑ & S↑ & P↑ \\
\hline
Pure XR & .704 & 15.3 & .650 & 14.9 \\
G-Hybrid ($h=16$) & .982\tiny{+.278} & 35.8\tiny{+20.5} & .968\tiny{+.318} & 32.9\tiny{+18.0} \\
G-Hybrid ($h=64$)& .987\tiny{+.283} & 37.4\tiny{+22.1} & .974\tiny{+.324} & 34.8\tiny{+19.9} \\
G-Res & .987\tiny{+.283} & 37.5\tiny{+22.2} & .975\tiny{+.325} & 34.9\tiny{+20.0} \\
UNet++ - Small & .857\tiny{+.153} & 24.1\tiny{+8.9} & .834\tiny{+.184} & 23.7\tiny{+8.8} \\
UNet++ - Medium & .877\tiny{+.173} & 27.1\tiny{+11.8} & .858\tiny{+.208} & 26.9\tiny{+12.0} \\
UNet++ - Large & .914\tiny{+.210} & 29.2\tiny{+14.0} & .895\tiny{+.245} & 28.7\tiny{+13.8} \\
CompenNet++ & .991\tiny{+.287} & 39.1\tiny{+23.8} & .983\tiny{+.333} & 36.3\tiny{+21.4} \\
\hline
\end{tabular}
}
\end{table}

\begin{table}[]
\centering\tablestyle
\caption{Full Scenario~5 ablation (mixed geometric +
   non-geometric): all Sequential compositions and hidden dimensions.
   Subscript values are the improvement over the Pure XR baseline
   (``$+$'' for SSIM/PSNR, ``$-$'' for LPIPS/RMSE).}
\label{tab:supp-scenario5full}
\fittable{
\begin{tabular}{l|cccc}
\hline
& \multicolumn{4}{c|}{Scenario 5}  \\
Hidden Dimensions & SSIM↑ & PSNR↑ & LPIPS↓ & RMSE↓ \\
\hline
Pure XR & .641 & 14.6 & .635 & .186 \\
\hline
\multicolumn{5}{l}{\textit{Sequential-1: G-Res $\rightarrow$ Patch2Pixel-v2 ($p=5$)}} \\
\hline
64, 64 & .942\tiny{+.301} & 27.4\tiny{+12.7} & .063\tiny{-.572} & .043\tiny{-.143} \\
\hline
\multicolumn{5}{l}{\textit{Sequential-2: Patch2Pixel-v2 ($p=5$) $\rightarrow$ G-Res}} \\
\hline
64, 64 & .941\tiny{+.300} & 27.3\tiny{+12.7} & .076\tiny{-.559} & .043\tiny{-.143} \\
\hline
\multicolumn{5}{l}{\textit{Sequential-3: G-Res $\rightarrow$ C-Res $\rightarrow$ Patch2Pixel-v2 ($p=5$)}} \\
\hline
64, 64, 64 & .943\tiny{+.302} & 27.5\tiny{+12.9} & .060\tiny{-.575} & .042\tiny{-.144} \\
\hline
\multicolumn{5}{l}{\textit{Sequential-4: C-Res $\rightarrow$ Patch2Pixel-v2 ($p=5$) $\rightarrow$ G-Res}} \\
\hline
64, 64, 64 & .940\tiny{+.299} & 27.4\tiny{+12.8} & .073\tiny{-.562} & .043\tiny{-.143} \\
\hline
\multicolumn{5}{l}{\textit{Sequential-5: G-Res $\rightarrow$ Patch2Pixel-v2 ($p=3$)}} \\
\hline
64, 64 & .943\tiny{+.302} & 27.5\tiny{+12.9} & .084\tiny{-.551} & .042\tiny{-.144} \\
32, 32 & .939\tiny{+.298} & 27.2\tiny{+12.6} & .093\tiny{-.542} & .044\tiny{-.142} \\
16, 16 & .923\tiny{+.282} & 26.4\tiny{+11.8} & .112\tiny{-.523} & .048\tiny{-.138} \\
\hline
\multicolumn{5}{l}{\textit{Sequential-5': G-Res $\rightarrow$ Patch2Pixel-v3 ($p=3$)}} \\
\hline
64, 64 & .943\tiny{+.302} & 27.5\tiny{+12.9} & .084\tiny{-.551} & .042\tiny{-.144} \\
64, 48 & .943\tiny{+.302} & 27.4\tiny{+12.8} & .085\tiny{-.550} & .043\tiny{-.143} \\
64, 32 & .940\tiny{+.299} & 27.2\tiny{+12.6} & .091\tiny{-.544} & .044\tiny{-.142} \\
32, 32 & .939\tiny{+.298} & 27.1\tiny{+12.5} & .091\tiny{-.544} & .044\tiny{-.142} \\
16, 16 & .929\tiny{+.288} & 26.6\tiny{+12.0} & .103\tiny{-.532} & .047\tiny{-.139} \\

\hline
\multicolumn{5}{l}{\textit{Sequential-5'': G-Res $\rightarrow$ Patch2Patch ($p_{in}$=5, $p_{out}$=3)}} \\
\hline
64, 128 & .913\tiny{+.272} & 26.7\tiny{+12.1} & .067\tiny{-.568} & .046\tiny{-.140} \\
64, 256 & .922\tiny{+.281} & 27.0\tiny{+12.4} & .057\tiny{-.578} & .045\tiny{-.141} \\
32, 128 & .918\tiny{+.277} & 26.9\tiny{+12.3} & .065\tiny{-.570} & .045\tiny{-.141} \\
32, 256 & .922\tiny{+.281} & 27.0\tiny{+12.4} & .059\tiny{-.576} & .045\tiny{-.141} \\
16, 128 & .911\tiny{+.270} & 26.5\tiny{+11.9} & .070\tiny{-.565} & .047\tiny{-.139} \\
16, 256 & .913\tiny{+.272} & 26.5\tiny{+11.9} & .062\tiny{-.573} & .047\tiny{-.139} \\
\hline
\multicolumn{5}{l}{\textit{Sequential-6: Patch2Pixel-v2 ($p=3$) $\rightarrow$ G-Res}} \\
\hline
64, 64 & .939\tiny{+.298} & 27.3\tiny{+12.6} & .093\tiny{-.542} & .043\tiny{-.143} \\
\hline
\multicolumn{5}{l}{\textit{Sequential 7: G-Res $\rightarrow$ C-Res $\rightarrow$ Patch2Pixel-v2 ($p=3$)}} \\
\hline
64, 64, 64 & .945\tiny{+.304} & 27.6\tiny{+13.0} & .082\tiny{-.553} & .042\tiny{-.144} \\
32, 32, 32 & .940\tiny{+.299} & 27.3\tiny{+12.7} & .091\tiny{-.544} & .043\tiny{-.143} \\
16, 16, 16 & .928\tiny{+.287} & 26.7\tiny{+12.1} & .107\tiny{-.528} & .046\tiny{-.140} \\
\hline
\multicolumn{5}{l}{\textit{Sequential-7': G-Res $\rightarrow$ C-Res $\rightarrow$ Patch2Pixel-v3 ($p=3$)}} \\
\hline
64, 64, 64 & .944\tiny{+.303} & 27.6\tiny{+13.0} & .083\tiny{-.552} & .042\tiny{-.144} \\
32, 32, 32 & .934\tiny{+.293} & 27.1\tiny{+12.5} & .097\tiny{-.538} & .044\tiny{-.142} \\
16, 16, 16 & .923\tiny{+.282} & 26.4\tiny{+11.8} & .106\tiny{-.529} & .048\tiny{-.138} \\
\hline
\multicolumn{5}{l}{\textit{Sequential-7'': G-Res $\rightarrow$ C-Res $\rightarrow$ Patch2Patch ($p_{in}$=5, $p_{out}$=3)}} \\
\hline
64, 64, 128 & .934\tiny{+.293} & 27.4\tiny{+12.8} & .053\tiny{-.582} & .043\tiny{-.143} \\
64, 64, 256 & .939\tiny{+.298} & 27.5\tiny{+12.9} & .048\tiny{-.587} & .042\tiny{-.144} \\
32, 32, 128 & .929\tiny{+.288} & 27.1\tiny{+12.5} & .056\tiny{-.579} & .044\tiny{-.142} \\
32, 32, 256 & .937\tiny{+.296} & 27.4\tiny{+12.8} & .049\tiny{-.586} & .043\tiny{-.143} \\
16, 16, 128 & .924\tiny{+.283} & 26.7\tiny{+12.1} & .061\tiny{-.574} & .046\tiny{-.140} \\
16, 16, 256 & .930\tiny{+.289} & 27.0\tiny{+12.4} & .054\tiny{-.581} & .045\tiny{-.141} \\
\hline
\multicolumn{5}{l}{\textit{Sequential-8: C-Res $\rightarrow$ Patch2Pixel-v2 ($p=3$)$\rightarrow$ G-Res}} \\
\hline
64, 64, 64 & .941\tiny{+.300} & 27.4\tiny{+12.8} & .089\tiny{-.546} & .043\tiny{-.143} \\
\hline
\multicolumn{5}{l}{\textit{Sequential-9: G-Res $\rightarrow$ C-Res (ablation)}} \\
\hline
64, 64 & .896\tiny{+.255} & 25.8\tiny{+11.2} & .172\tiny{-.463} & .051\tiny{-.135} \\
\hline
\multicolumn{5}{l}{\textit{Sequential-10: C-Res $\rightarrow$ G-Res (ablation)}} \\
\hline
64, 64 & .896\tiny{+.255} & 25.8\tiny{+11.2} & .172\tiny{-.463} & .051\tiny{-.135} \\
\hline
\multicolumn{5}{l}{\textit{Baseline Models}} \\
\hline
Unet++ - Small & .790\tiny{+.149} & 22.3\tiny{+7.7} & .149\tiny{-.486} & .077\tiny{-.109} \\
Unet++ - Medium & .831\tiny{+.190} & 24.5\tiny{+9.9} & .083\tiny{-.552} & .059\tiny{-.127} \\
Unet++ - Large & .868\tiny{+.227} & 25.8\tiny{+11.2} & .058\tiny{-.577} & .052\tiny{-.134} \\
CompenNet++ & .944\tiny{+.303} & 27.7\tiny{+13.1} & .033\tiny{-.602} & .041\tiny{-.145} \\
\hline
\end{tabular}
}
\vspace{0.2cm}

\end{table}

\section{Quantization Analysis}
\label{sec:supp-quant}
Table~\ref{tab:quant} reports the full quantization sweep results summarized in Section~\ref{sec:quant}. For each model, we sweep weight precision from 32 down to 4 bits to identify the minimum weight precision $N^*_w$ within ${\approx}0.1$~dB PSNR and $0.01$ SSIM of the FP32 baseline. Weights are then fixed at $N^*_w$ and activation precision is swept over the same 32--4 range, under the same criterion, to find $N^*_a$. Results confirm that the framework is well-suited to fixed-point hardware: 14-bit weights with 12-bit activations is near-lossless across all models (largest deviation $-0.13$\,dB PSNR), and Sequential-5$'$ tolerates 10-bit weights with 12-bit activations at less than $0.01$\,dB PSNR loss.

\begin{table*}[t]
\centering\tablestyle
\caption{Quantization Analysis Results (h=64)}
\label{tab:quant}
\fittable{%
\begin{tabular}{|lc|cc|ccc|cccc|}
\hline
\multirow{3}{*}{Model} & \multirow{3}{*}{Scenario} & \multicolumn{2}{c}{Baseline (FP32)} & \multicolumn{3}{c}{Weight-Only Quantization} & \multicolumn{3}{c}{Weight+Activation Quantization} \\
\cmidrule(lr){3-4} \cmidrule(lr){5-7} \cmidrule(lr){8-10}
& & \multirow{2}{*}{PSNR} & \multirow{2}{*}{SSIM} & \multirow{2}{*}{Bits} & \multirow{2}{*}{PSNR} & \multirow{2}{*}{SSIM} & \multirow{2}{*}{Bits (W+A)} & \multirow{2}{*}{PSNR} & \multirow{2}{*}{SSIM} \\
& & & &  & & &  & & \\
\hline
Patch2Pixel-v2 (p=3) & 1 & 27.69 & 0.8546 & 13b & 27.68 {\tiny(-0.01)} & 0.8539 {\tiny(-0.0007)} & 13b+12b & 27.65 {\tiny(-0.04)} & 0.8510 {\tiny(-0.0036)} \\
Patch2Pixel-v3 (p=3) &  1 & 27.52 & 0.8515 & 14b & 27.42 {\tiny(-0.10)} & 0.8515 {\tiny(+0.0000)} & 14b+11b & 27.39 {\tiny(-0.13)} & 0.8474 {\tiny(-0.0041)} \\
Patch2Pixel-v2 (p=5) &  2 & 23.26 & 0.8728 & 12b & 23.26 {\tiny(+0.00)} & 0.8725 {\tiny(-0.0003)} & 12b+11b & 23.24 {\tiny(-0.02)} & 0.8684 {\tiny(-0.0044)} \\
Patch2Pixel-v2 (p=3) &  2 & 23.05 & 0.8627 & 11b & 23.00 {\tiny(-0.05)} & 0.8626 {\tiny(-0.0001)} & 11b+11b & 22.98 {\tiny(-0.07)} & 0.8556 {\tiny(-0.0071)} \\
Patch2Pixel-v3 (p=3) &  2  & 22.99 & 0.8588 & 10b & 22.91 {\tiny(-0.08)} & 0.8580 {\tiny(-0.0008)} & 10b+9b & 22.88 {\tiny(-0.11)} & 0.8518 {\tiny(-0.0070)} \\
G-Res & 3 & 37.81 & 0.9870 & 11b & 37.74 {\tiny(-0.07)} & 0.9869 {\tiny(-0.0001)} & 11b+12b & 37.76 {\tiny(-0.05)} & 0.9869 {\tiny(-0.0001)} \\
G-Res &  4 & 35.24 & 0.9752 & 12b & 35.14 {\tiny(-0.10)} & 0.9749 {\tiny(-0.0003)} & 12b+12b & 35.16 {\tiny(-0.08)} & 0.9749 {\tiny(-0.0003)} \\
Sequential 5' (h=64,64) &  5 & 27.51 & 0.9419 & 10b & 27.48 {\tiny(-0.03)} & 0.9416 {\tiny(-0.0003)} & 10b+12b & 27.48 {\tiny(-0.00)} & 0.9415 {\tiny(-0.0004)} \\
sequential 7' (h=64,64,64) &  5 & 27.51 & 0.9411 & 11b & 27.51 {\tiny(-0.01)} & 0.9409 {\tiny(-0.0002)} & 11b+12b & 27.50 {\tiny(-0.01)} & 0.9406 {\tiny(-0.0002)} \\

\hline
\end{tabular}%
}
\vspace{0.5em}
\begin{flushleft}
\footnotesize
Weight-only quantization dequantizes weights to FP32 before computation (FP32 activations). Weight+activation quantization uses integer-only arithmetic. Bits format: W (weight bits) or W+A (weight + activation bits). Symmetric quantization with per-tensor scaling. Calibration: 10\% of test set.
\end{flushleft}
\end{table*}

%
%
\section{Video Workload Characterization}
\label{sec:supp-video}
This appendix provides the complete supporting material for the video
workload extension of Section~\ref{sec:video-extension}: pipeline and
resolution generalization, the runtime design space, dataset
construction, hardware platforms, the \emph{full} OpenMP calibration
(search ranges, workload, per-experiment winners), \emph{all ten}
nine-mode per-experiment result tables, the cross-cutting analyses
summarized in the main text, and threats to validity. All numbers are
taken from completed per-run \texttt{summary.csv} files; only
$\mathrm{fps}_{\mathrm{real}}{=}1000/(t_{\text{G-Res}}{+}
t_{\text{P2P}})$ and the quality columns are reported (harness
wall-clock / synthetic FPS are not deployment-relevant and are
omitted).

\begin{table}[t]
\centering\tablestyle
\caption{Design Space for Video Pre-distortion Exploration}
\label{tbl:modes}
\small
\begin{tabularx}{0.8\linewidth}{@{}lX@{}}
\toprule
\textbf{Mode} & \textbf{Description} \\
\midrule
\multicolumn{2}{@{}l}{\textbf{G-Res Operation Modes}} \\
\addlinespace[0.2em]

\texttt{mlp} &
Evaluates MLP on the full coordinate grid every frame. \\

\texttt{lut} &
Precomputes the fixed G-Res warp grid once for a given resolution and replaces the per-frame MLP with a bilinear resample of the stored grid. Since G-Res depends only on $(x,y)$, this is equivalent to \texttt{mlp} up to \texttt{float32} rounding. \\

\texttt{lutrevcache} &
Adds inter-frame reuse with a reverse LUT: for each input pixel, it stores the output pixels that read from it under the bilinear stencil. Only outputs whose source pixels changed since the previous frame are recomputed. \\

\midrule
\multicolumn{2}{@{}l}{\textbf{Patch2Pixel-v3 Operation Modes}} \\
\addlinespace[0.2em]

\texttt{mlp} &
Runs the per-pixel MLP on every output pixel of every frame, with no caching. \\

\texttt{3x3} &
Reuses the previous tonal output only when all nine pixels in the $3{\times}3$ dependency neighbourhood are unchanged.  \\

\texttt{center} &
Compares only the centre pixel of the dependency neighbourhood. This is cheaper than \texttt{3x3}, but may reuse a stale value if one of the eight neighbours changed. \\

\bottomrule
\end{tabularx}
\end{table}

\begin{table*}[t]
\centering\tablestyle
\caption{Cross-platform video workload summary for Sequential~5$'$. {Latencies are per-frame medians of the predistortion compute.} ``Best config'' minimizes $t_{\text{G-Res}}{+}t_{\text{P2P}}$.
}
\label{tab:video_xplatform}

\fittable{%
\begin{tabular}{llccl rrr rrr}
\toprule
Target & Compute & Dataset & $N$ & Best config
 & G-Res (ms) & P2P (ms) & Total (ms)
 & $\mathrm{fps}_{\mathrm{real}}$ & PSNR (dB) & SSIM \\
\midrule
Ryzen 9 5900X            & CPU & UGC & 256  & \texttt{lutrevcache}+\texttt{mlp} &  1.23 &  27.48 &   28.71 & 34.90 & 26.40 & 0.896 \\
Ryzen 9 5900X            & CPU & UVG & 1080 & \texttt{lut}+\texttt{mlp}         &  5.53 & 487.16 &  492.69 &  2.03 & 23.52 & 0.838 \\
RTX 3090                 & GPU & UGC & 360  & \texttt{lut}+\texttt{mlp}         &  1.15 &   1.03 &    2.18 & 530.37 & 25.44 & 0.877 \\
RTX 3090                 & GPU & UVG & 1080 & \texttt{lut}+\texttt{mlp}         &  6.06 &   7.96 &   14.02 & 72.20 & 23.52 & 0.838 \\
Raspberry Pi 5           & CPU & UGC & 256  & \texttt{lut}+\texttt{mlp}         &  2.43 &  83.52 &   85.95 & 11.65 & 26.40 & 0.896 \\
Raspberry Pi 5           & CPU & UVG & 1080 & \texttt{lut}+\texttt{mlp}         & 23.66 &1339.40 & 1363.06 &  0.73 & 23.51 & 0.838 \\
Jetson Orin Nano Super   & CPU & UGC & 256  & \texttt{lut}+\texttt{mlp}         &  2.85 &  80.26 &   83.11 & 12.04 & 26.40 & 0.896 \\
Jetson Orin Nano Super   & CPU & UVG & 1080 & \texttt{lut}+\texttt{mlp}         & 18.20 & 994.59 & 1012.79 &  0.99 & 23.51 & 0.838 \\
Jetson Orin Nano (iGPU)  & GPU & UGC & 360  & \texttt{lut}+\texttt{center}       &  4.55 &   9.03 &   13.58 & 82.29 & 25.45 & 0.877 \\
Jetson Orin Nano (iGPU)  & GPU & UVG & 1080 & \texttt{lut}+\texttt{mlp}         & 20.21 &  68.07 &   88.28 & 11.33 & 23.51 & 0.838 \\
\bottomrule
\end{tabular}}
\end{table*}

\subsection{Pipeline, Stages, and Resolution Generalization}
The video pipeline is the Sequential~5$'$ composition: a \textbf{G-Res}
geometric stage ($2{\to}64{\to}64{\to}2$ MLP with an identity
residual, $4{,}482$ parameters) producing a per-pixel warp grid
consumed by a bilinear resample (\texttt{border},
\texttt{align\_corners}); followed by a \textbf{Patch2Pixel-v3} tonal
stage ($29{\to}48{\to}48{\to}3$ MLP with a learned linear residual
projection, $4{,}029$ parameters). The $29$-D Patch2Pixel-v3 feature
is $(x,y)$ plus the centre RGB plus the $24$ RGB values of the eight
reflect-padded $3{\times}3$ neighbours ($P{=}3$); the output is
sigmoid-scaled to $[-2,2]$, added to the centre pixel, and clamped to
$[0,1]$. Total: $8{,}511$ \texttt{float32} parameters.

G-Res consumes only normalized coordinates, so its output is the
same warp field for every frame at a given resolution; this is what
makes the LUT substitution exact (to \texttt{float32} rounding) and
also lets the \emph{identical $256$-trained weights} run at $360$ and
$1080$ without retraining---G-Res is simply the same continuous
function $f:[0,1]^2{\to}[0,1]^2$ sampled on a denser grid, and the
Patch2Pixel-v3 neighbours are always the immediate physical
$3{\times}3$ pixels. Table~\ref{tab:supp-res} lists the
resolution-dependent quantities (all derived at startup from a single
\texttt{--input-size} flag); the ROI preserves $\approx$49\% pixel
coverage by construction. The measured PSNR cost of running outside
the training resolution is $\approx$1\,dB (UGC $256{\to}360$:
$26.40{\to}25.44$\,dB); UVG@1080 lands at $23.5$\,dB on every host.

\begin{table}[t]
\centering\tablestyle
\caption{Resolution-dependent quantities (one \texttt{--input-size}
knob propagates everywhere).}
\label{tab:supp-res}
\fittable{%
\begin{tabular}{lccc}
\toprule
Quantity & $N{=}256$ & $N{=}360$ & $N{=}1080$ \\
\midrule
Pixels per frame ($3N^2$) & 196{,}608 & 388{,}800 & 3{,}499{,}200 \\
ROI window & $179{\times}179$ & $252{\times}252$ & $755{\times}755$ \\
ROI / total pixels & 48.95\% & 49.00\% & 48.99\% \\
Forward-LUT size & 524\,KB & 1.0\,MB & 9.3\,MB \\
Reverse-LUT size & 1.3\,MB & 2.5\,MB & 23\,MB \\
Reverse-LUT edges ($4N^2$) & 262{,}144 & 518{,}400 & 4{,}665{,}600 \\
\bottomrule
\end{tabular}}
\end{table}

\subsection{Runtime Design Space}
The benchmark sweeps a $3{\times}3$ grid (Table~\ref{tab:supp-cfg}):
three G-Res modes $\times$ three Patch2Pixel-v3 cache modes,
identical on CPU and CUDA. \texttt{lut} replaces the G-Res MLP by a
precomputed warp grid (a single bilinear sample).
\texttt{lutrevcache} adds a CSR reverse-dependency graph so that only
output pixels whose source pixels changed are recomputed.
\texttt{center} reuses the previous Patch2Pixel-v3 output where the
centre pixel of the G-Res output is unchanged; \texttt{3x3} reuses
only where the full $3{\times}3$ G-Res-output neighbourhood (the
exact dependency set of Patch2Pixel-v3) is unchanged. The forward LUT
is the G-Res MLP evaluated offline on a uniform $N{\times}N$
coordinate grid; the reverse LUT is its bilinear-stencil transpose in
CSR form ($4N^2$ edges).

\begin{table}[t]
\centering\tablestyle
\caption{Nine video-runtime configurations (CPU and CUDA).}
\label{tab:supp-cfg}
\fittable{%
\begin{tabular}{cll}
\toprule
\# & G-Res mode & Patch2Pixel-v3 cache \\
\midrule
1 & \texttt{mlp}         & \texttt{none}   \\
2 & \texttt{mlp}         & \texttt{center} \\
3 & \texttt{mlp}         & \texttt{3x3}    \\
4 & \texttt{lut}         & \texttt{none}   \\
5 & \texttt{lut}         & \texttt{center} \\
6 & \texttt{lut}         & \texttt{3x3}    \\
7 & \texttt{lutrevcache} & \texttt{none}   \\
8 & \texttt{lutrevcache} & \texttt{center} \\
9 & \texttt{lutrevcache} & \texttt{3x3}    \\
\bottomrule
\end{tabular}}
\end{table}

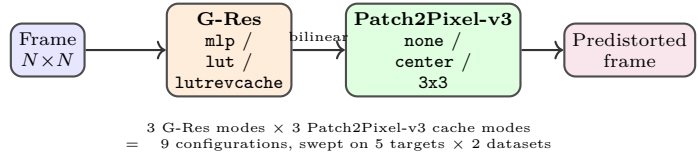
\begin{figure}[t]
\centering\tablestyle
\fittable{%
\begin{tikzpicture}[
  font=\scriptsize,
  box/.style={rectangle, rounded corners=1.5mm, draw=black!70,
    thick, align=center, minimum height=0.7cm, inner sep=3pt},
  arrow/.style={->, thick, draw=black!80}]
\node[box, fill=blue!10] (in)  at (0,0)  {Frame\\$N{\times}N$};
\node[box, fill=orange!14] (a1) at (2.4,0) {\textbf{G-Res}\\\texttt{mlp} /\\\texttt{lut} /\\\texttt{lutrevcache}};
\node[box, fill=green!12] (a12) at (5.1,0) {\textbf{Patch2Pixel-v3}\\\texttt{none} /\\\texttt{center} /\\\texttt{3x3}};
\node[box, fill=purple!10] (out) at (7.7,0) {Predistorted\\frame};
\draw[arrow] (in) -- (a1);
\draw[arrow] (a1) -- node[above,font=\tiny]{bilinear} (a12);
\draw[arrow] (a12) -- (out);
\node[font=\tiny, text width=8.4cm, align=center] at (3.85,-1.15)
 {$3$ G-Res modes $\times$ $3$ Patch2Pixel-v3 cache modes $=9$
  configurations, swept on $5$ targets $\times$ $2$ datasets};
\end{tikzpicture}}
\caption{Video runtime design space (self-contained schematic).}
\label{fig:supp-video-design}
\end{figure}

\subsection{Reverse-LUT Construction and Runtime}
\label{sec:supp-revlut}
We implement this functionality using a reverse lookup table.
Because every output pixel reads exactly four input pixels under the
bilinear stencil, the reverse LUT has a fixed total size of $4N^2$
input$\to$output edges (${\sim}262$k at $N{=}256$; ${\sim}4.66$M at
$N{=}1080$). We store
it as a compressed-sparse-row (CSR) array with an \texttt{offsets}
array of length $N^2{+}1$ and an \texttt{indices} array of length
$4N^2$. Per-input-pixel fan-in is heavy-tailed: at $N{=}256$ the
median is $2$ and the mean $4$, with a few corner pixels feeding
thousands of outputs. At runtime, the cache (i)~scans the input
frame to detect changed pixels (cost $\Theta(N^2)$), (ii)~walks the
CSR rows of the changed inputs to mark dependent outputs as dirty
(cost proportional to the number of dirty edges), and (iii)~recomputes only the dirty outputs from the forward LUT; the rest are read directly from the previous frame's cached G-Res output.
In hardware, the CSR walker maps naturally to a
sparse-indirect-address generator feeding a dirty-bit vector over
the output grid, with per-frame cost equal to the fixed change-
detection scan plus the dirty-output recompute. The structure
therefore pays off only when the dirty set is small relative to
$4N^2$, which is why \texttt{lutrevcache} is competitive on small
dependency graphs with high inter-frame coherence and regresses
otherwise.

\subsection{Datasets and Motion Stratification}
YouTube-UGC~\cite{wang2019youtubeugc}: the 360p H.264 CRF-10 subset,
$182$ clips across $12$ categories (HDR/VR/vertical excluded),
$20$\,s each ($\approx$600 frames at 30\,fps); per-category counts
are in Table~\ref{tab:supp-ugccat}. Six motion-stratified clips
(Table~\ref{tab:supp-ugc}) are selected by mean temporal
frame-change on the $256{\times}256$ grayscale stream; the selected
clips span a $1{,}295{\times}$ motion range. UVG
(\cite{mercat2020uvg}): $1920{\times}1080$ sources
transcoded once to H.264 CRF~18 and centre-cropped to
$1080{\times}1080$; two motion-stratified clips are used (HoneyBee,
mean change $9.60{\times}10^{-4}$; ReadySetGo,
$5.95{\times}10^{-2}$). Each UGC sweep is $6{\times}9{=}54$ rows;
each UVG sweep is $2{\times}9{=}18$ rows.

\begin{table}[t]
\centering\tablestyle
\caption{YouTube-UGC 360p subset by category.}
\label{tab:supp-ugccat}
\fittable{%
\begin{tabular}{lr}
\toprule
Category & Clips \\
\midrule
Sports & 24 \\ Lecture & 23 \\ NewsClip & 20 \\ HowTo & 20 \\
LiveMusic & 16 \\ Gaming & 16 \\ CoverSong & 16 \\ Animation & 14 \\
MusicVideo & 12 \\ Vlog & 10 \\ TelevisionClip & 6 \\ LyricVideo & 5 \\
\midrule
\textbf{Total} & \textbf{182} \\
\bottomrule
\end{tabular}}
\end{table}

\begin{table}[t]
\centering\tablestyle
\caption{Six motion-stratified YouTube-UGC benchmark clips.}
\label{tab:supp-ugc}
\fittable{%
\begin{tabular}{llcr}
\toprule
Group & Clip & Category & Mean change \\
\midrule
Smooth & Lecture\_360P-051a    & Lecture    & $1.157{\times}10^{-4}$ \\
Smooth & LyricVideo\_360P-5e87 & LyricVideo & $1.361{\times}10^{-4}$ \\
Medium & Gaming\_360P-586d     & Gaming     & $1.758{\times}10^{-2}$ \\
Medium & Vlog\_360P-2973       & Vlog       & $1.835{\times}10^{-2}$ \\
Fast   & Vlog\_360P-3121       & Vlog       & $1.386{\times}10^{-1}$ \\
Fast   & MusicVideo\_360P-24b0 & MusicVideo & $1.503{\times}10^{-1}$ \\
\bottomrule
\end{tabular}}
\end{table}

\subsection{Hardware Platforms}
\begin{table}[t]
\centering\tablestyle
\caption{Execution targets. OMP columns are the per-target,
per-dataset calibrated host thread counts (UGC\,/\,UVG).}
\label{tab:supp-hw}
\fittable{%
\begin{tabular}{lllcc}
\toprule
Host & Arch & CPU / GPU & OMP\,CPU & OMP\,GPU \\
\midrule
Ryzen 9 5900X & x86\_64 & 12C/24T + RTX 3090 (sm\_86) & 16 / 32 & 14 / 16 \\
Raspberry Pi 5 & aarch64 & 4$\times$ Cortex-A76 & 6 / 8 & -- \\
Jetson Orin Nano Super & aarch64 & 6$\times$ Cortex-A78AE + Ampere iGPU (sm\_87) & 28 / 18 & 8 / 10 \\
\bottomrule
\end{tabular}}
\end{table}
The Jetson is the replacement Orin Nano \emph{Super}
(MAXN\_SUPER 25\,W, JetPack 6.2.1 / L4T R36.4.7, CUDA 12.6). GPU
sweeps use a different host thread count than CPU sweeps because the
GPU-build timing window contains non-trivial host code (coordinate
packing, host/device copies, host bilinear resample).

\subsection{OpenMP Thread-Count Calibration}
The single swept runtime knob is \texttt{OMP\_NUM\_THREADS}; it
governs every \texttt{\#pragma omp parallel for} region (the bilinear
warp application, the two Patch2Pixel-v3 host loops, and the
activation/linear kernels) and thus directly moves $t_{\text{G-Res}}$
and $t_{\text{P2P}}$. Before each main run, every device sweeps its
candidate range (Table~\ref{tab:supp-omprange}) with the dominant
configuration (\texttt{lut}+\texttt{none}) over a short calibration
workload (Table~\ref{tab:supp-ompwork}) and picks the value
minimising $t_{\text{G-Res}}{+}t_{\text{P2P}}$, separately per
(machine, dataset, kind); ties break toward the lower thread count.
The chosen value transfers well to the other eight modes because
Patch2Pixel-v3 dominates runtime in all of them. GPU sensitivity is
real (RTX~3090: $+57\%$ UGC, $+79\%$ UVG between best and worst host
thread count) but lower on the iGPU ($+15.6\%$ UGC, $+28.4\%$ UVG),
where the kernel itself is the bottleneck.
Table~\ref{tab:supp-omp} lists the ten per-experiment winners.

\begin{table}[t]
\centering\tablestyle
\caption{OpenMP search ranges per machine.}
\label{tab:supp-omprange}
\fittable{%
\begin{tabular}{lll}
\toprule
Machine & CPU candidates & GPU host candidates \\
\midrule
Ryzen 9 5900X & $2,4,\dots,32$ (16 values) & $2,4,\dots,32$ (16) \\
Raspberry Pi 5 & $1,2,3,4,5,6,8,10$ (8) & --- \\
Jetson Orin Nano Super & $2,4,\dots,28$ (14, extended) & $2,4,\dots,18$ (9) \\
\bottomrule
\end{tabular}}
\end{table}

\begin{table}[t]
\centering\tablestyle
\caption{Calibration workload (videos $\times$ frames-per-video).}
\label{tab:supp-ompwork}
\fittable{%
\begin{tabular}{lllcc}
\toprule
Machine & Dataset & Res & \#vids & Frames/vid \\
\midrule
Ryzen 9 5900X & UGC & 256/360 & 6 & 40 \\
Ryzen 9 5900X & UVG & 1080 & 2 & 15 \\
Raspberry Pi 5 & UGC & 256 & 6 & 25 \\
Raspberry Pi 5 & UVG & 1080 & 2 & 6 \\
Jetson Orin Nano Super & UGC & 256/360 & 6 & 40 \\
Jetson Orin Nano Super & UVG & 1080 & 2 & 15 \\
\bottomrule
\end{tabular}}
\end{table}


\begin{table}[t]
\centering\tablestyle
\caption{Per-experiment OpenMP winners. ``Margin'' is the gap to the
next-best thread count. }
\label{tab:supp-omp}
\fittable{%
\begin{tabular}{clllcl}
\toprule
Exp & Host & Dataset & Kind & OMP & Margin \\
\midrule
1  & Ryzen 5900X            & UGC & CPU & 16 & $+3.50\%$ \\
2  & Raspberry Pi 5         & UGC & CPU &  6 & $+2.16\%$ \\
3  & Jetson Orin Nano Super & UGC & CPU & 28 & $+0.33\%$ (boundary) \\
4  & Ryzen 5900X / RTX 3090 & UGC & GPU & 14 & $+2.31\%$ \\
5  & Jetson iGPU            & UGC & GPU &  8 & $+1.68\%$ (noise zone) \\
6  & Ryzen 5900X            & UVG & CPU & 32 & $+0.37\%$ (boundary) \\
7  & Raspberry Pi 5         & UVG & CPU &  8 & $+0.42\%$ \\
8  & Jetson Orin Nano Super & UVG & CPU & 18 & $+0.14\%$ (true optimum) \\
9  & Ryzen 5900X / RTX 3090 & UVG & GPU & 16 & $+2.33\%$ \\
10 & Jetson iGPU            & UVG & GPU & 10 & $+1.00\%$ \\
\bottomrule
\end{tabular}}
\end{table}

\subsection{Full Per-Experiment Results}
Tables~\ref{tab:supp-e1}--\ref{tab:supp-e10} give all nine modes for each of the ten experiments (means over the dataset's videos; G-Res/P2P are means of per-video medians; ``Dist'' is the
benchmark-only optical-distortion stage, excluded from $\mathrm{fps}_{\mathrm{real}}$). \mbox{P2P\,$=$\,Patch2Pixel-v3};
``G''/``P''/``D'' are G-Res/Patch2Pixel-v3/Distortion latency in ms.

\begin{table}[t]
\centering\tablestyle\caption{Exp~1 --- Ryzen~9 5900X CPU, UGC @256 (OMP\,16).}
\label{tab:supp-e1}
\fittable{%
\begin{tabular}{llrrrrrr}
\toprule
G-Res & P2P & G ms & P ms & D ms & fps & PSNR & SSIM \\
\midrule
\texttt{mlp}&\texttt{none}&31.44&28.41&30.93&16.73&26.4007&0.8960\\
\texttt{mlp}&\texttt{center}&31.79&28.27&31.27&16.68&26.4004&0.8960\\
\texttt{mlp}&\texttt{3x3}&31.87&29.01&30.88&16.44&26.4007&0.8960\\
\texttt{lut}&\texttt{none}&2.45&27.96&30.84&33.06&26.4007&0.8960\\
\texttt{lut}&\texttt{center}&2.57&27.54&30.68&33.35&26.4004&0.8960\\
\texttt{lut}&\texttt{3x3}&2.52&28.23&30.77&32.65&26.4007&0.8960\\
\texttt{lutrevcache}&\texttt{none}&1.23&27.48&30.84&34.90&26.3996&0.8960\\
\texttt{lutrevcache}&\texttt{center}&1.23&27.82&30.71&34.51&26.3962&0.8959\\
\texttt{lutrevcache}&\texttt{3x3}&1.24&27.64&30.61&34.69&26.3996&0.8960\\
\bottomrule
\end{tabular}}
\end{table}

\begin{table}[t]
\centering\tablestyle
\caption{Exp~2 --- Raspberry~Pi~5 CPU, UGC @256 (OMP\,6).}
\label{tab:supp-e2}
\fittable{%
\begin{tabular}{llrrrrrr}
\toprule
G-Res & P2P & G ms & P ms & D ms & fps & PSNR & SSIM \\
\midrule
\texttt{mlp}&\texttt{none}&94.48&84.63&52.29&5.58&26.4007&0.8960\\
\texttt{mlp}&\texttt{center}&94.59&84.69&52.20&5.58&26.4004&0.8960\\
\texttt{mlp}&\texttt{3x3}&94.80&85.68&52.31&5.54&26.4007&0.8960\\
\texttt{lut}&\texttt{none}&2.43&83.52&52.10&11.65&26.4007&0.8960\\
\texttt{lut}&\texttt{center}&2.38&84.46&52.04&11.53&26.4004&0.8960\\
\texttt{lut}&\texttt{3x3}&2.47&85.42&52.03&11.39&26.4007&0.8960\\
\texttt{lutrevcache}&\texttt{none}&2.57&84.41&52.15&11.51&26.3996&0.8960\\
\texttt{lutrevcache}&\texttt{center}&2.47&85.29&52.29&11.41&26.3962&0.8959\\
\texttt{lutrevcache}&\texttt{3x3}&2.42&86.71&52.31&11.23&26.3996&0.8960\\
\bottomrule
\end{tabular}}
\end{table}

\begin{table}[t]
\centering\tablestyle\caption{Exp~3 --- Jetson Orin Nano Super CPU, UGC @256
(OMP\,28; boundary winner).}
\label{tab:supp-e3}
\fittable{%
\begin{tabular}{llrrrrrr}
\toprule
G-Res & P2P & G ms & P ms & D ms & fps & PSNR & SSIM \\
\midrule
\texttt{mlp}&\texttt{none}&90.23&77.68&70.03&5.96&26.4007&0.8960\\
\texttt{mlp}&\texttt{center}&89.15&75.89&69.36&6.06&26.4004&0.8960\\
\texttt{mlp}&\texttt{3x3}&89.91&78.03&69.52&5.96&26.4007&0.8960\\
\texttt{lut}&\texttt{none}&2.85&80.26&69.10&12.04&26.4007&0.8960\\
\texttt{lut}&\texttt{center}&2.99&82.46&68.62&11.71&26.4004&0.8960\\
\texttt{lut}&\texttt{3x3}&3.15&82.90&67.76&11.62&26.4007&0.8960\\
\texttt{lutrevcache}&\texttt{none}&2.59&82.64&68.66&11.75&26.3996&0.8960\\
\texttt{lutrevcache}&\texttt{center}&2.50&82.73&68.01&11.74&26.3962&0.8959\\
\texttt{lutrevcache}&\texttt{3x3}&2.48&84.77&68.09&11.47&26.3996&0.8960\\
\bottomrule
\end{tabular}}
\end{table}

\begin{table}[t]
\centering\tablestyle\caption{Exp~4 --- RTX~3090 GPU, UGC @360
(host OMP\,14).}\label{tab:supp-e4}
\fittable{%
\begin{tabular}{llrrrrrr}
\toprule
G-Res & P2P & G ms & P ms & D ms & fps & PSNR & SSIM \\
\midrule
\texttt{mlp}&\texttt{none}&2.21&1.01&56.26&335.99&25.4365&0.8774\\
\texttt{mlp}&\texttt{center}&2.22&1.74&56.02&268.09&25.4361&0.8774\\
\texttt{mlp}&\texttt{3x3}&2.23&2.00&56.11&247.75&25.4365&0.8774\\
\texttt{lut}&\texttt{none}&1.15&1.03&56.97&530.37&25.4365&0.8774\\
\texttt{lut}&\texttt{center}&1.14&1.77&56.79&378.99&25.4361&0.8774\\
\texttt{lut}&\texttt{3x3}&1.15&2.03&57.09&338.38&25.4365&0.8774\\
\texttt{lutrevcache}&\texttt{none}&2.22&1.07&54.05&371.91&25.4354&0.8773\\
\texttt{lutrevcache}&\texttt{center}&2.22&1.81&53.36&288.73&25.4332&0.8772\\
\texttt{lutrevcache}&\texttt{3x3}&2.23&2.10&54.01&261.50&25.4354&0.8773\\
\bottomrule
\end{tabular}}
\end{table}

\begin{table}[t]
\centering\tablestyle\caption{Exp~5 --- Jetson Orin Nano iGPU, UGC @360
(host OMP\,8).}\label{tab:supp-e5}
\fittable{%
\begin{tabular}{llrrrrrr}
\toprule
G-Res & P2P & G ms & P ms & D ms & fps & PSNR & SSIM \\
\midrule
\texttt{mlp}&\texttt{none}&18.75&10.35&122.02&34.37&25.4550&0.8773\\
\texttt{mlp}&\texttt{center}&18.79&8.35&121.90&37.59&25.4547&0.8773\\
\texttt{mlp}&\texttt{3x3}&18.86&12.27&122.96&32.72&25.4550&0.8773\\
\texttt{lut}&\texttt{none}&3.94&10.78&119.76&68.02&25.4550&0.8773\\
\texttt{lut}&\texttt{center}&4.55&9.03&120.69&82.29&25.4547&0.8773\\
\texttt{lut}&\texttt{3x3}&4.23&13.15&120.53&61.87&25.4550&0.8773\\
\texttt{lutrevcache}&\texttt{none}&5.32&10.80&119.79&63.37&25.4541&0.8773\\
\texttt{lutrevcache}&\texttt{center}&5.50&8.94&120.92&84.44&25.4524&0.8772\\
\texttt{lutrevcache}&\texttt{3x3}&5.46&13.49&120.95&61.32&25.4541&0.8773\\
\bottomrule
\end{tabular}}
\end{table}

\begin{table}[t]
\centering\tablestyle\caption{Exp~6 --- Ryzen~9 5900X CPU, UVG @1080
(OMP\,32; boundary winner).}\label{tab:supp-e6}
\fittable{%
\begin{tabular}{llrrrrrr}
\toprule
G-Res & P2P & G ms & P ms & D ms & fps & PSNR & SSIM \\
\midrule
\texttt{mlp}&\texttt{none}&574.08&484.67&433.96&0.94&23.5234&0.8380\\
\texttt{mlp}&\texttt{center}&575.34&497.90&447.88&0.93&23.5230&0.8380\\
\texttt{mlp}&\texttt{3x3}&576.15&504.89&447.88&0.93&23.5234&0.8380\\
\texttt{lut}&\texttt{none}&5.53&487.16&433.48&2.03&23.5234&0.8380\\
\texttt{lut}&\texttt{center}&5.46&500.15&450.39&1.98&23.5230&0.8380\\
\texttt{lut}&\texttt{3x3}&5.49&506.60&450.41&1.95&23.5234&0.8380\\
\texttt{lutrevcache}&\texttt{none}&28.77&486.65&434.14&1.94&23.5195&0.8378\\
\texttt{lutrevcache}&\texttt{center}&28.75&499.42&448.49&1.89&23.5208&0.8379\\
\texttt{lutrevcache}&\texttt{3x3}&28.68&506.64&447.99&1.87&23.5195&0.8378\\
\bottomrule
\end{tabular}}
\end{table}

\begin{table}[t]
\centering\tablestyle\caption{Exp~7 --- Raspberry~Pi~5 CPU, UVG @1080
(OMP\,8).}\label{tab:supp-e7}
\fittable{%
\begin{tabular}{llrrrrrr}
\toprule
G-Res & P2P & G ms & P ms & D ms & fps & PSNR & SSIM \\
\midrule
\texttt{mlp}&\texttt{none}&1526.53&1330.00&988.59&0.35&23.5135&0.8377\\
\texttt{mlp}&\texttt{center}&1527.62&1355.36&993.22&0.35&23.5131&0.8377\\
\texttt{mlp}&\texttt{3x3}&1524.51&1368.42&993.05&0.35&23.5135&0.8377\\
\texttt{lut}&\texttt{none}&23.66&1339.40&996.68&0.73&23.5135&0.8377\\
\texttt{lut}&\texttt{center}&23.43&1357.55&989.94&0.72&23.5131&0.8377\\
\texttt{lut}&\texttt{3x3}&23.15&1369.39&996.95&0.72&23.5135&0.8377\\
\texttt{lutrevcache}&\texttt{none}&53.91&1335.28&986.10&0.72&23.5123&0.8377\\
\texttt{lutrevcache}&\texttt{center}&55.30&1355.27&990.21&0.71&23.5125&0.8377\\
\texttt{lutrevcache}&\texttt{3x3}&54.51&1370.43&989.88&0.70&23.5123&0.8377\\
\bottomrule
\end{tabular}}
\end{table}

\begin{table}[t]
\centering\tablestyle\caption{Exp~8 --- Jetson Orin Nano Super CPU, UVG @1080
(OMP\,18; true optimum after range extension).}\label{tab:supp-e8}
\fittable{%
\begin{tabular}{llrrrrrr}
\toprule
G-Res & P2P & G ms & P ms & D ms & fps & PSNR & SSIM \\
\midrule
\texttt{mlp}&\texttt{none}&1116.79&984.18&1045.85&0.48&23.5135&0.8377\\
\texttt{mlp}&\texttt{center}&1115.58&1005.78&1060.20&0.47&23.5131&0.8377\\
\texttt{mlp}&\texttt{3x3}&1118.95&1019.58&1059.39&0.47&23.5135&0.8377\\
\texttt{lut}&\texttt{none}&18.20&994.59&1049.85&0.99&23.5135&0.8377\\
\texttt{lut}&\texttt{center}&17.41&1013.25&1062.38&0.97&23.5131&0.8377\\
\texttt{lut}&\texttt{3x3}&17.73&1030.74&1063.34&0.95&23.5135&0.8377\\
\texttt{lutrevcache}&\texttt{none}&57.02&996.84&1047.30&0.95&23.5123&0.8377\\
\texttt{lutrevcache}&\texttt{center}&57.08&1015.39&1058.76&0.93&23.5125&0.8377\\
\texttt{lutrevcache}&\texttt{3x3}&56.86&1034.09&1059.05&0.92&23.5123&0.8377\\
\bottomrule
\end{tabular}}
\end{table}

\begin{table}[t]
\centering\tablestyle\caption{Exp~9 --- RTX~3090 GPU, UVG @1080
(host OMP\,16).}\label{tab:supp-e9}
\fittable{%
\begin{tabular}{llrrrrrr}
\toprule
G-Res & P2P & G ms & P ms & D ms & fps & PSNR & SSIM \\
\midrule
\texttt{mlp}&\texttt{none}&13.79&7.87&429.52&46.17&23.5234&0.8380\\
\texttt{mlp}&\texttt{center}&13.90&14.89&426.61&34.73&23.5230&0.8380\\
\texttt{mlp}&\texttt{3x3}&13.82&17.86&425.94&31.71&23.5234&0.8380\\
\texttt{lut}&\texttt{none}&6.06&7.96&441.54&72.20&23.5234&0.8380\\
\texttt{lut}&\texttt{center}&6.15&15.27&431.91&46.88&23.5230&0.8380\\
\texttt{lut}&\texttt{3x3}&6.13&17.95&433.75&41.57&23.5234&0.8380\\
\texttt{lutrevcache}&\texttt{none}&29.05&8.13&427.18&27.81&23.5195&0.8378\\
\texttt{lutrevcache}&\texttt{center}&28.96&14.97&426.78&23.33&23.5208&0.8379\\
\texttt{lutrevcache}&\texttt{3x3}&28.95&18.66&426.93&21.18&23.5195&0.8378\\
\bottomrule
\end{tabular}}
\end{table}

\begin{table}[t]
\centering\tablestyle\caption{Exp~10 --- Jetson Orin Nano iGPU, UVG @1080
(host OMP\,10).}\label{tab:supp-e10}
\fittable{%
\begin{tabular}{llrrrrrr}
\toprule
G-Res & P2P & G ms & P ms & D ms & fps & PSNR & SSIM \\
\midrule
\texttt{mlp}&\texttt{none}&113.09&72.65&1089.33&5.38&23.5135&0.8377\\
\texttt{mlp}&\texttt{center}&112.91&96.66&1084.57&4.77&23.5131&0.8377\\
\texttt{mlp}&\texttt{3x3}&113.31&140.83&1093.41&4.02&23.5135&0.8377\\
\texttt{lut}&\texttt{none}&20.21&68.07&1066.68&11.33&23.5135&0.8377\\
\texttt{lut}&\texttt{center}&21.84&83.52&1072.96&9.49&23.5131&0.8377\\
\texttt{lut}&\texttt{3x3}&22.45&125.18&1074.88&7.11&23.5135&0.8377\\
\texttt{lutrevcache}&\texttt{none}&58.94&71.15&1065.20&7.71&23.5123&0.8377\\
\texttt{lutrevcache}&\texttt{center}&58.98&82.34&1072.04&7.09&23.5125&0.8377\\
\texttt{lutrevcache}&\texttt{3x3}&59.34&132.13&1072.40&5.32&23.5123&0.8377\\
\bottomrule
\end{tabular}}
\end{table}

\subsection{Cross-Cutting Analyses}

\paragraph{G-Res mode vs.\ latency.}
Table~\ref{tab:supp-a1} reports G-Res latency by mode and the
\texttt{mlp}$\to$\texttt{lut} speedup. The LUT is bit-equivalent to
the MLP (PSNR/SSIM stable to $<0.01$\,dB / $10^{-4}$ across modes);
the small \texttt{lutrevcache} PSNR drop (e.g.\ $26.3996$ vs.\
$26.4007$ in Exp~1) is the rounding cost of carrying cached
\texttt{float32} outputs across frames.

\begin{table}[t]
\centering\tablestyle\caption{G-Res latency by mode and \texttt{mlp}$\to$%
\texttt{lut} speedup.}\label{tab:supp-a1}
\fittable{%
\begin{tabular}{lrrrr}
\toprule
Target / dataset & mlp ms & lut ms & rev ms & mlp$\to$lut \\
\midrule
Ryzen / UGC256 (CPU)        &  31.44 &  2.45 &  1.23 & $12.8{\times}$ \\
Pi 5 / UGC256 (CPU)         &  94.48 &  2.43 &  2.57 & $38.9{\times}$ \\
Jetson / UGC256 (CPU)       &  90.23 &  2.85 &  2.59 & $31.7{\times}$ \\
Ryzen / UVG1080 (CPU)       & 574.08 &  5.53 & 28.77 & $103.8{\times}$ \\
Pi 5 / UVG1080 (CPU)        &1526.53 & 23.66 & 53.91 & $64.5{\times}$ \\
Jetson / UVG1080 (CPU)      &1116.79 & 18.20 & 57.02 & $61.4{\times}$ \\
RTX 3090 / UGC360 (GPU)     &   2.21 &  1.15 &  2.22 & $1.92{\times}$ \\
Jetson iGPU / UGC360 (GPU)  &  18.75 &  3.94 &  5.32 & $4.76{\times}$ \\
RTX 3090 / UVG1080 (GPU)    &  13.79 &  6.06 & 29.05 & $2.28{\times}$ \\
Jetson iGPU / UVG1080 (GPU) & 113.09 & 20.21 & 58.94 & $5.60{\times}$ \\
\bottomrule
\end{tabular}}
\end{table}

\paragraph{G-Res vs.\ Patch2Pixel-v3 share.}
After the LUT, Patch2Pixel-v3 dominates: $95$--$99\%$ of model
compute on every CPU sweep, $47$--$57\%$ on the dGPU
(Table~\ref{tab:supp-share}).

\begin{table}[t]
\centering\tablestyle\caption{Best G-Res latency vs.\ Patch2Pixel-v3
(\texttt{none}); $\mathrm{P2P}/(\mathrm{G}{+}\mathrm{P2P})$.}
\label{tab:supp-share}
\fittable{%
\begin{tabular}{lrrr}
\toprule
Target / dataset & best G ms & P2P ms & P2P share \\
\midrule
Ryzen / UGC256 (CPU)    &  1.23 &  27.48 & 95.7\% \\
Pi 5 / UGC256 (CPU)     &  2.43 &  83.52 & 97.2\% \\
Ryzen / UVG1080 (CPU)   &  5.53 & 487.16 & 98.9\% \\
Pi 5 / UVG1080 (CPU)    & 23.66 &1339.40 & 98.3\% \\
RTX 3090 / UGC360 (GPU) &  1.15 &   1.03 & 47.3\% \\
RTX 3090 / UVG1080 (GPU)&  6.06 &   7.96 & 56.8\% \\
\bottomrule
\end{tabular}}
\end{table}

\paragraph{Reverse-cache vs.\ motion.}
The reverse cache helps only when consecutive frames share many
unchanged input pixels. Per-video G-Res latency (\texttt{lut} vs.\
\texttt{lutrevcache}, Patch2Pixel-v3 \texttt{none};
Table~\ref{tab:supp-motion}) shows it wins on all $256^2$ x86 clips,
regresses on higher-motion clips on the slower Pi, and is
catastrophic at $1080^2$ on every platform.

\begin{table}[t]
\centering\tablestyle\caption{Reverse-cache vs.\ motion: per-video G-Res ms,
\texttt{lut}$\to$\texttt{rev} ($\Delta$). ``sm''/``fa'' $=$
smoothest/fastest clip.}\label{tab:supp-motion}
\fittable{%
\begin{tabular}{lrrrrrr}
\toprule
Target / dataset & sm lut & sm rev & $\Delta$ & fa lut & fa rev & $\Delta$ \\
\midrule
Ryzen / UGC256   & 0.83 & 0.47 & $-43\%$ & 3.03 & 2.34 & $-23\%$ \\
Pi 5 / UGC256    & 1.68 & 1.48 & $-12\%$ & 2.53 & 3.78 & $+49\%$ \\
Jetson / UGC256  & 2.00 & 1.00 & $-50\%$ & 3.30 & 4.11 & $+25\%$ \\
Ryzen / UVG1080  & 5.30 & 22.12 & $+317\%$ & 5.76 & 35.42 & $+515\%$ \\
Pi 5 / UVG1080   & 21.42 & 48.00 & $+124\%$ & 25.91 & 59.82 & $+131\%$ \\
Jetson / UVG1080 & 17.56 & 48.85 & $+178\%$ & 18.84 & 65.18 & $+246\%$ \\
\bottomrule
\end{tabular}}
\end{table}

\paragraph{Patch2Pixel-v3 cache vs.\ hardware.}
Table~\ref{tab:supp-p2pcache} isolates the cache effect
(G-Res\,$=$\,\texttt{lut}). It is within noise / a slight cost on
every CPU and on the dGPU it always slows the tonal stage down. The
only host where a cache nets positive is the bandwidth-limited Jetson
iGPU at UGC-360, where \texttt{center} beats \texttt{none}
($9.03$ vs.\ $10.78$\,ms); at UVG-1080 even the iGPU reverts to the
dGPU pattern. The cached CUDA kernel still launches one thread per
pixel, so unchanged pixels do not free GPU work---they only avoid the
Patch2Pixel-v3 MLP; on the high-bandwidth dGPU the per-pixel test
plus \texttt{atomicAdd} exceeds the saved MLP, while on the
low-bandwidth iGPU the avoided evaluations net positive at the small
image size.

\begin{table}[t]
\centering\tablestyle\caption{Patch2Pixel-v3 ms by cache mode (G-Res\,$=$\,%
\texttt{lut}).}\label{tab:supp-p2pcache}
\fittable{%
\begin{tabular}{lrrrl}
\toprule
Target / dataset & none & center & 3x3 & verdict \\
\midrule
Ryzen / UGC256 (CPU)        &  27.96 &  27.54 &  28.23 & noise \\
Pi 5 / UGC256 (CPU)         &  83.52 &  84.46 &  85.42 & slight cost \\
Ryzen / UVG1080 (CPU)       & 487.16 & 500.15 & 506.60 & slight cost \\
Pi 5 / UVG1080 (CPU)        &1339.40 &1357.55 &1369.39 & slight cost \\
RTX 3090 / UGC360 (GPU)     &   1.03 &   1.77 &   2.03 & cache slower \\
RTX 3090 / UVG1080 (GPU)    &   7.96 &  15.27 &  17.95 & cache slower \\
Jetson iGPU / UGC360 (GPU)  &  10.78 &   9.03 &  13.15 & \textbf{center wins} \\
Jetson iGPU / UVG1080 (GPU) &  68.07 &  83.52 & 125.18 & cache slower \\
\bottomrule
\end{tabular}}
\end{table}

\paragraph{GPU model-only speedup vs.\ same-host CPU.}
Table~\ref{tab:supp-gpu} contrasts the model-only GPU latency
(\texttt{lut}+\texttt{none}) with the same model on the same host's
CPU. At UVG-1080 the dGPU is $35.1{\times}$ faster than its CPU and
the edge iGPU is $11.5{\times}$ faster than its CPU; the UGC GPU rows
are cross-resolution (GPU @360 vs.\ CPU @256) and are not directly
comparable.

\begin{table}[t]
\centering\tablestyle\caption{GPU model-only (\texttt{lut}+\texttt{none}) vs.\
same-host CPU (\texttt{lut}+\texttt{none}).}\label{tab:supp-gpu}
\fittable{%
\begin{tabular}{lrrr}
\toprule
Platform / dataset & GPU G+P2P ms & CPU G+P2P ms & speedup \\
\midrule
RTX 3090 / UVG1080      & 14.02 &  492.69 & $35.1{\times}$ \\
Jetson iGPU / UVG1080   & 88.28 & 1012.79 & $11.5{\times}$ \\
RTX 3090 / UGC360       &  2.18 & \multicolumn{2}{c}{cross-resolution (CPU @256)} \\
Jetson iGPU / UGC360    & 14.72 & \multicolumn{2}{c}{cross-resolution (CPU @256)} \\
\bottomrule
\end{tabular}}
\end{table}

\paragraph{Quality vs.\ platform and resolution.}
PSNR/SSIM are model-determined, not platform-determined
(Table~\ref{tab:supp-quality}): within each (dataset, resolution)
cell all hosts agree to $\approx$0.01\,dB / $10^{-4}$, confirming
numerical portability across x86, ARM, the CUDA dGPU, and the
embedded iGPU.

\begin{table}[t]
\centering\tablestyle\caption{Quality by dataset/resolution cell.}
\label{tab:supp-quality}
\fittable{%
\begin{tabular}{llrr}
\toprule
Cell & Hosts & PSNR (dB) & SSIM \\
\midrule
UGC @256 (CPU)      & Exps 1,2,3   & $26.40\pm0.001$ & $0.8960\pm10^{-4}$ \\
UGC @360 (GPU)      & Exps 4,5     & $25.44$ / $25.45$ & $0.8774$ / $0.8773$ \\
UVG @1080 (CPU+GPU) & Exps 6--10   & $23.52$ / $23.51$ & $0.838$ \\
\bottomrule
\end{tabular}}
\end{table}


\subsection{Reproducibility}
The engine is a self-contained C++17 implementation (CMake $\geq$3.14,
\texttt{-O3} release, OpenMP auto-detected; optional CUDA with
\texttt{sm\_86;sm\_87}); a custom \texttt{float32} tensor engine
(no BLAS/OpenCV/cuDNN); \texttt{ffmpeg} via \texttt{popen()}.
Per-frame latency uses \texttt{std::chrono::steady\_clock} medians;
PSNR uses joint channel/spatial MSE, SSIM uses an $11{\times}11$
Gaussian window ($\sigma{=}1.5$, $C_1{=}10^{-4}$, $C_2{=}9{\times}
10^{-4}$). Each of the ten experiments writes a per-run
\texttt{summary.csv} ($54$ rows for UGC, $18$ for UVG), aggregated
by an analysis script into the tables above; OpenMP thread counts are
fixed by the per-host calibration script preceding each run.

\section{Hardware Model Details}
\label{sec:supp-hardware}

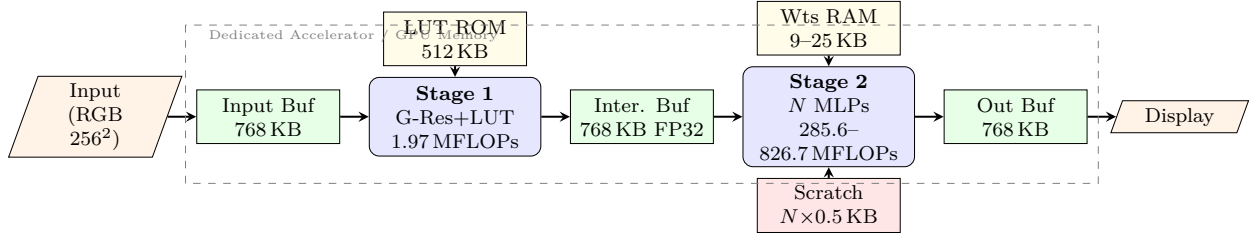
\begin{figure*}[t]
\centering\tablestyle
\fittable{
\begin{tikzpicture}[
    node distance=0.6cm,
    block/.style={rectangle,draw,fill=blue!10,text width=2.2cm,
                  text centered,minimum height=1.1cm,rounded corners},
    memory/.style={rectangle,draw,fill=green!10,text width=1.8cm,
                   text centered,minimum height=0.75cm},
    io/.style={trapezium,trapezium left angle=70,
               trapezium right angle=110,draw,fill=orange!10,
               text width=1.4cm,text centered},
    arrow/.style={->,thick,>=stealth}]
\node[io] (cam) {Input\\(RGB $256^2$)};
\node[memory, right=0.4cm of cam]  (b1) {Input Buf\\768\,KB};
\node[block,  right=0.4cm of b1]   (g)
    {\textbf{Stage 1}\\G-Res+LUT\\1.97\,MFLOPs};
\node[memory, above=0.15cm of g, fill=yellow!10] (lut) {LUT ROM\\512\,KB};
\node[memory, right=0.4cm of g]    (b2) {Inter. Buf\\768\,KB FP32};
\node[block,  right=0.4cm of b2]   (p)
    {\textbf{Stage 2}\\$N$ MLPs\\285.6--826.7\,MFLOPs};
\node[memory, above=0.15cm of p, fill=yellow!10] (w) {Wts RAM\\9--25\,KB};
\node[memory, below=0.15cm of p, fill=red!10]    (s)
    {Scratch\\$N{\times}$0.5\,KB};
\node[memory, right=0.4cm of p]    (b3) {Out Buf\\768\,KB};
\node[io, right=0.4cm of b3] (disp) {Display};
\draw[arrow](cam)--(b1); \draw[arrow](b1)--(g);
\draw[arrow](lut)--(g);  \draw[arrow](g)--(b2);
\draw[arrow](b2)--(p);   \draw[arrow](w)--(p);
\draw[arrow](s)--(p);    \draw[arrow](p)--(b3);
\draw[arrow](b3)--(disp);
\draw[dashed,gray]
  ($(b1.north west)+(-0.15,0.9)$) rectangle ($(b3.south east)+(0.15,-0.55)$);
\node[gray,font=\tiny] at ($(b1.north west)+(2.2,0.75)$)
  {Dedicated Accelerator / GPU Memory};
\end{tikzpicture}}
\caption{Sequential~5$'$ dataflow. Stage~1 is a LUT lookup plus bilinear interpolation; Stage~2 runs $N$ parallel Patch2Pixel MLPs. All buffers and weights are FP32.}
\label{fig:data_flow}
\end{figure*}

\begin{table}[t]
\centering\tablestyle
\caption{Per-frame resource requirements for Sequential~5$'$. All quantities assume FP32; no quantization is applied (cf.\ Sec.~\ref{sec:quant}). Sizes use 1~KB${=}1024$~bytes. ``Transfer (Reads)'' counts read traffic only and excludes the intermediate- and output-frame writes ($2\times768$~KB), unlike the Transfer column of Table~\ref{tab:model_summary}, which is reads plus writes.}
\label{tab:resource_per_stage}
\fittable{%
\begin{tabular}{l c c c}
\toprule
Metric & Stage~1 (G-Res) & Stage~2 (Patch2Pixel) & Total \\
       &                 & $h=32 \quad h=64$     & $h=32$ / $h=64$ \\
\midrule
Compute (FLOPs) & 1.97 M & 285.6 M \quad 826.8 M & 287.6 M / 828.8 M \\
Static (ROM)    & 512 KB & 8.6 KB \quad 24.9 KB   & 521 KB / 537 KB \\
Dynamic (RAM)   & 1{,}536 KB & 1{,}536 KB \quad 1{,}536 KB   & 2{,}304 KB + scratch \\
{Transfer (Reads)}      & 3{,}584 KB & 777 KB \quad 793 KB & 4{,}361 KB / 4{,}377 KB per frame \\\bottomrule
\end{tabular}%
}
\end{table}

For FPGA (field-programmable gate array) implementation, each MLP instance uses 16 processing elements, each sustaining one MAC per cycle, or 2~FLOPs/cycle. Thus,

\begin{equation}
P_{\mathrm{peak,FPGA}} = \mathrm{Freq} \times N \times 32,
\label{eq:fpga_peak}
\end{equation}
where $N$ is the number of parallel MLP instances. This mapping fits the evaluated devices: $N=2$ uses 32 of the 56 CrossLink-NX multipliers, matching the reported 57\% utilization, while the Zynq ZU3EG has 360 DSP (digital signal processing) slices and can support $N=16$ within budget.

{The platform parameters are taken from device specifications: 360 DSP slices for Zynq ZU3EG~\cite{AMD_DS891} and 56 $18{\times}18$ multipliers for CrossLink-NX LIFCL-40~\cite{Lattice_CLNX_DS}. We use 600~MHz for Zynq PL (programmable logic) and 200~MHz for CrossLink-NX.}

Sequential~5$'$ is pixel-independent, so each stage can be fully pipelined without inter-pixel stalls. However, Stage~2 consumes the intermediate frame produced by Stage~1, so frame latency is the sum of the two stage latencies. Assuming four Stage~1 interpolation units and $N$ Stage~2 MLP instances, Stage~1 requires $C_1 \in [5,10]$ cycles per pixel, while Stage~2 with $h=32$ requires $2{,}048$ MACs per pixel. With 16 MACs/cycle per instance,
\[
C_2 = \frac{2{,}048}{16} = 128
\]
cycles per pixel. The resulting latency model is:
\begin{equation}
T_{\mathrm{latency}} \approx
\frac{H W}{4} \cdot \frac{C_1}{\mathrm{Freq}}
+
\frac{H W}{N} \cdot \frac{C_2}{\mathrm{Freq}}.
\label{eq:latency}
\end{equation}

For $H=W=256$, Zynq ZU3EG at $N=16$ and 600~MHz gives
\[
T_1 \approx 0.14\text{--}0.27~\mathrm{ms}, \qquad
T_2 \approx 0.87~\mathrm{ms},
\]
or $T_{\mathrm{latency}} \approx 1.01\text{--}1.15$~ms. CrossLink-NX at $N=2$ and 200~MHz gives
\[
T_1 \approx 0.41\text{--}0.82~\mathrm{ms}, \qquad
T_2 \approx 20.97~\mathrm{ms},
\]
or $T_{\mathrm{latency}} \approx 21.38\text{--}21.79$~ms.

\section{Limitations and Future Work}
\label{sec:supp-limitations}
Our approach has several limitations. First, the two-phase (formula-agnostic) scheme is bounded by the fidelity of the Phase~1 forward mimic: when $\mathrm{NN}_1$ cannot fully reproduce the true distortion system, the Phase~2 corrector trained through it inherits that error, and its correction becomes correspondingly less reliable. This shows in the ${\approx}0.9$\,dB gap between two-phase and known-formula training on Scenario~5; achievable predistortion is ultimately bounded by the invertibility and mimic-fidelity of the forward model rather than by the corrector's capacity. Second, the per-pixel tonal stage is resolution-bounded: trained at $256^2$, it does not exploit higher input resolution as well as a convolutional backbone, trailing CompenNet++ by ${\approx}3$\,dB at $1024^2$. Third, chromatic dispersion (color fringes) remains the artifact where our models lag the monolithic baselines, as the per-pixel tonal MLP lacks the multi-scale spatial support that dispersion correction requires. Fourth, we report FLOPs and memory transfers as proxies rather than measured energy: per-frame energy and memory-bandwidth utilization would require a synthesized implementation and are not reported here. Finally, and most importantly, our evaluation uses closed-form differentiable distortions and a single real projector--camera capture rather than through-the-optics measurements on physical AR hardware; per-device calibration, sensor noise, and temporal drift are not modelled. Future work includes improving Phase~1 mimic fidelity (e.g., higher-capacity or physically-grounded forward models), a wider-context or multi-scale tonal stage to address dispersion and higher resolutions, and validation on measured optical pipelines with perceptual metrics.

\end{document}